%% file: main_arxiv.tex
\documentclass[final,letterpaper,11pt]{article}

\usepackage{siunitx}
\usepackage{microtype}
\usepackage{subfigure}
\usepackage[utf8]{inputenc} 
\usepackage[T1]{fontenc}    
\usepackage{hyperref}       
\usepackage{url}            

\usepackage{amsmath, amssymb, amsthm, setspace, fancyhdr, lastpage, graphicx, caption, stackengine, longtable ,bbm, indentfirst, enumitem, amsbsy, color,bm, mathtools, relsize, latexsym}
\usepackage{upgreek}
\usepackage{psfrag,epsf}
\usepackage[toc,page]{appendix}
\usepackage{makecell}
\usepackage{multirow}
\usepackage{booktabs}  
\usepackage{url}

\newtheorem{theorem}{Theorem}[section] 
\newtheorem{lemma}{Lemma}[section] 
\newtheorem{corollary}{Corollary}[section]
\newtheorem{proposition}{Proposition}[section]
\newtheorem{definition}{Definition}[section]

\newtheorem{claim}{Claim}[section]
\newtheorem{remark}{Remark}[section]
\newtheorem{algorithm}{Algorithm}

\newcommand\independent{\protect\mathpalette{\protect\independenT}{\perp}}
\def\independenT#1#2{\mathrel{\rlap{$#1#2$}\mkern2mu{#1#2}}}

\usepackage{hyperref}
\usepackage[round]{natbib}
\usepackage[normalem]{ulem}

\allowdisplaybreaks

\newcommand{\rn}[1]{%
  \textup{\expandafter{\romannumeral#1}}%
}
\newcommand{\RN}[1]{%
  \textup{\uppercase\expandafter{\romannumeral#1}}%
}

\DeclareMathOperator*{\argmin}{\arg\!\min}
\newcommand{\E}{\mathbb{E}}
\newcommand{\Pb}{\mathbb{P}}

\newcommand{\var}{\text{var}}

\newcommand{\Qb}{\mathbb{Q}}
\newcommand{\Pn}{\mathbb{P}_n}

\newcommand{\R}{\mathbb{R}}
\DeclarePairedDelimiter\floor{\lfloor}{\rfloor}

\makeatletter
\newcommand{\customlabel}[2]{%
   \protected@write \@auxout {}{\string \newlabel {#1}{{#2}{\thepage}{#2}{#1}{}} }%
   \hypertarget{#1}{#2}
}
\makeatother

\usepackage{color}

\newcommand{\stout}[1]{\ifmmode\text{\sout{\ensuremath{#1}}}\else\sout{#1}\fi}
\usepackage{xcolor}

\allowdisplaybreaks

\begin{document}

\title{\bf {Causal effects based on distributional distances}\thanks{
The authors would like to thank Ilmun Kim for many helpful discussions. 
Part of this work was completed while Kwangho Kim was a doctoral student at Carnegie Mellon University. The authors report there are no competing interests to declare.
}}
\author{Kwangho Kim\thanks{Department of Statistics, Korea University, 145 Anam-ro, Seongbuk-gu, Seoul 02841, South Korea; email: \url{kwanghk@korea.ac.kr}.} \and Jisu Kim\thanks{Department of Statistics, Seoul National University, 1, Gwanak-ro, Gwanak-gu, Seoul 08826, Republic of Korea; email: \url{jkim82133@snu.ac.kr}.}\and Edward H. Kennedy\thanks{Department of Statistics \& Data Science, Carnegie Mellon University, 5000 Forbes Ave, Pittsburgh, PA 15213, United States; email: \url{edward@stat.cmu.edu}.}
}

\date{}
\maketitle

\begin{abstract}
Comparing counterfactual distributions can provide more nuanced and valuable measures for causal effects, going beyond typical summary statistics such as averages. In this work, we consider characterizing causal effects via distributional distances, focusing on two kinds of target parameters. The first is the counterfactual outcome density. We propose a doubly robust-style estimator for the counterfactual density and study its rates of convergence and limiting distributions. We analyze asymptotic upper bounds on the $L_q$ and the integrated $L_q$ risks of the proposed estimator, and propose a bootstrap-based confidence band. The second is a novel distributional causal effect defined by the $L_1$ distance between different counterfactual distributions. We study three approaches for estimating the proposed distributional effect: smoothing the counterfactual density, smoothing the $L_1$ distance, and imposing a margin condition. For each approach, we analyze asymptotic properties and error bounds of the proposed estimator, and discuss potential advantages and disadvantages. We go on to present a bootstrap approach for obtaining confidence intervals, and propose a test of no distributional effect. We conclude with a numerical illustration and a real-world example.
\end{abstract}

\noindent%

\pagenumbering{arabic}
\vspace*{.5in}

\section{Introduction}


Causal inference tools are increasingly being employed across a broad spectrum of disciplines, including medicine, sociology, criminology, business, and government research. Practitioners typically characterize causal effects by the mean difference in outcomes between treatment and control groups, commonly referred to as the average treatment effect (ATE):
\begin{equation} \label{def:the-ate}
\E[Y^1 - Y^0],
\end{equation}
where $Y^a$ denotes the \textit{counterfactual} or \textit{potential outcome} that would have been observed under $A=a$ \citep{rubin1974estimating}. Although this is a simple and intuitive summary, it may obscure crucial treatment effects on the outcome distribution. As a simple example, suppose that $Y^0=0$ yet $\Pb(Y^1=1)=\Pb(Y^1=-1)=1/2$, resulting in the ATE of exactly zero. In this case, the treatment has no effect on average, despite the fact that it harms half of the population while benefiting the other half. Similarly, a multimodal structure may suggest the presence of underlying subgroups with varying responses to a given treatment. Policymakers could greatly benefit from leveraging this more nuanced understanding to inform more effective decision-making.

Direct knowledge of the shape of counterfactual densities can be highly practical and insightful for real-world applications. Contrasting the shape of the density under different interventions could inform more nuanced information about treatment mechanism, especially in the presence of treatment heterogeneity. This could ultimately help optimize treatment policies or motivate the development of more effective treatments, such as lowering the risk of severe treatment side effects \citep{kennedy2021semiparametric}. Among the multiple objectives of this paper, one key goal is to develop an efficient estimator for the counterfactual outcome density, which poses significantly different challenges compared to standard counterfactual mean estimation. 

We also consider a novel causal effect defined by a distributional distance between counterfactual distributions. Letting $p_a$ denote the marginal density of $Y^a$, we target the $L_1$  distance between the densities $p_0, p_1$ corresponding to two treatment levels $A=0,1$, respectively, i.e.,
\begin{equation} \label{def:target-param-SRE}
D(p_1,p_0)  = \int |p_1(y) - p_0(y)| dy.
\end{equation}
In other words, the target parameter of interest in our estimation is defined by the integrated absolute difference between two counterfactual densities. The $L_1$ distance is a widely-used metric on spaces of probability measures, and has been studied extensively in the observational but not the causal setting \citep[e.g.,][]{lepski1999estimation,hallin2004kernel,jiao2018minimax,han2020estimation}. Unlike the case where distributional treatment effects are defined in terms of quantiles or cumulative distribution functions (CDFs), which are potentially complex curves, a one-number summary of distributional effects can be used as a simple first step in assessing whether there is effect modification beyond a mean shift.   

Although our methods can be easily extended to the general $L_r$ distance, $r \geq 1$, here we focus on the $L_1$ distance for a number of reasons. First, the $L_1$ distance is easily interpretable as the average absolute difference in densities, and has a unique connection to the probability scale: namely, half the $L_1$ distance is the total variation distance, the maximum possible difference in probabilities of the same event under $A=0$ versus $A=1$. For example if the $L_1$ distance is $D(p_1,p_0)=1/2$ then there is a set of outcomes whose probabilities under treatment versus control could differ by up to $25\%$. Further, it is invariant under monotone transformations of $Y$, which is not the case for many other distances, including the $L_2$ distance \citep{devroye1985}. Furthermore, even if one is interested in quantiles/CDFs, the $L_1$ distance can be used to test hypotheses that these quantities differ. Further, due to its non-smoothness, the $L_1$ distance introduces intriguing technical challenges that do not arise when estimating with general $L_r$ distances for $r > 1$.

\begin{figure}[t!]
\begin{center}
{\includegraphics[width=.55\textwidth]{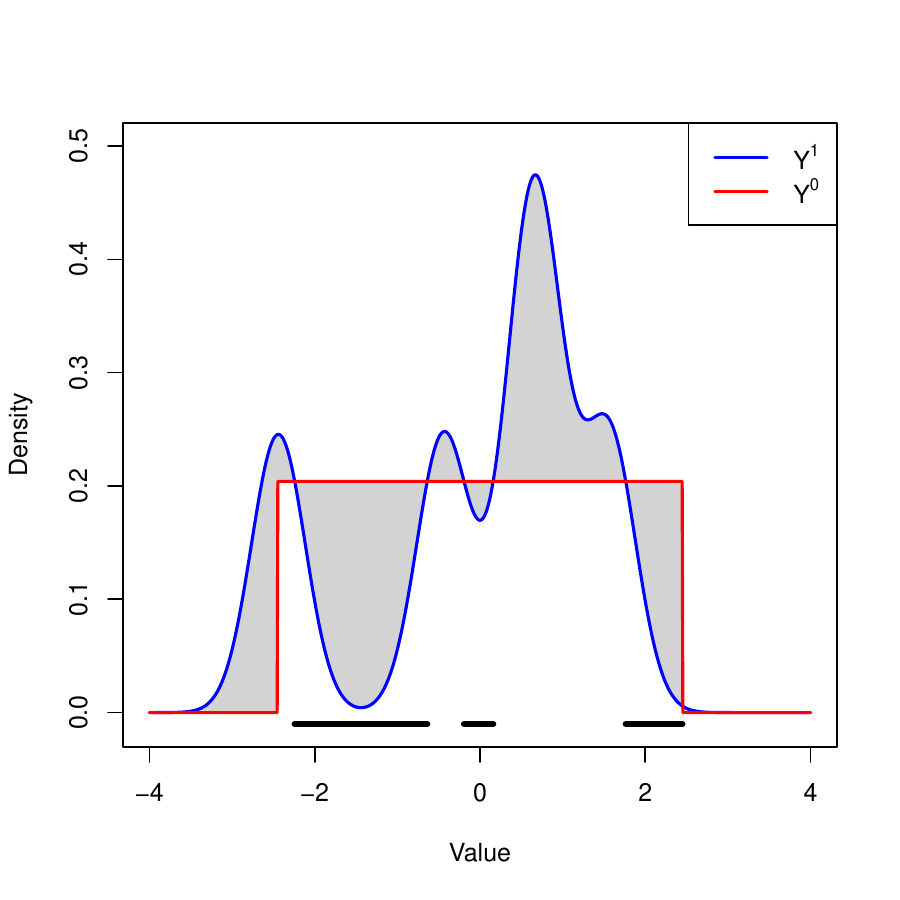}}
\caption{\em Densities for two counterfactual distributions, which have exactly the same mean and variance, but differ in $L_1$ distance (shaded area) by 0.628. The chance of having an outcome in the three black intervals along the x-axis is only 22.5\% under treatment, but 53.9\% under control, so the $L_1$ distance between the two distributions is 0.628.} \label{fig:example}
\end{center}
\end{figure}

As a simple illustration, consider the densities $p_0$ and $p_1$ displayed in Figure \ref{fig:example}. These have exactly the same mean (zero) and variance (two), but clearly differ in important ways. The distribution under control is simply uniform, whereas the distribution under treatment is multimodal, suggesting the presence of potentially important subgroups. Although the mean and variance are the same for the two distributions, the $L_1$ distance between them is approximately 0.628. This means that the chance of a subject in this population having some set of outcomes could differ by up to 31.4\% if they were treated versus had they taken control. For example, if treated, a subject here would have a 77.5\% chance of having an outcome outside the region $[-2.243,-0.638] \cup [-0.193, 0.159] \cup [1.764, 2.449]$ (denoted by the three black intervals in Figure \ref{fig:example}); however, under  control this chance would only be 46.1\%. (We note that this extremal set where the probability can differ most is simply the set of $y$ values where one density is larger, i.e., $p_1(y) \geq p_0(y)$ or its complement.)

\begin{remark} 
We stress that we view the proposed density-based causal effect as a complementary tool to the ATE and quantile- or CDF-based effects, providing potentially more nuanced information that may be useful; our goal is certainly not to do away with other standard causal effects entirely.
\end{remark}

\subsubsection*{Relation to previous work}
Here we give a brief review of some related literature, and refer to cited references for more details. A number of authors have considered distributional treatment effects defined in terms of quantiles or CDFs \citep[e.g.,][]{machado2005counterfactual, melly2006estimation, firpo2007efficient, rothe2010nonparametric, chernozhukov2013inference, landmesser2016decomposition, diaz2017efficient}. (Note that estimating the CDF of a distribution is sufficient for estimating its quantiles.) Just as in the non-counterfactual case, there is a substantial difference in targeting CDFs versus density-based treatment effects. Although arguably more visually appealing and interpretable, density estimation has an added statistical complexity; unlike CDFs, densities are not estimable at root-n rates in a nonparametric model. Moreover, efficient CDF estimators can be obtained simply by employing the standard doubly robust estimators with thresholded outcomes of the form $\mathbbm{1}(Y \leq y)$, just as in the spirit of counterfactual mean estimation. We refer to \citet{kennedy2021semiparametric} for more details on trade-offs in targeting CDF versus density effects.

In contrast to the quantile/CDF effects, density estimation with counterfactuals (or equivalently missing data) has received far less attention. There have been a few proposals, including by \citet{dinardo1996labor}, \cite{robins2001comment}, and  \cite{rubin2006extending}, which describe counterfactual density estimation as an interesting problem and propose some potential estimators, but do not analyze them in detail. They also do not consider distributional distance functionals based on the densities. Importantly, \citet{robins2001comment} proposed a doubly robust-style counterfactual density estimator, and gave a conjecture about its properties, but no proof. One of our contributions is analyzing a version of this estimator. More recently, \citet{westling2020unified} considered the related problem of density estimation in the presence of right censoring, but under a monotone density assumption. \citet{kennedy2021semiparametric} studied counterfactual density estimation by projecting the density function onto a finite-dimensional model and then developing an efficient estimator for the projected target. This differs from our approach based on \citet{robins2001comment}, which will be discussed in greater detail in Section \ref{sec:counterfactual-density-estimation}.

To our knowledge, only one paper has developed methods for distributional distance estimation in the counterfactual framework; \citet{kennedy2021semiparametric} proposed a nonparametric estimator based on a generalized, but twice differentiable distance metric.
Outside the counterfactual framework, recent work has made important advances in estimating distributional distances from observational data. \cite{Kandasamy14} studied estimation of smooth distance functionals using the theory of influence functions and sample splitting. For the non-smooth $L_1$ distance, \cite{jiao2018minimax} gave minimax lower bounds in the non-counterfactual and discrete case. \citet{lepski1999estimation} and \citet{han2020estimation} have studied nonparametric estimation of general $L_r$-norms and gave minimax rates for any $r\geq 1$, in the non-counterfactual and continuous case. Further references can be found in aforementioned papers. 


\subsubsection*{Overview of contributions}
We make nine main contributions. First, based on the proposal by \citet{robins2001comment}, we develop doubly-robust style methods for nonparametric counterfactual density estimation and analyze their asymptotic properties. Second, we give the associated pointwise and global $L_q$ risk, which will be useful to assess optimality of the proposed estimator. Third, we define a novel distributional causal effect \eqref{def:target-param-SRE} measuring the difference between counterfactual densities in terms of the $L_1$ distance. Next, we go on to develop efficient estimators for the proposed causal effect, based on the three different approaches, namely by using 4) the estimated counterfactual densities, 5) a smooth approximation of the absolute value function, and 6) the margin condition. Further we study their rates of convergence and the $L_2$ risk, as well as propose a bootstrap approach for constructing confidence bands/intervals. Eighth, based on our theoretical finding, we develop a novel test of no distributional effect. Finally we conclude with a simple numerical illustration and a real-world example.

\section{Preliminaries}

\subsection{Setup and Identification} \label{sec:setup} 
Throughout, we consider treatments $A \in \mathcal{A}\coloneqq\{0,1\}$ that are binary (or without loss of generality, discrete), real-valued continuous outcomes $Y \in \mathcal{Y} \subseteq \mathbb{R}^d$, and covariates $X \in \mathcal{X} \subseteq \mathbb{R}^k$. Although $d=1$ is the most common case in practice, we allow $d > 1$. Suppose we have access to an i.i.d. sample $(Z_1,...,Z_n)$ where $Z=(X,A,Y)\sim\Pb$ for some distribution $\Pb$. In observational studies, the treatment happened naturally according to some unknown process, and was not under experimenter’s control. Thus, we require the following standard identification assumptions \citep[e.g.,][Chapter 12]{imbens2015causal}:
\begin{itemize}[leftmargin=*]
	\item \customlabel{assumption:c1}{(C1)} \textit{Consistency}: $Y=Y^a$ if $A=a$
	\item \customlabel{assumption:c2}{(C2)} \textit{No unmeasured confounding}: $A \independent Y^a \mid X$
	\item \customlabel{assumption:c3}{(C3)} \textit{Positivity}: $\Pb(A=a|X) \geq \varepsilon  \text{ a.s. }$ for some $\varepsilon>0$
\end{itemize}
Consistency \ref{assumption:c1} ensures that potential outcomes are defined uniquely by a subject’s own treatment level, not others’ levels (i.e., no interference). No unmeasured confounding \ref{assumption:c2} (or exhangeability) will hold if the collected covariates can explain treatment assignment, to the extent that after conditioning on them treatment is not further related to potential outcomes. Positivity \ref{assumption:c3} requires everyone to have some chance of being treated at all treatment levels. Throughout the paper, we assume that Assumptions \ref{assumption:c1} - \ref{assumption:c3} are valid.

Our setting includes randomized studies as they are simply a subset of observational studies (where the exhangeability condition holds by design). When $X = \emptyset$ and $\Pb(A=1) = 1/2$, it becomes a classic randomized controlled trial. When $X$ has $N$ discrete values, i.e., $X = \{1,...,N\}$, then we would have a multi-source randomized experiment where we have $N$ distinct sources of randomization (or a conditionally randomized experiment with discrete $X$).

Let $\mathcal{B}$ denote the Borel $\sigma$-field of subsets of $\mathcal{Y}$, and $B \in \mathcal{B}$. Then the conditional distribution and marginal density of the counterfactual outcome $Y^a$ are identified as
\begin{align*}
    \Pb\left(Y^a \in B \mid X=x\right) &= \Pb\left(Y \in B \mid X=x, A=a \right) \\
    &= \int_{B} \nu_a(y \mid x)dy
\end{align*}
and
\begin{align} \label{eqn:counterfactual-density}
    p_a(y)= \int_{\mathcal{X}} \nu_a(y \mid x)d\Pb(x),
\end{align}
respectively.
$\nu_a$ is the conditional density function before averaged over the covariates; when $d=1$, $\nu_a(y \mid x) = \frac{\partial}{\partial y}\Pb\left( Y \leq y \mid X=x, A=a \right)$. 

In this paper, we are essentially concerned with two kinds of target parameters. The first is the marginal counterfactual density $p_a$ given in \eqref{eqn:counterfactual-density}, and the second is the $L_1$ distance between two counterfactual densities defined in \eqref{def:target-param-SRE}, which is identified by
\begin{equation} \label{identify.exp::obs}
\begin{aligned} 
\psi \coloneqq D(p_1, p_0) &= \int |{p_1}(y) - {p_0}(y)| dy \\
&=\int \left|  \int \Big\{ \nu_1(y \mid x) -  \nu_0(y \mid x) \Big\} d\Pb(x) \right| dy.
\end{aligned}
\end{equation}
As discussed in the previous section, $\psi$ represents the proposed density-based distributional causal effect, enabling us to capture more subtle information on effect modification.

\textbf{Notation.} Hereafter, for a given function $f$, we use the notation
$\|f\|_{q}=\left(\int|f(z)|^{q}d\mathbb{P}(z)\right)^{\frac{1}{q}}$
as the $L_{q}(\mathbb{P})$-norm of $f$. When applied to finite-dimensional vectors, $\|\cdot\|_{q}$ is understood as the standard $q$-norm. In particular, we let $\Vert f \Vert$ denote the $L_{2}(\mathbb{P})$-norm in order to simplify notation and avoid any confusion with the Euclidean norm $\|\cdot\|_{2}$. Furthermore, we let ${\Pb}$ denote the conditional expectation given the sample operator $\hat{f}$, as in $\mathbb{P}(\hat{f})=\int\hat{f}(z)d\mathbb{P}(z)$. Notice that $\Pb(\hat{f})$ is random only if $\hat{f}$ depends on samples, in which case $\Pb(\hat{f}) \neq \E(\hat{f})$. Otherwise $\Pb$ and $\E$ can be used exchangeably. For example, if $\hat{f}$ is constructed on a separate (training) sample $\mathsf{D}^n = (Z_1,...,Z_n)$, then ${\Pb}\left\{\hat{f}(Z)\right\} = \E\left\{\hat{f}(Z) \mid \mathsf{D}^n \right\}$ for a new observation $Z \sim \Pb$. Lastly, we use the shorthand $a_n \lesssim b_n$ to denote $a_n \leq C b_n$ for some universal constant $C > 0$.

\subsection{Bootstrap and Stochastic Convergence of Empirical Process}
\label{subsec:bootstrap-validity}

To construct valid confidence intervals for further quantifying the uncertainty in our estimates, we employ bootstrap methods grounded in empirical process theory. Originally introduced by \citet{Efron1979}, bootstrapping is a technique for estimating the variance of an estimator, thereby enabling the construction of confidence sets. 
The asymptotic validity of the bootstrap procedure requires the stochastic convergence of an empirical process. Here, we provide a brief review for the key techniques that are essential to construct valid confidence sets of our proposed estimators. We refer to \citet{van1996weak, vanderVaart2000, Kosorok2008, gine2021mathematical} for further details.

Suppose an i.i.d sample $(Z_{1},...,Z_{n})\sim\mathbb{P}$
on $\mathcal{Z}$, and let $\mathbb{P}_{n}=\frac{1}{n}\sum_{i=1}^{n}\delta_{Z_{i}}$
be the empirical measure. Let $(Z_{1}^{*},\ldots,Z_{n}^{*})$ be a
bootstrapped sample drawn with replacement from the original
sample $(Z_{1},\ldots,Z_{n})$, and let $\mathbb{P}_{n}^{*}=\frac{1}{n}\sum_{i=1}^{n}\delta_{Z_{i}^{*}}$
be the corresponding bootstrap empirical measure. Bootstrapping is used to learn about the unknown measure $\mathbb{P}_{n}-\mathbb{P}$ using the known and computable
measure $\mathbb{P}_{n}^{*}-\mathbb{P}_{n}$.

One theoretical guarantee for bootstrapping is that $\sqrt{n}(\mathbb{P}_{n}-\mathbb{P})$
and $\sqrt{n}(\mathbb{P}_{n}^{*}-\mathbb{P}_{n})$ converges to the same
Brownian Bridge. Let $\mathcal{F}\subset\mathbb{R}^{\mathcal{Z}}$
be a class of measurable functions. We let $\ell_{\infty}(\mathcal{F})$
be the collection of all bounded functions $\phi:\mathcal{F}\to\mathbb{R}$
equipped with the sup norm (or uniform norm) $\|\cdot\|_{\infty}$. A random measure
$\mu$ is understood in $\ell_{\infty}(\mathcal{F})$ as $\mu(f)=\int fd\mu$.
For random measures $\{\mu_{n}\}_{n\in\mathbb{N}}$ and $\mu$, we
say $\mu_{n}\to\mu$ weakly in $\ell_{\infty}(\mathcal{F})$ if and
only if $\mathbb{E}\left[\phi(\mu_{n})\right]\to\mathbb{E}\left[\phi(\mu)\right]$
for every bounded continuous map $\phi:\ell_{\infty}(\mathcal{F})\to\mathbb{R}$. This is formalized in the following theorem.

\begin{theorem}
	\label{thm:weakconv-bootstrap}
	(\citet[Theorem 2.6]{Kosorok2008})
	$\sqrt{n}(\mathbb{P}_{n}-\mathbb{P})\to\mathbb{G}$ weakly in $\ell_{\infty}(\mathcal{F})$
	 if and only if $\sqrt{n}(\mathbb{P}_{n}^{*}-\mathbb{P}_{n})\to\mathbb{G}$
	a.s. weakly in $\ell_{\infty}(\mathcal{F})$ for a limit process $\mathbb{G}$. If either convergence happens, the limit process $\mathbb{G}$ is a centered Gaussian process
	with $Cov[\mathbb{G}(f),\mathbb{G}(g)]=\int fgd\mathbb{P}-\int fd\mathbb{P}\int gd\mathbb{P}$ for $f,g \in \mathcal{F}$.
\end{theorem} 

Therefore, once $\sqrt{n}(\mathbb{P}_{n}-\mathbb{P})\to\mathbb{G}$
weakly in $\ell_{\infty}(\mathcal{F})$ is shown, Theorem \eqref{thm:weakconv-bootstrap}
implies that $\sqrt{n}(\mathbb{P}_{n}^{*}-\mathbb{P}_{n})\to\mathbb{G}$
weakly in $\ell_{\infty}(\mathcal{F})$ a.s. as well, and the unknown
measure $\sqrt{n}(\mathbb{P}_{n}-\mathbb{P})$ can be asymptotically
approximated by the known and computable measure $\sqrt{n}(\mathbb{P}_{n}^{*}-\mathbb{P}_{n})$.
One way to show $\sqrt{n}(\mathbb{P}_{n}-\mathbb{P})\to\mathbb{G}$
weakly in $\ell_{\infty}(\mathcal{F})$ (i.e., $\mathcal{F}$ is $\Pb$-Donsker) is to use the bracketing entropy argument as detailed in, for example, \citet[][Chapter 2.5]{van1996weak}. Appendix \ref{sec:appendix-empirical-processes} provides extra details on the essential technical tools used in our inferential approaches.

	
	
	

\section{Counterfactual Density Estimation} \label{sec:counterfactual-density-estimation}

\subsection{Proposed Estimator and Asymptotic Properties}

Finding an efficient estimator for the counterfactual density is challenging and still an open problem. In observational studies, the identifying expression \eqref{identify.exp::obs} involves conditional densities that depend not only on $Y$ but also on potentially continuous, higher-dimensional covariates $X \in \mathbb{R}^k$. Such conditional densities cannot be estimated at root-n rates in nonparametric models \citep{efromovich2007conditional}, and a plug-in estimator for $p_a$ based on the conditional density estimates typically yields slow and suboptimal convergence rates \citep{bickel1993efficient,van2003unified}.

In this section, we develop an efficient estimator for $p_a$ by targeting the kernel-smoothed version of the true counterfactual density
\begin{equation} \label{eqn:psi_h}
 p_{a,h}(y) \coloneqq \E \left\{ \frac{1}{h^d}K\left(\frac{\Vert Y^a-y \Vert_2}{h}  \right) \right\},
\end{equation}
with a kernel $K:\mathbb{R}^{d}\to\mathbb{R}$ that is an integrable function satisfying $\int K(u)du = 1$. As in the classical kernel density estimation, the smoothing bias vanishes as the bandwidth $h$ goes to zero. To the best of our knowledge, this approach was first proposed by \cite{robins2001comment}, though without any detailed theoretical exploration, and has not been formally studied since then. Our approach also differs from the recent work by \citet{kennedy2021semiparametric} in which the density is approximated with some pre-specified finite-dimensional model.

The kernel-smoothed counterfactual density in \eqref{eqn:psi_h} is identified as
\begin{equation} \label{eqn:p_a_h}
    \begin{aligned}
    p_{a,h}(y) &= \E \left\{ \E \left[ \frac{1}{h^d}K\left(\frac{ \Vert Y-y \Vert_2}{h} \right) \Big\vert X, A=a  \right] \right\}\\
    & \equiv \E \left\{ \E \left[ K_{h,y}(Y) \Big\vert X, A=a  \right] \right\},
    \end{aligned}
\end{equation}
where we let $ K_{h,y}(Y)=\frac{1}{h^d}K\left(\frac{\Vert Y-y \Vert_2}{h}  \right)$. The identifying expression for $D(p_{1,h},p_{0,h})$ follows directly from substituting \eqref{eqn:p_a_h}, forming the basis of our first estimator for the distributional effect, as discussed in Section \ref{subsec:density-effects-smooth-pa}.

Before proceeding, we make the following assumption on the marginal density $p_a$ and the kernel function $K$.

\begin{itemize}[leftmargin=*]
        \item \customlabel{assumption:A1}{(A1)} \textit{Bounded kernel with bounded support}: The kernel function $K$ is bounded in $L_{\infty}$, i.e.,
	  $\Vert K \Vert_\infty < \infty$, and there exists $R_{K}<\infty$ such that $\mathrm{supp}(K)\subset\mathbb{B}_{R_{K}}(0)$
	where $\mathbb{B}_{R_{K}}(0)=\{u\in\mathbb{R}^{d}:\,\|u\|_{2}\leq R_{K}\}$.
\end{itemize}

Assumption \ref{assumption:A1} is mild, as $K$ is user-specified. Indeed, it holds for various types of kernel functions in common use \citep[e.g.,][]{Tsybakov10, GyorfiKKW2002}. To simplify notation, we introduce the following nuisance functions
\begin{align*}
    & \pi_a(X)=\Pb(A=a \mid X), \\
    & \mu_{a,y}(X)=\E\{K_{h,y}(Y) \mid X, A=a\},
\end{align*}
and let $\widehat{\pi}_a, \widehat{\mu}_{a,y}$ be some estimators of ${\pi}_a, {\mu}_{a,y}$, respectively. We drop the dependence of $\mu_{a,y}$ on $h$ to ease notation. We use $\eta$ to denote a set of the nuisance functions $\{{\pi}_a, {\mu}_{a,y}: a \in \mathcal{A}, y \in \mathcal{Y}\}$. 

Now we propose our estimator for the smoothed counterfactual density $p_{a,h}$ as
\begin{align} \label{eqn:doubly-robust}
\widehat{p}_{a,h}(y) \equiv \widehat{p}_{a,h}(y; \widehat{\eta}) = \Pb_n \left\{ \frac{\mathbbm{1}(A=a)}{\widehat{\pi}_a(X)}\Big(K_{h,y}(Y) - \widehat{\mu}_{A,y}(X) \Big) +\widehat{\mu}_{a,y}(X) \right\}.
\end{align}
The estimator \eqref{eqn:doubly-robust} resembles the doubly robust (or semiparametric) estimator for the ATE, except that we have replaced $Y$ in the original estimator with the kernel-weighted outcome $K_{h,y}(Y)$. When $h$ is fixed and not depending on sample size, \eqref{eqn:doubly-robust} is $\sqrt{n}$-consistent, asymptotically normal, and efficient, even when $\pi_a$ and $\mu_{a,y}$ are estimated flexibly with nonparametric methods without necessarily committing a priori to particular estimators. For more details on the doubly robust estimators and related topics, we refer the interested readers to \cite{bickel1993efficient, tsiatis2006semiparametric, van2003unified, kennedy2016semiparametric, kennedy2022semiparametric} and references therein. Shortly, we will discuss the case whereby $h$ varies with sample size as well.


\begin{remark} \label{remark:mu-and-conditional-density}
A natural way to construct the estimator $\widehat\mu_{a,y}$ is to regress the kernel-weighted outcome $K_{h,y}(Y)$ on $(X,A)$. Another option would be to estimate the conditional density $\nu_a(\cdot \mid x)$ first and then use
$$ \widehat\mu_{a,y}(x) = \int K_{h,y}(u) \widehat{\nu}_a(u \mid x) \ du. $$
It follows that under Assumption \ref{assumption:A1}, 
\begin{align*}
    \left\Vert \widehat{\mu}_{a,y}(X)-{\mu}_{a,y}(X) \right\Vert 
    \leq h\Vert K \Vert_{\infty} R_K^d \underset{y}{\sup} \left\Vert \widehat{\nu}_a(y \mid X) - {\nu}_a(y \mid X) \right\Vert.
\end{align*}
\end{remark}

Before we analyze the error bounds, we enumerate additional assumptions regarding our nuisance estimators as below.

\begin{itemize}[leftmargin=*]
	\item \customlabel{assumption:A2}{(A2)} \textit{Uniform boundedness for $1/\widehat{\pi}_{a}$}: $\left\Vert 1/\widehat{\pi}_{a}\right\Vert _{\infty}$ is finite. 
	
	\item \customlabel{assumption:A3}{(A3)} \textit{Sample splitting}: The nuisance estimators are computed in a separate independent sample. 
\end{itemize}
Assumption \ref{assumption:A3} enables us to accommodate the added complexity from estimating both nuisance and density functions without relying on complicated empirical process conditions (e.g., Donsker-type or low entropy conditions \citep{vanderVaart2000}). If one is willing to rely on appropriate empirical process conditions instead of \ref{assumption:A3}, then the nuisance components can be estimated on the same sample without sample splitting. However this would limit the flexibility of the nuisance estimators \citep{kennedy2016semiparametric, kennedy2022semiparametric}. 

\begin{remark}[Sample splitting] \label{rmk:sample-splitting}
For nuisance estimation, we can always create separate independent samples by splitting the data in half (or in folds) at random. The full sample size efficiency can be attained by
swapping the samples as in cross-fitting \citep[e.g.,][]{zheng2010asymptotic, kennedy2016semiparametric, Chernozhukov17, newey2018cross}. We analyze a single split procedure in this paper for simplicity and theoretical emphasis. However, extending to averages across independent splits is straightforward (see \citet{kennedy2019nonparametric} for an example).
\end{remark}

In the next theorem, we give pointwise asymptotic bounds for $\widehat{p}_{a,h}(y)-p_{a}(y)$. 

\begin{theorem}\label{thm:asymptotics-pa-hat}
    Suppose $p_a(y) \leq p_{\max}$ for some $p_{\max} < \infty$, $\forall y \in \mathcal{Y}$. Then under Assumptions \ref{assumption:A1} - \ref{assumption:A3}, we have
    \begin{align*}
    \widehat{p}_{a,h}(y)-p_{a}(y) & = (\mathbb{P}_{n}-\mathbb{P})({f}_{h,y}^{a}) + p_{a,h}(y) - p_{a}(y) \\
    & \quad + O_\Pb\left( \Vert\widehat{\mu}_{a,y}-{\mu}_{a,y}\Vert\Vert\widehat{\pi}_{a}-{\pi}_{a}\Vert + R_{h,n} \right),
    \end{align*}
    where $f_{h,y}^{a}(Z) \coloneqq \frac{\mathbbm{1}(A=a)}{{\pi}_{a}(X)}\left(K_{h,y}(Y)-{\mu}_{A,y}(X)\right)+{\mu}_{a,y}(X)$, and    
    \begin{align*}
        R_{h,n} = 
        \begin{cases}
        \frac{\left\Vert {\pi}_{a} - \widehat{\pi}_{a} \right\Vert_\infty}{\sqrt{nh^d}} + \frac{\left\Vert \widehat{\mu}_{a,y}-{\mu}_{a,y} \right\Vert}{\sqrt{n}}, & \text{for variable } h\\
        \frac{\left\Vert {\pi}_{a} - \widehat{\pi}_{a} \right\Vert + \left\Vert \widehat{\mu}_{a,y}-{\mu}_{a,y} \right\Vert}{\sqrt{n}},       & \text{for fixed } h.
        \end{cases}
    \end{align*}
\end{theorem} 

A proof of Theorem \ref{thm:asymptotics-pa-hat} and all other results can be found in the Appendix. 
If we assume that $p_a$ is smooth enough, the bounded-density condition $p_a \leq p_{\max} < \infty$ could be replaced by a specific kernel characteristic (see Remark \ref{rmk:new-boundedness-condition-on-kernel}).
{Importantly, the result in Theorem \ref{thm:asymptotics-pa-hat} verifies and expands on the conjectured rate from \citet{robins2001comment}. In particular, when combined with the standard bounds on the smoothing bias (discussed further in the following subsection), this shows that the doubly robust kernel estimator can achieve the same rate as an oracle with access to the actual counterfactuals, under some weak high-level conditions on the nuisance error. This is similar in spirit to recent results in heterogeneous treatment effect estimation \citep{nie2017quasi, foster2019orthogonal, kennedy2020optimal}, but to the best of our knowledge has not been provided for density estimation.} 

The following nonparametric condition on nuisance component estimators is pretty standard in the literature, and could be satisfied under some structural conditions on the nuisance functions such as smoothness and sparsity \citep[][Section 4]{kennedy2016semiparametric}.
\begin{itemize}[leftmargin=*]
	\item \customlabel{assumption:A4}{(A4)} \textit{Nonparametric conditions on $\widehat{\pi}_{a}, \widehat{\mu}_{a,y}$}: For any $a \in \mathcal{A}$, $y \in \mathcal{Y}$, $\Vert\widehat{\mu}_{a,y}-{\mu}_{a,y}\Vert = o_\Pb(1)$, $\Vert\widehat{\pi}_{a}-{\pi}_{a}\Vert = o_\Pb(1)$ and $\Vert\widehat{\mu}_{a,y}-{\mu}_{a,y}\Vert\Vert\widehat{\pi}_{a}-{\pi}_{a}\Vert=o_{\Pb}(n^{-\frac{1}{2}})$. 
\end{itemize}
If we further assume \ref{assumption:A4} under the fixed bandwidth condition, by Theorem \ref{thm:asymptotics-pa-hat} and the central limit theorem, we have the following limiting distribution with respect to the smoothed density $p_{a,h}(y)$ for any $a \in \mathcal{A}$, $y \in \mathcal{Y}$: 
\begin{align} \label{eqn:CLT-density-pointwise}
    \widehat{p}_{a,h}(y)-p_{a,h}(y) \overset{d}{\longrightarrow} N\left(0,\var(f_{h,y}^{a})\right).
\end{align}

\eqref{eqn:CLT-density-pointwise} could be used to construct a pointwise confidence interval with respect to $p_{a,h}$. To obtain uniform confidence bands across $y \in \mathcal{Y}$, we need a more comprehensive result that describes the asymptotic behavior of $\widehat{p}_{a,h}(y)$ across a continuum of $y$ values. This will be discussed in Section \ref{subsec:confidence_band_for_density} in detail.

\subsection{Risk Bounds}

Here, we analyze asymptotic upper bounds on the $L_q$ and the integrated $L_q$ risks, which will be useful for assessing optimality of the proposed estimator. We start with introducing some additional regularity conditions on the density $p_a$ and on the kernel $K$. Let $\floor{\beta}{}$ denote the greatest integer strictly less than $\beta \in \R$. Also, given a vector $\alpha=(\alpha_1,\ldots,\alpha_d)^\top \in \R^d$ and a multivariate function $f:\R^d \rightarrow \R$, define $D^{(\alpha)}f = \frac{\partial^{\alpha_1+ \cdots \, + \alpha_d}}{\partial x_1^{\alpha_1} \cdots \, \partial x_d^{\alpha_d}}f$, the order-$\alpha$ partial derivatives of $f$. Then we consider the \emph{H\"{o}lder class} $\Sigma(\beta,L)$ on $\R^d$ defined as the set of $l = \floor{\beta}{}$ times differentiable functions $f : \R^d \rightarrow \R$ such that for some positive numbers $\beta, L$,
\[
 \vert D^{(l)}f(v) - D^{(l)}f(w) \vert \leq L\Vert v - w \Vert_1^{\beta-l}, \quad \forall v,w \in \mathbb{R}^d,
\]
and denote by $\mathcal{P}_\Sigma(\beta,L)$ the associated class of densities, defined as
\[
\mathcal{P}_\Sigma(\beta,L)=\left\{p \biggm\vert p \geq 0, \int p(v)dv = 1, \text{and } p \in \Sigma(\beta,L) \text{ on } \R^d \right\}.
\]
Further, for an integer $l \geq 1$, we say that $K$ is a \emph{kernel of order} $l$ if 
\[
\int K(u)du = 1, \qquad \int u^jK(u)du = 0, \, j=1,\ldots,l.
\]
See \citet[][Section 1.2]{Tsybakov10} for examples of the kernel of order $l$. Now we have the following result on the {$L_q$ risk} bounds for the proposed estimator $\widehat{p}_{a,h}$ at an arbitrary fixed point $y$.

\begin{theorem} \label{thm:Lq-risk-pointwise} %
Suppose that $p_a \in \mathcal{P}_\Sigma(\beta,L)$ and $K$ is a kernel of order $\floor{\beta}{}$ satisfying $\int \Vert u \Vert_1^\beta \vert K(u) \vert du < \infty$. Further, define the stochastic term:
\begin{align} \label{eqn:remainder-1}
    R_{1,n}^q = n^{-\frac{q}{2}} \left\Vert \widehat{\mu}_{a,y} - {\mu}_{a,y} \right\Vert_q^q + \left\Vert\widehat{\mu}_{a,y}-{\mu}_{a,y}\right\Vert^q \left\Vert\widehat{\pi}_{a}-{\pi}_{a}\right\Vert^q.
\end{align}
Then under Assumptions \ref{assumption:A1} - \ref{assumption:A3}, for $2 \leq q < \infty$, $\forall y \in \R$, $\forall h > 0$, we have
\begin{align*}
    \left\Vert \widehat{p}_{a,h}(y)-p_{a}(y) \right\Vert_q^q \lesssim n^{-(q-1)}h^{-d(q-1)} + n^{-\frac{q}{2}}h^{-\frac{dq}{2}} + h^{q\beta} + R_{1,n}.
\end{align*}
Hence, taking $h \asymp n^{-\frac{1}{2\beta + d}}$, the proposed estimator satisfies
\begin{align*}
    \underset{y \in \mathcal{Y}}{\sup} \underset{p_a \in \mathcal{P}_\Sigma(\beta,L)}{\sup}\left\Vert \widehat{p}_{a,h}(y)-p_{a}(y) \right\Vert_q^q  \lesssim n^{-\frac{q\beta}{2\beta+d}} + R_{1,n}.
\end{align*}
\end{theorem}

\begin{remark} \label{rmk:new-boundedness-condition-on-kernel}
    The condition $\int \vert K(u) \vert \Vert u \Vert_1^\beta du < \infty$ does, in fact, imply the bounded-density condition $p_a(y) \leq p_{\max} < \infty$, $\forall y \in \mathcal{Y}$; it could be shown that one may set $p_{\max} = O\left( \int \vert K(u) \vert \Vert u \Vert_1^\beta du + \left\Vert K \right\Vert_\infty \right)$ (see Remark \ref{rmk:thm3.2-p_max-representation} in the Appendix).
\end{remark}

The analysis of the global risk of $\widehat{p}_{a,h}$ will also be informational. An important global criterion is the \emph{integrated $L_q$ risk} defined by
\begin{align*}
    \Pb\left\{ \int \left\vert \widehat{p}_{a,h}(y)-p_{a}(y) \right\vert^q dy \right\}.
\end{align*}

To this end, we consider the \emph{Nikol’ski class} of functions $\mathcal{H}(q,\beta,L)$, defined as the set of $l = \floor{\beta}{}$ times differentiable functions $f : \R^d \rightarrow \R$ satisfying
\[
\left[ \int \left\vert D^{(l)}f(v+t) - D^{(l)}f(v) \right\vert^q dv \right]^{1/q} \leq L\Vert t \Vert_1^{\beta-l}, \quad \forall t \in \R^d,
\]
and define the associated class of densities as
\[
\mathcal{P}_\mathcal{H}(q,\beta,L)=\left\{p \biggm\vert p \geq 0, \int p(v)dv = 1, \text{and } p \in \mathcal{H}(q,\beta,L) \text{ on } \R^d \right\}.
\]


In the following theorem, we analyze upper bounds on the integrated $L_q$ risk.

\begin{theorem} \label{thm:Lq-risk-integrated} %
Suppose that $p_a \in \mathcal{P}_\mathcal{H}(q,\beta,L)$ such that $\Vert p_a \Vert_{L_\frac{q}{2}} < \infty$, $\forall 2 \leq q < \infty$. Also, let $K$ be a kernel of order $\floor{\beta}{}$ satisfying $\int \Vert u \Vert_1^\beta \vert K(u) \vert du < \infty$. Define the stochastic term:
\begin{align} \label{eqn:remainder-2}
    R_{2,n}^q = n^{-\frac{q}{2}} \int \left\Vert \widehat{\mu}_{a,y} - {\mu}_{a,y} \right\Vert_q^q dy + \left\Vert\widehat{\pi}_{a}-{\pi}_{a}\right\Vert^q \int \left\Vert\widehat{\mu}_{a,y}-{\mu}_{a,y}\right\Vert^q dy.
\end{align} 
Then, under Assumptions \ref{assumption:A1} - \ref{assumption:A3}, $\forall h > 0$, we have
\begin{align*}
    \Pb\left\{ \int \left\vert \widehat{p}_{a,h}(y)-p_{a}(y) \right\vert^q dy \right\} & \lesssim n^{-(q-1)} h^{-d(q-1)} + n^{-\frac{q}{2}} h^{-\frac{dq}{2}} + h^{q\beta} + R_{2,n}
\end{align*}
Furthermore, taking $h \asymp n^{-\frac{1}{2\beta + d}}$, the proposed estimator satisfies
\begin{align*}
    \underset{p_a \in \mathcal{P}_\mathcal{H}(q,\beta,L)}{\sup} \Pb\left\{ \int \left\vert \widehat{p}_{a,h}(y)-p_{a}(y) \right\vert^q dy \right\}  \lesssim n^{-\frac{q\beta}{2\beta+d}} + R_{2,n}.
\end{align*}
\end{theorem}

Some people may find analysis of error bounds with respect to $p_{a,h}$ for the fixed $h$ more beneficial, especially when they do not want to impose strong regularity conditions on the true density and the kernel function. In what follows we present the result for the relatively simple fixed-bandwidth case as well.

\begin{corollary} \label{cor:Lr-risk-fixed-h}
Suppose that Assumptions \ref{assumption:A1} - \ref{assumption:A3} hold. Then, for some fixed $h$ and $\quad 2 \leq q < \infty$, we have
\[
\left\Vert\widehat{p}_{a,h}(y)-p_{a,h}(y)\right\Vert_q^q \lesssim \left\Vert\widehat{\mu}_{a,y}-{\mu}_{a,y}\right\Vert^q \left\Vert\widehat{\pi}_{a}-{\pi}_{a}\right\Vert^q + n^{-\frac{q}{2}}, 
\]
and
\[
\Pb\left\{ \int \left\vert \widehat{p}_{a,h}(y)-p_{a,h}(y) \right\vert^q dy \right\} \lesssim \left\Vert\widehat{\pi}_{a}-{\pi}_{a}\right\Vert^q \int \left\Vert\widehat{\mu}_{a,y}-{\mu}_{a,y}\right\Vert^q dy + n^{-\frac{q}{2}}.
\]
\end{corollary}

The discussion on variable- versus fixed-bandwidth analysis becomes more nuanced in statistical inference. We will revisit this issue in Section \ref{subsec:confidence_band_for_density} in more detail.

\subsubsection{Connections to non-counterfactual minimax bounds}

Theorems \ref{thm:Lq-risk-pointwise}, \ref{thm:Lq-risk-integrated} and Corollary \ref{cor:Lr-risk-fixed-h} provide an important clue for assessing optimality of the proposed estimator. 
One may compare our rates to the non-counterfactual "oracle" minimax lower bounds. Minimax approaches in nonparametric density estimation have been extensively studied over the decades; the reader is referred to, for example, \citet{hasminskii1990density, Tsybakov10, hardle2012wavelets} for a thorough overview. Our results in Theorem \ref{thm:Lq-risk-pointwise} and Theorem \ref{thm:Lq-risk-integrated} show that under appropriate rate requirements on nuisance estimation, in the non-counterfactual setting, the proposed estimator attains the minimax lower bound with respect to the pointwise and the integrated risk, associated with the class of densities $\mathcal{P}_\Sigma(\beta,L)$ and $\mathcal{P}_\mathcal{H}(q,\beta,L)$, respectively.

The minimax lower bounds for counterfactual density estimation - when the oracle rate is not achievable - are unknown. However, we conjecture that our proposed estimator is minimax optimal as well, since the counterfactual case only adds complexity in the form of extra nuisance estimation. In particular, we believe that it resembles the minimax rate for conditional average treatment effects, which combines rates for nonparametric regression and functional estimation \citep{kennedy2022minimax}. We omit a formal proof as it is beyond the scope of this paper.

A similar argument can be made for the fixed-bandwidth case. From Corollary \ref{cor:Lr-risk-fixed-h}, for example when $q=2$, we may achieve fast $\sqrt{n}$ rates if the second-order nuisance errors converge at faster-than-$\sqrt{n}$ rates. It has been shown that in the non-counterfactual setting, for the kernel density estimator with fixed bandwidth the optimal rate of convergence is $\sqrt{n}$ with respect to the $L_2$ risk \citep[][Proposition 16]{KimSRW2019}. The minimax lower bounds for estimating generic kernel-smoothed parameters with fixed bandwidth have not been explicitly studied in the literature.



\section{Density Effect based on $L_1$-distance}
\label{sec:densiti-effects-estimator}

In this section, we discuss three different approaches to estimating the proposed density-based causal effect in \eqref{identify.exp::obs}, and analyze their asymptotic properties and error bounds.

\subsection{Smoothing Counterfactual Density} 
\label{subsec:density-effects-smooth-pa}

Given that we have proposed our counterfactual density estimator $\widehat{p}_{a,h}$ in \eqref{eqn:doubly-robust}, the following plug-in estimator arises naturally:
\begin{equation} \label{eqn:density-effect-estimator-plug-in}
\widehat{\psi}^{cd}_h = D(\widehat{p}_{1,h},\widehat{p}_{0,h}),
\end{equation}
where we simply compute the $L_1$ distance between two counterfactual density estimates obtained using $\widehat{p}_{a,h}$. To evaluate performance of the proposed plug-in estimator $\widehat{\psi}^{cd}_h$, we analyze the root-mean-square error (the square root of the $L_2$ risk) in the following theorem.

\begin{theorem} \label{thm:psi-sm-counterfactual-density}
Assume that the conditions \ref{assumption:A1} - \ref{assumption:A3} hold, and define
\begin{align*}
    R_{2,n}^\prime = n^{-\frac{1}{2}} \int \left\Vert \widehat{\mu}_{a,y} - {\mu}_{a,y} \right\Vert dy 
    + \left\Vert\widehat{\pi}_{a}-{\pi}_{a}\right\Vert \int \left\Vert\widehat{\mu}_{a,y}-{\mu}_{a,y}\right\Vert dy.
\end{align*}
Then we have
\begin{align*}
    \left\Vert \widehat{\psi}^{cd}_h  - \psi \right\Vert &\lesssim n^{-\frac{1}{2}} h^{-\frac{d}{2}} + \max_a \int \left\vert p_{h,a}(y) - p_{a}(y) \right\vert dy + R_{2,n}^\prime.
\end{align*}
\end{theorem}

As in the previous section, in order to analyze the integrated bias term $\int \left\vert p_{h,a}(y) - p_{a}(y) \right\vert dy$ we need to rely on a restricted subset of densities. Since our density effects are defined with respect to the $L_1(\R)$-norm, we may assume that $p_a$ belongs to the class of densities corresponding to the Nikol’ski class $\mathcal{H}(1,\beta,L)$. In the next corollary, we give the upper bounds of the $L_2$ risk for both varying and fixed bandwidths.

\begin{corollary}
\label{cor:psi-sm-counterfactual-density}
Suppose that $p_a \in \mathcal{P}_\mathcal{H}(1,\beta,L)$ and $K$ is a kernel of order $\floor{\beta}{}$ satisfying $\int \Vert u \Vert_1^\beta \vert K(u) \vert du < \infty$. If we further assume that $\Vert K \Vert_2 < \infty$ and Assumptions \ref{assumption:A2}, \ref{assumption:A3} hold, then taking $h \asymp n^{-\frac{1}{2\beta + d}}$, we have
\begin{align*}
\left\Vert \widehat{\psi}^{cd}_h  - \psi \right\Vert &\lesssim  n^{-\frac{\beta}{2\beta + d}} + R_{2,n}^\prime.
\end{align*}
If we are rather interested in ${\psi}^{cd}_h \equiv D({p}_{1,h},{p}_{0,h})$ with a fixed bandwidth $h$, then, under the conditions \ref{assumption:A2},\ref{assumption:A3}, and that $\Vert K \Vert_2 < \infty$ we get
\begin{align*}
\left\Vert \widehat{\psi}^{cd}_h-{\psi}^{cd}_h \right\Vert\lesssim n^{-\frac{1}{2}} + R_{2,n}^\prime.
\end{align*}
\end{corollary}

$\widehat{\psi}^{cd}_h$ is simple and easy to interpret, and conjectured to be optimal up to a logarithmic factor, as noted in the following remark.

\begin{remark}
In the non-counterfactual setting, the minimax lower bounds for the $L_1$-norm have been studied in nonparametric regression \citep{lepski1999estimation} and in Gaussian white noise models \citep{han2020estimation}, but not in density estimation to our knowledge. This makes the discussion on minimax optimality of $\widehat{\psi}^{cd}_h$ more difficult than $\widehat{p}_{a,h}$. We conjecture that the lower bounds for the $L_1$-norm in counterfactual density estimation are also on the order of $(n\log n)^{-\frac{\beta}{2\beta + d}}$, as found in those works. In our forthcoming paper, we plan to pursue this in greater depth.
\end{remark}

Since the $L_1$ distance $D(\cdot, \cdot)$ is not Hadamard differentiable, we cannot apply the delta method to \eqref{eqn:CLT-density-pointwise} to derive a limiting distribution of the proposed plug-in estimator $\widehat{\psi}^{cd}_h$. Alternatively, we obtain the weak convergence of $\widehat{\psi}^{cd}_h$ in Section \ref{subsec:confidence_band_for_density} and utilize this result to analyze the asymptotic properties of $\widehat{\psi}^{cd}_h$. We will return to this topic in section \ref{subsec:inference-density-effect}.

\subsection{Smoothing $L_1$ Distance} \label{subsec:smoothed-L1-distance}

Next, we develop our density effect estimator by smoothing the $L_1$ distance function, not the counterfactual densities themselves. 
This approach is delineated in \citet[][]{kennedy2021semiparametric}. For $t \in \R$, let $h_s(t)$ be an approximation of the absolute value function $\vert t \vert$, with a smoothing parameter $s$ controlling the approximation error of order $g(s)$: i.e.,
\[
\left\vert h_s(t) - \vert t \vert \right\vert \lesssim g(s).
\]
We assume that the function $h_s$ is twice continuously differentiable and has bounded second derivatives. Simple examples include $\sqrt{t^2+s^2}$, $t\tanh(ts)$, and $t\text{erf}(ts)$. There also exist polynomial and rational approximations $h_s(t)$ of degree $s$ , giving $g(s) = s^{-1}$ and $\exp(-s)$, respectively \citep{newman1964rational}. We remain agnostic about the choice of formula for $h_s$, as each option presents distinct advantages and disadvantages. Based on $h_s$, one may obtain the smoothed $L_1$ distance between densities $p,q$ as $\int h_s(p-q)$. We define the smoothed approximation of our target effect $\psi$ by
\[
    \psi^{sm}_s \coloneqq \int h_s\left(p_1(y)-p_0(y)\right) dy.
\]

Let $\widehat{\nu}_a(y\mid x)$ an estimator for the conditional density function $\nu_a(y\mid x)$.
We also let $\widehat{p}_a$ be an estimator of the counterfactual density under $A=a$. The default choice could be the simple plug-in estimator $\widehat{p}_a(y) = \Pn\left\{\widehat{\nu}_a(y\mid X) \right\}$; yet one may consider other estimators as well (see Remark \ref{rmk:other-estimator-for-hat-p_a}). By Lemma 1 in \citet{kennedy2021semiparametric}, the efficient influence function of $\psi^{sm}_s$ is derived as follows:
\begin{equation} \label{eqn:EIF-psi-sm}
    \begin{aligned}
        \phi^{sm}_s(Z;\eta^{sm}) &= \frac{(2A-1)}{\pi_A(X)}\left\{h'_s(p_1(Y)-p_0(Y)) - \int h'_s(p_1(y)-p_0(y))\nu_A(y \mid X)dy\right\} \\
        & \quad + \int h'_s(p_1(y)-p_0(y))\left\{\nu_1(y \mid X) - \nu_0(y \mid X) \right\}dy \\
        & \quad - \int\int h'_s(p_1(y)-p_0(y))\left\{\nu_1(y \mid x) - \nu_0(y \mid x) \right\}dy d\Pb(x),
    \end{aligned}
\end{equation}
where $\eta^{sm}=\{\pi_a, \nu_a, p_a\}$ denotes the relevant nuisance functions. Based on \eqref{eqn:EIF-psi-sm}, we propose the following bias-corrected estimator for $\psi^{sm}_s$:
\begin{align} \label{eqn:estimator-psi-sm}
    \widehat{\psi}^{sm}_s = \int h_s\left(\widehat{p}_1(y)-\widehat{p}_0(y)\right) dy + \Pn\left\{ \phi^{sm}_s(Z;\widehat{\eta}^{sm}) \right\}.
\end{align}
We proceed by introducing an additional consistency assumption on the nuisance estimators, which is necessary for formally stating the main theorem in this subsection. In the rest of the paper, with a slight abuse of notation, we define $\Vert \widehat{\nu}_a - \nu_a \Vert^2=\int \zeta_a(x)^2 d\Pb(x)$, where $\zeta_a(x)=\int\left\vert \widehat{\nu}_a(y \mid x) - \nu_a(y \mid x) \right\vert dy$.

\begin{itemize}[leftmargin=*]
	\item \customlabel{assumption:A5}{(A5)} \textit{Consistency on} $\widehat{\pi}_a, \widehat{\nu}_a$: 
    \[
    \Vert h'_s\Vert_{\infty} \max_a\left\{ \left\Vert\frac{1}{\widehat{\pi}_a}\right\Vert_{\infty} \right\} \left( \sum_a \left\{\Vert\widehat{\pi}_a - \pi_a \Vert + \Vert \widehat{\nu}_a - \nu_a \Vert \right\} \right) = o_{\Pb}(1).
    \]    
\end{itemize}
$\Vert h'_s\Vert_{\infty}$ is finite for the vast majority of options for $h_s$. Hence, Assumption \ref{assumption:A5} typically reduces to a mild consistency assumption on $\widehat{\pi}_a, \widehat{\nu}_a$ with no requirement on rates of convergence. 
In the next theorem, we describe an asymptotic behavior of the proposed estimator $\widehat{\psi}^{sm}_s$ and compute the $L_2$-norm error with respect to $\psi$. 

\begin{theorem}\label{thm:error-bound-psi-sm}
Suppose that Assumptions \ref{assumption:A2}, \ref{assumption:A3}, \ref{assumption:A5} hold. Then
\begin{align*}
    \widehat{\psi}^{sm}_s - \psi  & = (\Pn-\Pb)\phi^{sm}_s(Z;\eta^{sm}) + \psi^{sm}_s - \psi \\
    & \quad + O_{\Pb}\left( \sum_a \left\{ \Vert h'_s\Vert_{\infty} \Vert \widehat{\pi}_a - \pi_a \Vert \Vert \widehat{\nu}_a - \nu_a \Vert + \Vert h''_s\Vert_{\infty}\Vert \widehat{p}_a - p_a \Vert^2 \right\}  \right) + o_{\Pb}\left(\frac{1}{\sqrt{n}}\right),
\end{align*}
and
\begin{align*}
    \left\Vert  \widehat{\psi}^{sm}_s - \psi \right\Vert &= O_\Pb\Bigg(\left\vert \psi^{sm}_s - \psi \right\vert + \frac{\Vert h'_s \Vert_\infty}{\sqrt{n}} + \sum_a\left\{ \Vert h'_s\Vert_{\infty}\Vert \widehat{\pi}_a - \pi_a \Vert \Vert \widehat{\nu}_a - \nu_a \Vert + \Vert h''_s\Vert_{\infty}\Vert\widehat{p}_a - p_a \Vert^2 \right\} \\
    & \qquad \quad + \Vert h''_s\Vert_{\infty}\Vert\widehat{p}_0 - p_0 \Vert \Vert\widehat{p}_1 - p_1 \Vert \Bigg) + o_\Pb\left( \frac{1}{\sqrt{n}} \right).
\end{align*}
\end{theorem}

If $\text{Area}(\mathcal{Y}) = \int_\mathcal{Y}dy < \infty$, then one may get $\vert \psi^{sm}_s  - \psi \vert \lesssim g(s)$. In this case, the $L_2$-norm error is minimized with the following choice of $s$:
\begin{align*}
    s^* = \argmin_s \left\{g(s) + \Vert h'_s \Vert_\infty \max\left\{\frac{1}{\sqrt{n}},r_{1,n}\right\} + \Vert h''_s \Vert_\infty r_{2,n} \right\},
\end{align*}
where
\begin{align*}
    & \Vert \widehat{\pi}_a - \pi_a \Vert \Vert \widehat{\nu}_a - \nu_a \Vert = O_\Pb(r_{1,n}), \\
    & \Vert\widehat{p}_0 - p_0 \Vert \Vert\widehat{p}_1 - p_1 \Vert + \max_a\Vert\widehat{p}_a - p_a \Vert^2 = O_\Pb(r_{2,n}).
\end{align*}
Theorem \ref{thm:error-bound-psi-sm} shows that the estimation error consists of the terms that are second-order in the nuisance estimation error and the smoothing approximation error $\vert \psi^{sm}_s - \psi \vert$. For fixed $s$, $\widehat{\psi}^{sm}_s$ is asymptotically normal and efficient with respect to $\psi^{sm}_s$ under weak nonparametric conditions, such that $\Vert \widehat{\pi}_a - \pi_a \Vert \Vert \widehat{\nu}_a - \nu_a \Vert + \Vert \widehat{p}_a - p_a \Vert^2 = o_\Pb(1/\sqrt{n}), \forall a \in \mathcal{A}$. In this case, we have the non-vanishing asymptotic bias $\vert \psi^{sm}_s - \psi \vert$. For varying $s$, $\widehat{\psi}^{sm}_s \overset{p}{\longrightarrow} \psi$ can be achieved under relatively very mild conditions such that $g(s) + \max_a \left\{ \Vert h'_s\Vert_{\infty} \Vert \widehat{\pi}_a - \pi_a \Vert \Vert \widehat{\nu}_a - \nu_a \Vert + \Vert h''_s\Vert_{\infty}\Vert \widehat{p}_a - p_a \Vert^2 \right\}=o_\Pb(1)$ provided that $\mathcal{Y}$ is bounded. Achieving particular rates of convergence, however, can be more challenging than the case of $\widehat{\psi}^{cd}_h$ in the previous subsection; stronger rate conditions for the nuisance estimators may be required since the terms like $\Vert h''_s \Vert_\infty$ typically grow to infinity as $s \rightarrow 0$. In general, $\sqrt{n}$-consistency is unattainable for $\widehat{\psi}^{sm}_s$. For inference, one may use a conservative approach by constructing a valid confidence interval for $\psi^{sm}_s$ then increasing the radius by $\vert \psi^{sm}_s - \psi \vert$. We shall go into more detail about the inferential procedure in Section \ref{subsec:inference-density-effect}. 

Importantly, although $h_s$ is not exactly the $L_1$ distance we care about,
if the absolute value function is upper- or lower-bounded by $h_s$ on entire $\R$ (e.g., $t\tanh(ts) \leq \vert t \vert \leq \sqrt{t^2+s^2}$), it could provide a way to interpret our smooth approximation in some sense; for example, when $h_s$ approximates the absolute value function from below, one may simply use $\psi^{sm}_s$ to test the null hypothesis $p_0 = p_1$ from a more conservative standpoint. Moreover, we may not necessarily care about the exact $L_1$ distance, and any reasonable approximation will suffice. In this case, $\widehat{\psi}^{sm}_s$ enables to achieve the fast rate of convergence to the desired level of the smooth approximation $\psi^{sm}_s$.

The efficient influence function in \eqref{eqn:EIF-psi-sm} reduces to zero when $p_1 = p_0$, which presents some complications for inference. We will also revisit this issue in Section \ref{subsec:inference-density-effect}.

\begin{remark} \label{rmk:other-estimator-for-hat-p_a}
We have suggested using the plug-in estimator $\widehat{p}_a(y) = \Pn\left\{\widehat{\nu}_a(y\mid X) \right\}$ of the counterfactual density, due to the fact that the result in Theorem \ref{thm:error-bound-psi-sm} only depends on the second-order error $\Vert\widehat{p}_a - p_a \Vert^2$ and that rates for estimating $p_a$ will not be slower than rates for estimating $\nu_a$ \citep[see][Remark 4]{kennedy2021semiparametric}. However, in principle, the results in this section could apply to other estimators as well; for instance, one could use the proposed doubly robust estimator in \eqref{eqn:doubly-robust}, which may offer faster convergence rates than the plug-in estimator.
\end{remark}

\subsection{Under Margin Condition} \label{subsec:margin-condition}
Note that one may rewrite our target effect as
\begin{align} \label{eqn:target-effect-ind-functions}
    \psi = \int \left( p_1(y) - p_0(y) \right) \mathbbm{1}\left\{p_1(y) > p_0(y) \right\} dy + \int \left( p_0(y) - p_1(y) \right) \mathbbm{1}\left\{p_0(y) > p_1(y) \right\}  dy.
\end{align}

Unlike in the previous approaches, here we do not use smoothing methods but instead make an assumption to control the behavior of $p_0, p_1$ around the threshold $p_0 = p_1$, so that each component in \eqref{eqn:target-effect-ind-functions} behaves as a smooth functional. Specifically, we require the following margin condition:

\begin{definition}[Margin Condition] \label{assumption:MC}
For some $\alpha > 0$ and for all $t$, we have that
\[
    \Pb(\vert p_1 - p_0  \vert \leq t) \lesssim t^{\alpha}.
\]
\end{definition}
The above margin condition restricts the probability that the two counterfactual density functions get too close to each other. This is analogous to that used in classification \citep{audibert2007fast}, clustering \citep{levrard2018quantization}, optimal treatment regime \citep{luedtke2016optimal, luedtke2016statistical, pmlr-v202-kim23ab}, as well as other problems involving estimation of non-smooth parameters such as \citet{kennedy2019survivor, kennedy2020sharp, levis2023covariate}.

We are interested in estimating the parameter of the form
\begin{align*}
    \int \gamma(y) \left(p_1(y) - p_0(y)\right) dy &= \int \int \gamma(y) \left\{\nu_1(y \mid X) - \nu_0(y \mid X) \right\} d\Pb(x) dy \\
    &= \E\left\{\E\left(\gamma(Y) \mid X, A=1 \right)-\E\left(\gamma(Y) \mid X, A=0 \right)\right\},
\end{align*}
for some deterministic function $\gamma : \mathcal{Y} \rightarrow \R$. The uncentered influence function for the above parameter if the function $\gamma$ were known would be
\begin{equation} \label{eqn:etf-mc-component}
    \begin{aligned}
        \underline{\varphi}(Z;\gamma,\eta^{mc}) = \frac{(2A-1)}{\pi_A(X)}\left\{\gamma(Y) - \int \gamma(y)\nu_A(y \mid X)dy\right\} 
        + \int \gamma(y)\left\{\nu_1(y \mid X) - \nu_0(y \mid X) \right\}dy,
    \end{aligned}
\end{equation}
where $\eta^{mc} = \{\pi_a, \nu_a, p_a\}$. This motivates the estimator $\Pn\left\{\underline{\varphi}(Z;\widehat{\gamma}, \widehat{\eta}^{mc}) \right\}$, where $\widehat{\gamma}$ is obtained through plug-in principles. 


Now, for any $y \in \mathcal{Y}$, define the indicator functions
\begin{align*}
    \gamma^{+}(y) = \mathbbm{1}\left\{p_1(y) > p_0(y) \right\} \quad \text{and} \quad \gamma^{-}(y) = \mathbbm{1}\left\{p_0(y) > p_1(y) \right\},
\end{align*}
and let
\begin{align*}
    \widehat{\gamma}^{+}(y) = \mathbbm{1}\left\{\widehat{p}_1(y) > \widehat{p}_0(y) \right\} \quad \text{and} \quad \widehat{\gamma}^{-}(y) = \mathbbm{1}\left\{\widehat{p}_0(y) > \widehat{p}_1(y) \right\}
\end{align*}
denote their estimates, respectively. Then we let
\begin{align*}
    \varphi^{mc}(Z;\gamma^{mc},\eta^{mc}) = \underline{\varphi}(Z;\gamma^{+},\eta^{mc}) + \underline{\varphi}(Z;\gamma^{-},\eta^{mc}),
\end{align*}
which is the uncentered influence function for $\psi$ under the margin condition,
where $\gamma^{mc}=\{\gamma^{+}, \gamma^{-}\}$.
Therefore, our proposed estimator for $\psi$ is
\begin{align} \label{eqn:estimator-mc}
    \widehat{\psi}^{mc} = \Pn\left\{\varphi^{mc}(Z;\widehat{\gamma}^{mc}, \widehat{\eta}^{mc}) \right\},
\end{align}
with $\widehat{\gamma}^{mc}=\{\widehat{\gamma}^{+}, \widehat{\gamma}^{-}\}$. In contrast to the previous two approaches, the estimator in \eqref{eqn:estimator-mc} does not require a smoothing parameter. In what follows, we describe additional regularity conditions required to state our main result in this subsection.

\begin{itemize}[leftmargin=*]
	\item \customlabel{assumption:A6}{(A6)} \textit{Consistency on} $\widehat{\pi}_a, \widehat{\nu}_a, \widehat{p}_a$: 
	\[
    \max_a \left\{ \Vert\widehat{\pi}_a - \pi_a \Vert + \Vert \widehat{\nu}_a - \nu_a \Vert + \Vert \widehat{p}_a - p_a \Vert_{\infty} \right\} = o_{\Pb}(1).
    \]
\end{itemize}

In the next theorem, we characterize the asymptotic properties and the $L_2$-norm error of the proposed estimator.

\begin{theorem}\label{thm:error-bound-psi-mc}
Suppose that Assumptions \ref{assumption:A2}, \ref{assumption:A3}, \ref{assumption:A6} hold, and that the margin condition holds for some $\alpha$. Further assume that $Y$ has the density function $p$ such that $\inf_{y \in \mathcal{Y}} p(y) \geq p_{\min} > 0$, and that $\underset{y \in \mathcal{Y}}{\sup}\Vert {\nu}_a(y \mid X) \Vert < \infty$. Then,
\begin{align*}
    \widehat{\psi}^{mc} - \psi  & = (\Pn-\Pb)\varphi^{mc}(Z;\gamma^{mc},\eta^{mc}) \\
    & \quad + O_{\Pb}\left( \sum_a \left\{ \Vert \widehat{\pi}_a - \pi_a \Vert \Vert \widehat{\nu}_a - \nu_a \Vert + \left(\Vert \widehat{p}_a - p_a \Vert_{\infty} \right)^{\alpha+1} \right\}  \right) + o_{\Pb}\left(\frac{1}{\sqrt{n}}\right),
\end{align*}
and 
\begin{align*}
    \left\Vert \widehat{\psi}^{mc} - \psi \right\Vert = O_{\Pb}\left( \sum_a \left\{ \Vert \widehat{\pi}_a - \pi_a \Vert \Vert \widehat{\nu}_a - \nu_a \Vert + \left(\Vert \widehat{p}_a - p_a \Vert_{\infty} \right)^{\alpha+1} \right\} +  \frac{1}{\sqrt{n}} \right).
\end{align*}
\end{theorem}

Theorem \ref{thm:error-bound-psi-mc} shows that the proposed estimator is consistent, with the rate of convergence that is second-order in nuisance estimation errors. If the second-order nuisance errors converge to zero at a faster than $\sqrt{n}$ rates, $\widehat{\psi}^{mc}$ is asymptotically normal, and efficient. This condition on the nuisance estimation could be satisfied under weak nonparametric conditions; for example, if $\alpha = 1$ and the nuisance estimators converge at faster than $n^{1/4}$ rates \citep[e.g.,][]{kennedy2020sharp}. It also shows that the $L_2$-norm error bounds attain $\sqrt{n}$ rates as long as $\sum_a \Vert \widehat{\pi}_a - \pi_a \Vert \Vert \widehat{\nu}_a - \nu_a \Vert + \left(\Vert \widehat{p}_a - p_a \Vert_{\infty} \right)^{\alpha+1} = O_\Pb(1/\sqrt{n})$. Note that $\sqrt{n}$ rates are not attainable with the previous approaches proposed in Sections \ref{subsec:density-effects-smooth-pa} and \ref{subsec:smoothed-L1-distance}.

\begin{remark}
    Using other versions of the margin condition, one can avoid imposing the lower bound condition on the density function $p$ in Theorem \ref{thm:error-bound-psi-mc}. For instance, one could restrict the area directly, rather than the probability as in Definition \ref{assumption:MC}, as follows:
    \[
        \int_{y \in \mathcal{Y}} \mathbbm{1}\left\{ \lvert p_1(y) - p_0(y) \rvert \leq t \right\} dy \lesssim t^\alpha.
    \]    
\end{remark}


\section{Inference}
\label{sec:confidence_interval}

In this section, we present a bootstrap approach to obtaining pointwise and simultaneous confidence bands for our target parameters. Specifically, we detail the asymptotic property of the proposed counterfactual density estimator \eqref{eqn:doubly-robust} uniformly across a continuum of $y \in \mathcal{Y}$, and propose a bootstrap algorithm for obtaining uniform confidence bands. We go on to present bootstrap algorithms for constructing confidence intervals for the density effect \eqref{identify.exp::obs} based on each of the proposed estimators in Section \ref{sec:densiti-effects-estimator}, and develop a novel test of no distributional effect.

\subsection{Uniform Inference for Counterfactual Density}\label{subsec:confidence_band_for_density}

As to uniform inference for the proposed counterfactual density estimator, we use the fixed-bandwidth approach as advocated by many previous studies \citep[e.g.,][]{wasserman2006all,Chen17,rinaldo2010, ChaFasLec13}. Unlike the case where $h$ is variable (or adaptive), we focus on the smoothed density $p_{a,h}$ with a fixed $h$ and use a confidence set centered at $\widehat{p}_{a,h}$. This introduces a non-vanishing bias, as well as an undercoverage issue, meaning that the usual $(1-\alpha)$ confidence set for $\widehat{p}_{a,h}$ will not asymptotically cover $p_a$ with probability tending to $1-\alpha$. 

Nonetheless, there are several crucial benefits of targeting $p_{a,h}$ instead of $p_a$. 
With variable $h$, it is not possible for $\widehat{p}_{a,h}$ to attain $\sqrt{n}$ rates under nonparametric conditions, as is often the case when the target parameter is a curve \citep[e.g.,][]{kennedy2017nonparametric, kennedy2019robust}. A class of the kernel functions with a fixed bandwidth $\{K_{h,y}:y\in\mathbb{R}^{d}\}$ exhibits uniform regularity. As the bandwidth shrinks, the kernel function becomes sharper and more singular. Thus, maintaining a fixed bandwidth ensures a minimal regularity, characterized by a bounded VC dimension, that is uniformly shared across all kernel functions in the class. This allows $\widehat{p}_{a,h}$ to achieve fast parametric $\sqrt{n}$ rates and the desired weak convergence under relatively mild conditions. Moreover, by changing our target to $p_{a,h}$, we do not need strong smoothness assumptions on the density to analyze the bias $p_{a,h} - p_a$; the kernel-smoothed density may exist even if $Y^a$ itself does not have a density in the usual sense \citep{rinaldo2010}. Fixed bandwidths can also better reflect practical data analysis, since we typically face a single dataset with a particular sample size, rather than a sequence of datasets of increasing size. Last but not least, with varying bandwidth, one may require impractical undersmoothing where we sacrifice risk for coverage. Hence, despite being off-centered, an accurate confidence set with $p_{a,h}$ is potentially more useful than a poor confidence set with $p_a$. However, we highlight that our proposed inferential procedures can also be used for hypothesis testing concerning the true counterfactual densities, which we will discuss in greater detail in Section \ref{subsubsec:testing-no-effect}.

\begin{remark}\label{remark:bandwidth} (\textit{Bandwidth selection}) 
In the fixed bandwidth analysis, the bandwidth $h$ must be specified in advance through a separate procedure. Here we list a few potential options to tune $h$ in our setting. First, suppose that we have a priori knowledge about underlying outcomes, such as that units whose outcome values are not too far apart are roughly similar in some ways. Then we could potentially find a good cutoff value to determine which units should be treated as a neighborhood, and use this to find a proper $h$. A second natural option could be Silverman's rule of thumb as in \cite{Chen17}, in hopes of the underlying density close to being normal, where we simply use a sample standard deviation of observed outcome. Third, we could proceed based on a more data-driven approach. We split our data into two and estimate the density with a sequence of different $h$'s on one and estimate a corresponding pseudo-risk on the other. Then we can pick $h$ that minimizes the estimated pseudo-risk. This cross-validation-like approach to parameter/model selection in causal inference has also appeared in previous work of \citet{vanDudoit2003unified, kennedy2017nonparametric, kennedy2019robust, kennedy2021semiparametric}, for example. 
\end{remark}

Now, we detail the main large-sample property of $\widehat{p}_{a,h}$ given a fixed $h$.
In contrast to the previous result \eqref{eqn:CLT-density-pointwise} that only holds pointwise for a given $y$, the subsequent result hold uniformly in the sense that, when viewed as a random function on a compact set $\mathcal{Y}$, $\widehat{p}_{a,h}$ converges in distribution to a Gaussian process. To this end, we require an additional assumption as follows.

\begin{itemize}[leftmargin=*]
\item \customlabel{assumption:kernel_vc}{($\text{B1}$)} \textit{Bounded VC dimension}: For the kernel function $K$ such that $\Vert K \Vert_\infty < \infty$, given $h>0$, $\mathcal{G}_{h}:=\left\{ K_{h,y}:\,y\in\mathbb{R}^{d}\right\} $
is a uniformly bounded VC-class with dimension $\nu$; i.e., there
exist positive numbers $\Updelta$ and $\nu$ such that, for every probability
measure $Q$ on $\mathbb{R}^{d}$ and for every $\epsilon\in(0,1)$,
the covering numbers $\mathcal{N}\left(\mathcal{G}_{h},L_{2}(Q),\epsilon h^{-d}\Vert K \Vert_{\infty}\right)$
satisfy 
\[
\mathcal{N}\left(\mathcal{G}_{h},L_{2}(Q),\epsilon h^{-d}\Vert K \Vert_{\infty}\right)\leq\left(\frac{\Vert K \Vert_{\infty}\Updelta}{\epsilon h^{d}}\right)^{\nu},
\]
where the covering numbers is the minimal number of open balls of
radius $\epsilon$ with respect to $L_{2}(Q)$ distance whose centers
are in $\mathcal{G}{}_{h}$ to cover $\mathcal{G}{}_{h}$.
\end{itemize}

One sufficient condition for Assumption \ref{assumption:kernel_vc} is the bounded domain and Lipschitz kernel as described in the following lemma.

\begin{lemma}[Lemma 14 in \citet{KimSRW2019}]
\label{lem:kde_vc}
Suppose $Y$ has a bounded support, i.e., there exists $R_{\mathcal{Y}}<\infty$
with $\mathcal{Y}\subset\mathbb{B}_{\mathbb{R}^{d}}(0,R_{\mathcal{Y}})$.
Also suppose the kernel $K$ is $M_{K}$-Lipschitz, i.e., for all $u_{1},u_{2}\in\mathbb{R}^{d}$,
$|K(u_{1})-K(u_{2})|\leq L_{K}\|u_{1}-u_{2}\|$. Then for all $\epsilon\in\left(0,\Vert K \Vert_{\infty}\right)$,
the supremum of the $\epsilon$-covering number $\mathcal{N}(\mathcal{G}_{h},L_{2}(Q),\epsilon)$
over all measure $Q$ is upper bounded as 
\[
\sup_{Q}\mathcal{N}(\mathcal{G}_{h},L_{2}(Q),\epsilon)\leq\left(\frac{\Vert K \Vert_{\infty}+2R_{\mathcal{Y}}M_{K}}{\epsilon}\right)^{d}.
\]
\end{lemma}


In what follows, we introduce a slightly stronger version of Assumption \ref{assumption:A4}.
\begin{itemize}[leftmargin=*]
    \item \customlabel{assumption:A4'}{(A4$'$)} \textit{Nonparametric conditions on $\widehat{\pi}_{a}, \widehat{\mu}_{a,y}$}: 
    \begin{align*}
    &\underset{a\in\mathcal{A}}{\sup}\left\Vert \widehat{\pi}_{a}-\pi_{a}\right\Vert _{\infty}\sqrt{\log\left(1/\underset{a\in\mathcal{A}}{\sup}\left\Vert \widehat{\pi}_{a}-\pi_{a}\right\Vert _{\infty}\right)}\to 0 \quad \text{a.s.},\\
    &\underset{a\in\mathcal{A}}{\sup}\left\Vert \widehat{\mu}_{a,y}-\mu_{a,y}\right\Vert _{\infty}\sqrt{\log\left(1/\underset{a\in\mathcal{A}}{\sup}\left\Vert \widehat{\mu}_{a,y}-\mu_{a,y}\right\Vert _{\infty}\right)} \to 0 \quad \text{a.s.}, \\
    & \underset{y\in \mathcal{Y}, a\in\mathcal{A}}{\sup}\left\Vert \widehat{\mu}_{a,y}-{\mu}_{a,y}\right\Vert \left\Vert \widehat{\pi}_{a}-{\pi}_{a}\right\Vert =o_{\Pb}(1/\sqrt{n}).
    \end{align*}
\end{itemize}

The next result lays the foundation for our inferential procedures.

\begin{theorem} \label{thm:asymptotic_smooth_density} 
Assume that Assumptions \ref{assumption:A2}, \ref{assumption:A3}, \ref{assumption:A4'} and \ref{assumption:kernel_vc} hold. When $\sqrt{n}\left(\hat{p}_{h}-p_{h}\right)$ is understood as a stochastic process indexed by $(a,y)\in\mathcal{A}\times\mathbb{R}^{d}$ such that $\left\{ \sqrt{n}\left(\hat{p}_{h}-p_{h}\right)\right\} _{(a,y)}=\sqrt{n}\left(\widehat{p}_{a,h}(y)-p_{a,h}(y)\right)$, we have the following weak convergence as 
\[
\sqrt{n}\left(\hat{p}_{h}-p_{h}\right)\to\mathbb{G}\text{ weakly in }\ell^{\infty}(\mathcal{A}\times\mathbb{R}^{d})\text{ a.s.},
\]
where $\mathbb{G}$ is a centered Gaussian process with $Cov\left[\mathbb{G}(a_{1},y_{1}),\mathbb{G}(a,y_{2})\right]=\mathbb{E}\left[f_{y_{1}}^{a_{1}}f_{y_{2}}^{a_{2}}\right]-\mathbb{E}\left[f_{y_{1}}^{a_{1}}\right]\mathbb{E}\left[f_{y_{2}}^{a_{2}}\right]$, $f_{y}^{a} = \frac{\mathbf{1}(A=a)}{\pi_{a}(X)}\big(K_{h,y}(Y)-\mu_{A,y}(X)\big)+\mu_{a,y}(X)$.
\end{theorem}

Based on Theorem \ref{thm:asymptotic_smooth_density}, in what follows we present a bootstrap approach to obtaining uniform confidence bands for $p_{a,h}=\{p_{a,h}(y) : y \in \mathcal{Y} \}$. Throughout this section, we apply a sample-splitting procedure for each bootstrap algorithm, where nuisance estimators are constructed on one portion of the sample, while the remaining portion is reserved exclusively for bootstrapping.






\begin{algorithm}\mbox{}\\[-\baselineskip]
\begin{enumerate} \label{algorithm:density-band}
\item We generate $B$ bootstrap samples $\{\tilde{Z}_{1}^{1},\ldots,\tilde{Z}_{n}^{1}\},\ldots,\{\tilde{Z}_{1}^{B},\ldots,\tilde{Z}_{1}^{B}\}$,
by sampling with replacement from the original sample. 
\item On each bootstrap sample, compute $T_{i}^{a}=\sqrt{n}\left\Vert \widehat{p}_{a,h}^{i}-\widehat{p}_{a,h}\right\Vert _{\infty}$,
where $\widehat{p}_{a,h}^{i}$ is the proposed kernel density estimator $\widehat{p}_{a,h}$ computed on the $i$-th
bootstrap sample $\{\tilde{Z}_{1}^{i},\ldots,\tilde{Z}_{n}^{i}\}$. 
\item Compute $\alpha$-quantile $\hat{z}_{\alpha}^{a}=\inf\left\{ z:\,\frac{1}{B}\sum_{i=1}^{B}I(T_{i}^{a}>z)\leq\alpha\right\} $. 
\item Define $\widehat{C}_{\alpha}=\left\{ q:\left\Vert \widehat{p}_{a,h}-q\right\Vert _{\infty}\leq\hat{z}_{\alpha}^{a}/\sqrt{n}\right\} $. 
\end{enumerate}
\end{algorithm}

Algorithm \ref{algorithm:density-band} is easy to implement in practice, and provides a $(1-\alpha)$ uniform confidence band $\widehat{C}_{\alpha}$ of the form $\widehat{p}_{a,h}(y) \pm \hat{z}_{\alpha}^{a}/\sqrt{n}$.

The validity of the proposed bootstrap algorithm is based on the stochastic convergence of the empirical process, as briefly described in Section \ref{subsec:bootstrap-validity}. Suppose we have an original i.i.d sample set $(Z_{1},...,Z_{n})\sim\mathbb{P}$ and a bootstrapped set $(Z_{1}^{*},\ldots,Z_{n}^{*})$, and their empirical measures $\mathbb{P}_{n}$, $\mathbb{P}_{n}^{*}$ respectively. The main theory that underpins our bootstrapping algorithm is that the empirical process and its bootstrapped version converge to the same limiting distribution (Theorem \ref{thm:weakconv-bootstrap}).
In our case, it suffices to prove that $\sqrt{n}\left(\widehat{p}_{a,h}-p_{a,h}\right)$ and $\sqrt{n}\left(\widehat{p}_{a,h}^{*}-p_{a,h}\right)$ converge to the same limiting distribution for a fixed $h$. Convergence of $\sqrt{n}\left(\widehat{p}_{a,h}-p_{a,h}\right)$ across $y$ can be obtained from Theorem \ref{thm:asymptotic_smooth_density}, and applying Theorem \ref{thm:weakconv-bootstrap} implies that indeed $\sqrt{n}\left(\widehat{p}_{a,h}^{*}-p_{a,h}\right)$ converges to the same limiting distribution. The result is summarized in the following corollary.

\begin{corollary}
\label{cor:confidence_smooth_density}
Under the same conditions of Theorem~\ref{thm:asymptotic_smooth_density},
$\widehat{C}_{\alpha}$ constructed via Algorithm \ref{algorithm:density-band} are asymptotically valid uniform confidence bands, i.e. 
\[
\liminf_{n\to\infty}\mathbb{P}\left(p_{a,h}\in\widehat{C}_{\alpha}\right)\geq1-\alpha
\]
for a given level of $\alpha$. 
\end{corollary}


\subsection{Inference for Density Effect and Testing No Effect} \label{subsec:inference-density-effect}

In Section \ref{sec:densiti-effects-estimator}, we discussed the three approaches to estimating the  density-based effect based on the $L_1$ distance as defined in \eqref{identify.exp::obs}. In this section, we establish the large-sample properties for each of the proposed estimators, which form the basis for our inferential and testing procedures. Then we present bootstrap procedures to obtaining valid confidence intervals, and discuss a novel test of no distributional effect.

\subsubsection{Estimator based on smoothed counterfactual density} \label{subsubsec:inference-density-effect-cd}

Here, we discuss statistical inference for density effect estimation using the smoothed counterfactual density approach. As in Section \ref{subsec:confidence_band_for_density}, we adopt the fixed-bandwidth approach for $\widehat{p}_{a,h}$. 

\begin{remark}
    From the perspectives of geometry and topology, $p_h$ serves as a simplified yet robust representation that preserves the essential topological properties of $p$ \citep[Section 4.4]{FasyLWBS2014}. Hence, measuring $L_{1}$ distance between $p_{0,h}$ and $p_{1,h}$ could be understood as an attempt to focus on the essential difference between $p_{0}$ and $p_{1}$.
\end{remark}

Motivated by the triangle inequality:
\[
\left|D(\hat{p}_{1,h},\hat{p}_{0,h})-D(p_{1,h},p_{0,h})\right|\leq D(\hat{p}_{0,h},p_{0,h})+D(\hat{p}_{1,h},p_{1,h}),
\]
we describe the limiting distribution for $D\left(\widehat{p}_{a,h},p_{a,h}\right)$ in the next corollary.

\begin{corollary} \label{cor:asymptotic_l1_one}
Suppose that Assumptions \ref{assumption:A2}, \ref{assumption:A3}, and \ref{assumption:A4'} hold. Further assume that $\mathcal{Y}$ is bounded and $K$ is Lipschitz.  Then for a given $h$ and each $a\in\mathcal{A}$, we have
\[
\sqrt{n}D\left(\widehat{p}_{a,h},p_{a,h}\right)\to\int\left|\mathbb{G}(a,y)\right|dy\text{ weakly in }\mathbb{R}\text{ a.s.},
\]
where $\mathbb{G}$ is a centered Gaussian process with $Cov\left[\mathbb{G}(a_{1},y_{1}),\mathbb{G}(a_{2},y_{2})\right]=\mathbb{E}\left[f_{y_{1}}^{a_{1}}f_{y_{2}}^{a_{2}}\right]-\mathbb{E}\left[f_{y_{1}}^{a_{1}}\right]\mathbb{E}\left[f_{y_{2}}^{a_{2}}\right]$, $f_{y}^{a} = \frac{\mathbf{1}(A=a)}{\pi_{a}(X)}\big(K_{h,y}(Y)-\mu_{A,y}(X)\big)+\mu_{a,y}(X)$.
\end{corollary}

Corollary \ref{cor:asymptotic_l1_one} is deduced from Theorem \ref{thm:asymptotic_smooth_density}. We use the sufficient condition for Assumption \ref{assumption:kernel_vc}, which is described in Lemma \ref{lem:kde_vc}.

Given a target density effect $\theta$, we construct the confidence interval $\widehat{C}_{\alpha}$ centered at each proposed estimator $\widehat{\theta}$ with a width of $2c_{n}$, i.e., $
\widehat{C}_{\alpha}=[\widehat{\theta}-c_{n},\,\widehat{\theta}+c_{n}]
$. Then $\widehat{C}_{\alpha}$ is a valid $1-\alpha$ asymptotic confidence
set if and only if 
\begin{equation} \label{eqn:valid-CI}
\liminf_{n\to\infty}\mathbb{P}\left(\theta \in \widehat{C}_{\alpha}\right) = \underset{n\to\infty}{\lim\inf}\mathbb{P}(|\hat{\theta}-\theta|\leq c_{n})\geq1-\alpha.
\end{equation}

Based on Corollary \ref{cor:asymptotic_l1_one}, we propose the following bootstrap algorithm to construct a $(1-\alpha)$ confidence interval for $\psi^{cd}_h=D(p_{0,h},p_{1,h})$ for a given $h$ using our proposed estimator $\widehat{\psi}^{cd}_h$.

\begin{algorithm}\mbox{}\\[-\baselineskip]
\begin{enumerate} \label{algorithm:density-effect-cd1}
\item We generate $B$ bootstrap samples $\{\tilde{Z}_{1}^{1},\ldots,\tilde{Z}_{n}^{1}\},\ldots,\{\tilde{Z}_{1}^{B},\ldots,\tilde{Z}_{1}^{B}\}$,
by sampling with replacement from the original sample. 
\item On each bootstrap sample, compute $T_{i}^{a}=\sqrt{n}D(\widehat{p}_{a,h}^{i},\widehat{p}_{a,h})$,
where $\widehat{p}_{a,h}^{i}$ is the kernel density estimator $\widehat{p}_{a,h}$ computed on $i$-th
bootstrap samples $\{\tilde{Z}_{1}^{i},\ldots,\tilde{Z}_{n}^{i}\}$. 
\item Compute $\frac{\alpha}{2}$-quantile $\hat{z}_{\alpha/2}^{a}=\inf\left\{ z:\,\frac{1}{B}\sum_{i=1}^{B}I(T_{i}^{a}>z)\leq\frac{\alpha}{2}\right\} $. 
\item Define $\widehat{C}_{\alpha}=\left[D(\widehat{p}_{1,h},\widehat{p}_{0,h})-\frac{\hat{z}_{\alpha/2}^{0}}{\sqrt{n}}-\frac{\hat{z}_{\alpha/2}^{1}}{\sqrt{n}},\,D(\widehat{p}_{1,h},\widehat{p}_{0,h})+\frac{\hat{z}_{\alpha/2}^{0}}{\sqrt{n}}+\frac{\hat{z}_{\alpha/2}^{1}}{\sqrt{n}}\right]$. 
\end{enumerate}
\end{algorithm}

The interval obtained via Algorithm \ref{algorithm:density-effect-cd1} is an asymptotically valid $(1-\alpha)$ confidence interval for the target effect $\psi^{cd}_h$ under relatively weak conditions as shown in the following corollary.

\begin{corollary}
\label{cor:confidence_l1_one}
Assume that the same conditions of Corollary~\ref{cor:asymptotic_l1_one}
hold, and $\int\left|\mathbb{G}(a,y)\right|dy>0$.
Then, the intervals $\widehat{C}_{\alpha}$ constructed via Algorithm
\ref{algorithm:density-effect-cd1} are valid asymptotic confidence sets, i.e. 
\[
\liminf_{n\to\infty}\mathbb{P}\left(\psi^{cd}_h \in \widehat{C}_{\alpha}\right)\geq1-\alpha
\]
for given level of $\alpha$.
\end{corollary}

Even tighter confidence intervals for $\psi^{cd}_h$ can be obtained; however, these intervals are not valid when $p_{0,h} = p_{1,h}$. To achieve this, a slightly different asymptotic result is required, as demonstrated in the next corollary.

\begin{corollary} \label{cor:asymptotic_l1_twodiff}
Assume that
the same conditions of Corollary~\ref{cor:asymptotic_l1_one}
hold. Then for a given $h$, 
\[
\sqrt{n}D\left(\widehat{p}_{1,h}-\widehat{p}_{0,h},p_{1,h}-p_{0,h}\right)\to\int\left|\mathbb{G}(1,y)-\mathbb{G}(0,y)\right|dy\text{ weakly in }\mathbb{R}\text{ a.s.},
\]
where $\mathbb{G}$ is a centered Gaussian process with $Cov\left[\mathbb{G}(a_{1},y_{1}),\mathbb{G}(a_{2},y_{2})\right]=\mathbb{E}\left[f_{y_{1}}^{a_{1}}f_{y_{2}}^{a_{2}}\right]-\mathbb{E}\left[f_{y_{1}}^{a_{1}}\right]\mathbb{E}\left[f_{y_{2}}^{a_{2}}\right]$, 
$f_{y}^{a} = \frac{\mathbf{1}(A=a)}{{\pi}_{a}(X)}\big(K_{h,y}(Y)-\mu_{A,y}(X)\big)+\mu_{a,y}(X)$.
\end{corollary}

Unlike Corollary \ref{cor:asymptotic_l1_one}, Corollary \ref{cor:asymptotic_l1_twodiff} gives an asymptotic distribution of $D\left(\widehat{p}_{1,h}-\widehat{p}_{0,h},p_{1,h}-p_{0,h}\right)$ that contains both density components. This leads to the following bootstrap algorithm for obtaining $(1-\alpha)$ confidence intervals for $\psi^{cd}_h$.

\begin{algorithm}\mbox{}\\[-\baselineskip]
\begin{enumerate} \label{algorithm:density-effect-cd2}
\item We generate $B$ bootstrap samples $\{\tilde{Z}_{1}^{1},\ldots,\tilde{Z}_{n}^{1}\},\ldots,\{\tilde{Z}_{1}^{B},\ldots,\tilde{Z}_{1}^{B}\}$,
by sampling with replacement from the original sample. 
\item On each bootstrap sample, compute $T_{i}=\sqrt{n}D(\widehat{p}_{1,h}^{i}-\widehat{p}_{0,h}^{i},\widehat{p}_{1,h}-\widehat{p}_{0,h})$,
where $\widehat{p}_{a,h}^{i}$ is the kernel density estimator $\widehat{p}_{a,h}$ computed on $i$-th
bootstrap samples $\{\tilde{Z}_{1}^{i},\ldots,\tilde{Z}_{n}^{i}\}$. 
{
\item Compute $\alpha$-quantile $\hat{z}_{\alpha}=\inf\left\{ z:\,\frac{1}{B}\sum_{i=1}^{B}I(T_{i}>z)\leq\alpha\right\} $. 
\item Define $\widehat{C}_{\alpha}=\left[D(\widehat{p}_{1,h},\widehat{p}_{0,h})-\frac{\hat{z}_{\alpha}}{\sqrt{n}},\,D(\widehat{p}_{1,h},\widehat{p}_{0,h})+\frac{\hat{z}_{\alpha}}{\sqrt{n}}\right]$. }
\end{enumerate}
\end{algorithm}

The next corollary gives conditions under which the intervals produced by Algorithm \ref{algorithm:density-effect-cd2} are asymptotically valid.

\begin{corollary}
\label{cor:confidence_l1_twodiff}
Suppose that the same conditions of Corollary~\ref{cor:asymptotic_l1_twodiff}
hold, with the additional assumption that $\int\left|\mathbb{G}(1,y)-\mathbb{G}(0,y)\right|dy>0$.
Then, the intervals $\widehat{C}_{\alpha}$ constructed via Algorithm
\ref{algorithm:density-effect-cd2} are valid asymptotic confidence sets, i.e. 
\[
\liminf_{n\to\infty}\mathbb{P}\left(D(p_{0,h},p_{1,h})\in\widehat{C}_{\alpha}\right)\geq1-\alpha
\]
for given level of $\alpha$. 
\end{corollary}

When $p_{0,h}=p_{1,h}$, we have $\int\left|\mathbb{G}(1,y)-\mathbb{G}(0,y)\right|dy=0$, which contradicts the assumption of Corollary \ref{cor:confidence_l1_twodiff}. Therefore, unlike Algorithm \ref{algorithm:density-effect-cd1}, the intervals obtained via Algorithm \ref{algorithm:density-effect-cd2} are asymptotically valid when $p_{0,h} \neq p_{1,h}$, but not when $p_{0,h}=p_{1,h}$.

Although it is not valid at $p_{0,h}=p_{1,h}$, Algorithm \ref{algorithm:density-effect-cd2} provides tighter intervals than Algorithm \ref{algorithm:density-effect-cd1}.
The half widths of the confidence intervals obtained by Algorithms \ref{algorithm:density-effect-cd1} and \ref{algorithm:density-effect-cd2} are given by $\frac{\hat{z}_{\alpha/2}^{0}}{\sqrt{n}}+\frac{\hat{z}_{\alpha/2}^{1}}{\sqrt{n}}$ and $\frac{\hat{z}_{\alpha}}{\sqrt{n}}$, respectively. Further note that $\frac{\hat{z}_{\alpha/2}^{0}}{\sqrt{n}}+\frac{\hat{z}_{\alpha/2}^{1}}{\sqrt{n}}$
is an upper bound for the $\alpha$-quantile of $D(\hat{p}_{0,h},p_{0,h})+D(\hat{p}_{1,h},p_{1,h})$,
as implied by the following
\[
\liminf_{n\to\infty}P\left(D(\hat{p}_{0,h},p_{0,h})+D(\hat{p}_{1,h},p_{1,h})\leq\frac{\hat{z}_{\alpha/2}^{0}}{\sqrt{n}}+\frac{\hat{z}_{\alpha/2}^{1}}{\sqrt{n}}\right)\geq1-\alpha.
\]
Similarly, $\frac{\hat{z}_{\alpha}}{\sqrt{n}}$
is an upper bound for the $\alpha$-quantile of $D(\hat{p}_{1,h}-\hat{p}_{0,h},p_{1,h}-p_{0,h})$, as shown by
\[
\liminf_{n\to\infty}P\left(D(\hat{p}_{1,h}-\hat{p}_{0,h},p_{1,h}-p_{0,h})\leq\frac{\hat{z}_{\alpha}}{\sqrt{n}}\right)\geq1-\alpha.
\]
Now, by the triangle inequality it follows that
\[
\left|\left(\hat{p}_{1,h}(y)-\hat{p}_{0,h}(y)\right)-\left(p_{1,h}(y)-p_{0,h}(y)\right)\right|\leq\left|\hat{p}_{0,h}(y)-p_{0,h}(y)\right|+\left|\hat{p}_{1,h}(y)-p_{1,h}(y)\right|.
\]
Integrating on both sides of the above inequality gives
\[
D(\hat{p}_{1,h}-\hat{p}_{0,h},p_{1,h}-p_{0,h})\leq D(\hat{p}_{0,h},p_{0,h})+D(\hat{p}_{1,h},p_{1,h}),
\]
which leads to
\[
\frac{\hat{z}_{\alpha}}{\sqrt{n}}\leq\frac{\hat{z}_{\alpha/2}^{0}}{\sqrt{n}}+\frac{\hat{z}_{\alpha/2}^{1}}{\sqrt{n}}.
\]
Hence, the $(1-\alpha)$ confidence intervals obtained by Algorithm \ref{algorithm:density-effect-cd2} are tighter than those by Algorithm \ref{algorithm:density-effect-cd1} when $p_{0,h} \neq p_{1,h}$.

\subsubsection{Estimator based on smoothed $L_1$ distance} \label{subsubsec:inference-density-effect-sm}

Next, we discuss the inferential procedure for the estimator based on the smoothed $L_1$ distance. We require the following rate condition on the nuisance estimators $\widehat{\eta}^{sm}$:
\begin{itemize}[leftmargin=*]
	\item \customlabel{assumption:A4''}{(A4$''$)} \textit{Nonparametric conditions on 
 $\widehat{\pi}_{a}, \widehat{\nu}_a, \widehat{p}_a$}: \[
 \sup_{a\in \mathcal{A}}\left\{\Vert \widehat{\pi}_a - \pi_a \Vert \Vert \widehat{\nu}_a - \nu_a \Vert + \Vert \widehat{p}_a - p_a \Vert^2\right\}=o_\Pb(1/\sqrt{n}).
 \]
\end{itemize}

The next corollary describes the asymptotic properties of the estimator $\widehat{\psi}_{s}^{sm}$ in \eqref{eqn:estimator-psi-sm}.

\begin{corollary} \label{cor:asymptotic_smooth_l1} Suppose that
Assumptions \ref{assumption:A2}, \ref{assumption:A3}, \ref{assumption:A4''}, and \ref{assumption:A5}
hold. Further assume that $\Vert h'_{s}\Vert_{\infty},\Vert h''_{s}\Vert_{\infty}<\infty$ and $\sigma_{sm}^{2}\coloneqq\mathbb{E}\left[\phi_{s}^{sm}(Z;\eta^{sm})^{2}\right]\in(0,\infty)$. Then,
\begin{equation}
\sqrt{n}\left(\widehat{\psi}_{s}^{sm}-\psi_{s}^{sm}\right)\to\mathcal{N}(0,\sigma_{sm}^{2})\text{ weakly}.\label{eq:asymptotic_smooth_l1}
\end{equation}
\end{corollary}

The above weak convergence is obtained by using the result of Theorem \ref{thm:error-bound-psi-sm} and the central limit theorem.
Based on this result, we propose the following bootstrap algorithm to construct $(1-\alpha)$ confidence intervals for $\psi_{s}^{sm}$.

\begin{algorithm}\mbox{}\\[-\baselineskip]
\begin{enumerate} \label{algorithm:density-effect-sm}
\item We generate $B$ bootstrap samples $\{\tilde{Z}_{1}^{1},\ldots,\tilde{Z}_{n}^{1}\},\ldots,\{\tilde{Z}_{1}^{B},\ldots,\tilde{Z}_{1}^{B}\}$,
by sampling with replacement from the original sample. 
\item On each bootstrap sample, compute $T_{i}=\sqrt{n}\text{\ensuremath{\left|(\widehat{\psi}_{s}^{sm})^{i}-\widehat{\psi}_{s}^{sm}\right|}}$,
where $(\widehat{\psi}_{s}^{sm})^{i}$ is the estimator $\widehat{\psi}_{s}^{sm}$
computed on $i$-th bootstrap samples $\{\tilde{Z}_{1}^{i},\ldots,\tilde{Z}_{n}^{i}\}$. 
\item Compute $\alpha$-quantile $\hat{z}_{\alpha}=\inf\left\{ z:\,\frac{1}{B}\sum_{i=1}^{B}I(T_{i}>z)\leq\alpha\right\} $. 
\item Define $\widehat{C}_{\alpha}=\left[\widehat{\psi}_{s}^{sm}-\frac{\hat{z}_{\alpha}}{\sqrt{n}},\,\widehat{\psi}_{s}^{sm}+\frac{\hat{z}_{\alpha}}{\sqrt{n}}\right]$. 
\end{enumerate}
\end{algorithm}

In the next corollary, we show that the intervals produced by Algorithm \ref{algorithm:density-effect-sm} are asymptotically valid when $\psi_{s}^{sm} > 0$, but not when $\psi_{s}^{sm} = 0$.

\begin{corollary}
\label{cor:confidence_smooth_l1}
Assume that the same conditions of Corollary~\ref{cor:asymptotic_smooth_l1}
hold. Then the intervals $\widehat{C}_{\alpha}$ constructed via Algorithm
\ref{algorithm:density-effect-sm} are valid asymptotic confidence sets, i.e.,
\[
\liminf_{n\to\infty}\mathbb{P}\left(\psi_{s}^{sm}\in\widehat{C}_{\alpha}\right)\geq1-\alpha
\]
for given level of $\alpha$. 
\end{corollary}

The confidence intervals obtained by Algorithm \ref{algorithm:density-effect-sm} are guaranteed to be asymptotically valid only if $\psi_{s}^{sm} > 0$, i.e., when $p_1 \neq p_0$. When $p_1 = p_0$, the influence function $\phi^{sm}_s$ reduces to zero, which violates the premise of Corollary \ref{cor:confidence_smooth_l1} (see, for example, Section 5.2 of \citet{kennedy2021semiparametric} or Sections 12.3 and 20.1.1 of \citet{vanderVaart2000}). The presence of nuisance functions, combined with the fact that the terms like $\Vert h''_s \Vert_\infty$ grow to infinity as $s \rightarrow 0$, poses significant challenges for the inferential procedure when $p_1 = p_0$. This complication is not addressed within the scope of this work. Following \citet{kennedy2021semiparametric}, we recommend replacing $\hat{z}_{\alpha}$ with $\hat{z}_{\alpha}\vee 1$ in Step 4 of Algorithm \ref{algorithm:density-effect-sm}. This simple ad-hoc fix renders the interval $\widehat{C}_{\alpha}$ more conservative near the null $p_1 = p_0$ while maintaining its validity.

\subsubsection{Estimator based on margin condition} \label{subsubsec:inference-density-effect-mc}

We now discuss statistical inference for the estimator derived under the margin condition outlined in Definition \ref{assumption:MC}. To this end, we impose the following rate condition on nuisance estimators $\widehat{\eta}^{mc}$:
\begin{itemize}[leftmargin=*]
	\item \customlabel{assumption:A4'''}{(A4$'''$)} \textit{Nonparametric conditions on 
 $\widehat{\pi}_{a}, \widehat{\nu}_a, \widehat{p}_a$} under the margin condition: 
 \[
 \sup_{a \in\mathcal{A}}\left\{ \Vert \widehat{\pi}_a - \pi_a \Vert \Vert \widehat{\nu}_a - \nu_a \Vert + \left(\Vert \widehat{p}_a - p_a \Vert_{\infty} \right)^{\alpha+1} \right\}=o_{\mathbb{P}}\left(1/\sqrt{n}\right)
 \]
\end{itemize}

We describe the asymptotic properties of the estimator $\widehat{\psi}_{s}^{mc}$ in the next corollary.

\begin{corollary} \label{cor:asymptotic_margin} Suppose that Assumptions
\ref{assumption:A2}, \ref{assumption:A3}, \ref{assumption:A4'''}, and \ref{assumption:A6} hold, and that the margin condition holds for some $\alpha$. Further assume that $Y$ has the density function $p$ such that $\inf_{y \in \mathcal{Y}} p(y) \geq p_{\min} > 0$ and  $\sigma_{mc}^{2}\coloneqq\mathbb{E}\left[\varphi^{mc}(Z;\gamma^{mc},\eta^{mc})^{2}\right]<\infty$. Then,
\begin{equation}
\sqrt{n}\left(\widehat{\psi}^{mc}-\psi\right)\to\mathcal{N}(0,\sigma_{mc}^{2})\text{ weakly}.\label{eq:asymptotic_margin}
\end{equation}
\end{corollary}

Corollary \ref{cor:asymptotic_margin} is a direct consequence of Theorem \ref{thm:error-bound-psi-mc} and the central limit theorem. Similarly to the approach based on smoothing $L_1$ distance, we present the following bootstrap algorithm; it is essentially the same as Algorithm \ref{algorithm:density-effect-sm}, but we reiterate here for the sake of clarity and completeness.

\begin{algorithm}\mbox{}\\[-\baselineskip]
\begin{enumerate} \label{algorithm:density-effect-mc}
\item We generate $B$ bootstrap samples $\{\tilde{Z}_{1}^{1},\ldots,\tilde{Z}_{n}^{1}\},\ldots,\{\tilde{Z}_{1}^{B},\ldots,\tilde{Z}_{1}^{B}\}$,
by sampling with replacement from the original sample. 
\item On each bootstrap sample, compute $T_{i}=\sqrt{n}\text{\ensuremath{\left|(\hat{\psi}^{mc})^{i}-\hat{\psi}^{mc}\right|}}$,
where $(\hat{\psi}^{mc})^{i}$ is the estimator $\hat{\psi}^{mc}$
computed on $i$-th bootstrap samples $\{\tilde{Z}_{1}^{i},\ldots,\tilde{Z}_{n}^{i}\}$. 
\item Compute $\alpha$-quantile $\hat{z}_{\alpha}=\inf\left\{ z:\,\frac{1}{B}\sum_{i=1}^{B}I(T_{i}>z)\leq\alpha\right\} $. 
\item Define $\widehat{C}_{\alpha}=\left[\hat{\psi}^{mc}-\frac{\hat{z}_{\alpha}}{\sqrt{n}},\,\hat{\psi}^{mc}+\frac{\hat{z}_{\alpha}}{\sqrt{n}}\right]$. 
\end{enumerate}
\end{algorithm}

Next corollary shows that the intervals produced by Algorithm \ref{algorithm:density-effect-mc} are asymptotically valid under the same conditions used in Corollary~\ref{cor:asymptotic_margin}.

\begin{corollary}
\label{cor:confidence_smooth_margin}
Assume that the same conditions of Corollary~\ref{cor:asymptotic_margin}
hold. Then, the intervals $\widehat{C}_{\alpha}$ constructed via Algorithm
\ref{algorithm:density-effect-mc} are valid asymptotic confidence sets, i.e.,
\[
\liminf_{n\to\infty}\mathbb{P}\left(\psi\in\widehat{C}_{\alpha}\right)\geq1-\alpha
\]
for given level of $\alpha$. 
\end{corollary}

Note that $\sigma_{mc}^{2} = 0$ implies $p_0 = p_1$. Since the margin condition implies $p_0 \neq p_1$, we have $\sigma_{mc}^{2} > 0$. Unlike the previous approaches, inference at the null $p_0 = p_1$ is not feasible, regardless of how conservative we are willing to be, as it is assumed a priori to be a zero-probability event.





\subsubsection{Testing no effect} \label{subsubsec:testing-no-effect}

Probably it is testing the null hypothesis of no marginal distributional effect
\begin{align*}
H_0: p_0(y) = p_1(y) \quad \text{for all} \quad y \in \mathcal{Y}    
\end{align*}
that is of primary importance to the majority of practitioners. However, performing inference under the null hypothesis $H_0$ is notoriously challenging for a general semiparametric estimator. This difficulty arises because the target parameter $\psi$, along with its smoothed versions $\psi^{cd}_h$ and $\psi^{sm}_s$, is not pathwise differentiable under $H_0$, as $\psi=0$ (or $\psi^{cd}_h=0$, $\psi^{sm}_s=0$) is on the boundary of the parameter space. To our knowledge, there is no general solution for this. Here, we show that some of our inferential procedures in Section \ref{subsec:inference-density-effect} can be utilized for testing the null of no effect.

Based on Algorithm \ref{algorithm:density-effect-cd1}, asymptotically valid $(1-\alpha)$ confidence intervals for $\psi^{cd}_h$ can be constructed even when $p_{0,h} = p_{1,h}$. The next proposition shows that one may use the intervals obtained by Algorithm \ref{algorithm:density-effect-cd1} to test the null of no effect.

\begin{proposition} \label{prop:testing-the-null}
Let $\Pb,\Qb$ be two probability distributions on $\mathbb{R}^{d}$,
and $p,q$ be their continuous density functions. For a kernel function $K:\mathbb{R}^{d}\to[0,\infty)$, suppose that $K$ has a compact support, $K$ is continuous at $0$, and $K(0)>0$. If $p\neq q$, then there exists $h_{0}>0$
such that for all $h\in(0,h_{0}]$, $p_{h}\neq q_{h}$.
\end{proposition}

Let $\widehat{C}_{\alpha}$ denote the interval constructed using  Algorithm \ref{algorithm:density-effect-cd1}. Proposition \ref{prop:testing-the-null} implies when $h$ is sufficiently small, we can reject $H_0$ at level $\alpha$ if $0 \notin \widehat{C}_{\alpha}$. Importantly, it requires the bandwidth $h$ to be sufficiently small, but not necessarily go to $0$ or be adaptive to data. Similar results have been presented in the literature on topological data analysis where our aim is the reliable topological information rather than the exact density function, equivalent to us targeting the difference between two density functions rather than their exact inference. One criterion for choosing the bandwidth is to maximize statistically significant homological features by tracking the evolution of the persistence of the homological features as the bandwidth varies (see \citet[Section 7.1]{ChazalFLMRW2017} for details). One may adapt this strategy to choose $h$ for testing the null $H_0$.

Algorithm \ref{algorithm:density-effect-sm} with the estimator based on the smoothed $L_1$ distance can be used for testing $H_0$ as well. Suppose there exists a function $G(\cdot)$ such that $\left\vert h_s(t) - \vert t \vert \right\vert \leq G(s)$ and $\text{Area}(\mathcal{Y}) < \infty$. Then based on our discussion in Section \ref{subsubsec:inference-density-effect-sm}, one may use 
\begin{align} \label{eqn:CI-sm-conservative}
\left[\widehat{\psi}_{s}^{sm}-\frac{\hat{z}_{\alpha}\vee 1}{\sqrt{n}}-G(s)\text{Area}(\mathcal{Y}),\,\widehat{\psi}_{s}^{sm}+\frac{\hat{z}_{\alpha}\vee 1}{\sqrt{n}}+G(s)\text{Area}(\mathcal{Y})\right]
\end{align}
as a $(1-\alpha)$ asymptotically valid confidence interval for the original density effect $\psi$. If $0$ is not contained in the above interval, then we reject $H_0$ at level $\alpha$. However, for some $h_s$, finding the function $G$ may not be straightforward.

Regrettably, Algorithm \ref{algorithm:density-effect-cd2}, which yields tighter confidence intervals compared to Algorithm \ref{algorithm:density-effect-cd1}, cannot be employed for testing no effect, as it is invalid when $\psi^{cd}_h=0$. Algorithm \ref{algorithm:density-effect-mc} is not applicable as well, since we rely on the margin condition that rules out the event $\psi=0$ a priori. 

\begin{remark} \label{remark:alternative-bootstrap}
    In this section, we employ the standard bootstrap approach, where we do not adapt cross-fitting. Consequently, a portion of the sample must be dedicated exclusively to nuisance estimation. This does not impact the rate of convergence, and it remains unclear whether a general method exists that would allow us to achieve full-sample efficiency. Alternative bootstrap methods, such as the multiplier bootstrap, may be considered. The multiplier bootstrap does not require re-estimating the nuisance functions, as discussed in \citet{kennedy2019nonparametric}. To show the validity of the multiplier bootstrap, it is necessary to verify that the results from \citet{belloni2018uniformly} can be applied to our proposed estimators. We plan to explore this alternative approach in future work.
\end{remark}

\section{Numerical Illustration} \label{sec:num-illustration}

\begin{figure}[t!]
    \centering
    \begin{minipage}{.325\textwidth}
      \centering
      \includegraphics[width=.975\linewidth]{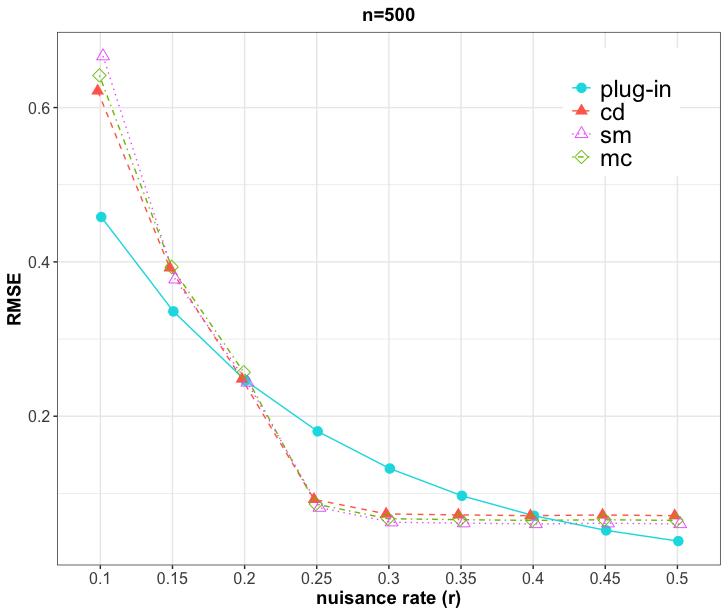}
    \end{minipage}
    \hfill%
    \begin{minipage}{.325\textwidth}
      \centering
      \includegraphics[width=.975\linewidth]{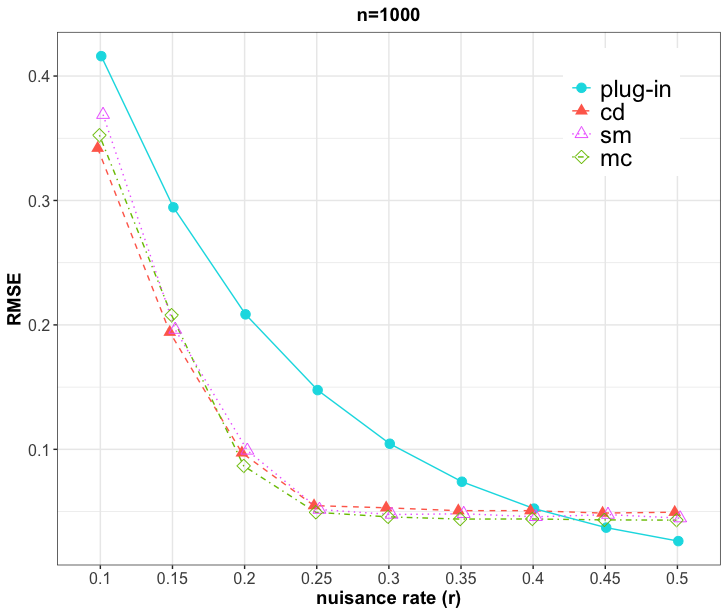}
    \end{minipage}
    \hfill%
    \begin{minipage}{.325\textwidth}
      \centering
      \includegraphics[width=.975\linewidth]{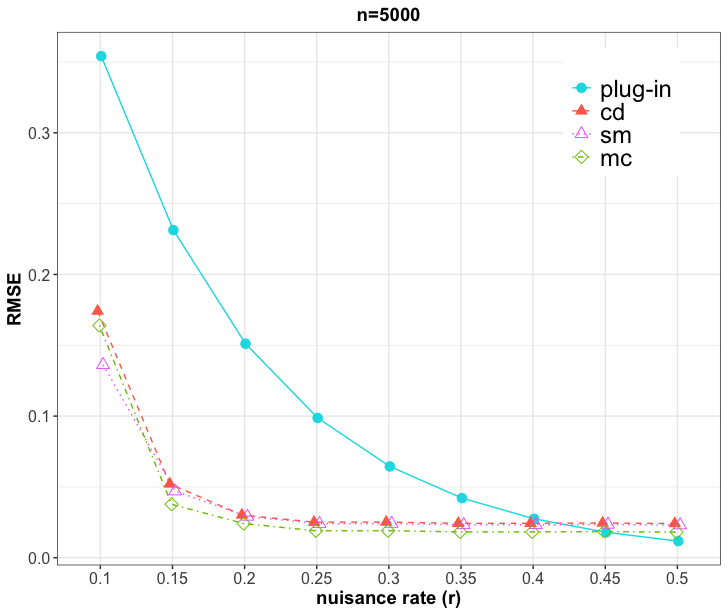}
    \end{minipage}
    \caption{RMSE versus nuisance error parameter $r$ across different sample sizes}
    \label{fig:simulation-rates}
\end{figure}

In order to assess the performance of the proposed estimators, we conduct a small simulation study with the following simplified scenario: $Y^0 \sim \text{Uniform}[-1/2, 1/2]$ and $Y^1 \sim \text{Uniform}[-1, 1]$, $X_1 \sim \text{Uniform}[0, 1]$, $A \sim \text{Bernoulli}(\pi_1(X_1))$ where $\pi_1(x)=\min\{\max\{x^2, 0.10\}, 0.90\}$, $X_2 = 1/2(AY^1 + (1-A)Y^0)$, $X=(X_1,X_2)$, and $Y=2X_2$. In this simplified setting, the margin condition in Definition \ref{assumption:MC} holds with any $\alpha > 0$. For the nuisance estimates, we define $\widehat{\pi}_1 = \text{expit}\left(\text{logit}(\pi_1(X)) + \epsilon_{\pi} \right)$, $\widehat{\mu}_{a,y} = \mu_{a,y} + \epsilon_{\mu}$, $\widehat{p}_a=\widehat{\nu}_a = p_a + \epsilon_{p}$, where $\epsilon_{\pi}, \epsilon_{\mu}, \epsilon_{p} \sim N(n^{-r}, n^{-2r})$. These choices guarantee that $\Vert \widehat{\pi}_a - {\pi}_a\Vert= O(n^{-r})$, $\Vert \widehat{\mu}_{a,y} - {\mu}_{a,y}\Vert= O(n^{-r})$, $\Vert \widehat{p}_a - p_a\Vert=\Vert \widehat{\nu}_a - {\nu}_a\Vert= O(n^{-r})$, $\forall a \in \{0,1\}$, which allows us to study the performance of the proposed estimators under different nuisance estimation errors solely depending on $r$. For $n \in \{500, 1000, 5000\}$, we generate data and compute the necessary nuisance estimates as described above, varying $r \in \{0.10 + 0.05j : j =0,1, \ldots, 8\}$. Then, we construct the three estimators $\widehat{\psi}^{cd}_h, \widehat{\psi}^{sm}_s, \widehat{\psi}^{mc}$ that we propose in Section \ref{sec:densiti-effects-estimator}, together with the plug-in estimator $\widehat{p}_a$ whose error is expected to be the order of the nuisance error $O(n^{-r})$. For $\widehat{\psi}^{cd}_h$, we use the Silverman kernel which is of order $3$ \citet{Tsybakov10}, setting $h=0.05$. For $\widehat{\psi}^{sm}$, we use $h_s(t) = t \text{tanh}(ts)$, yielding $G(s) = 1/s$, with $s=500$. Finally, we compute root-mean-squared error (RMSE) of each estimator across $500$ simulations for each $(n, r)$. Results are presented in Figure \ref{fig:simulation-rates}.

The results show that all three proposed estimators generally outperform the plug-in estimator, especially for larger sample sizes. Importantly, as predicted by our theoretical results in Section \ref{sec:densiti-effects-estimator}, when the nuisance error is only on the order of $O(n^{-1/4})$ their performance becomes comparable to the optimal performance of the plug-in estimator (i.e., parametric $\sqrt{n}$ rates). In our simulations, $\widehat{\psi}^{cd}_h$ and $\widehat{\psi}^{sm}_s$ appear to have a small non-vanishing smoothing bias, which might be improved further by choosing a different kernel and approximating function. 

Next, we assess the coverage of our proposed confidence intervals for each estimator. We posit two different scenarios in which we use the same data generation processes to produce Figure \ref{fig:example} (Scenario 1) and Figure \ref{fig:simulation-rates} (Scenario 2). In both scenarios, although $p_1$ and $p_0$ have the same mean (both have zero ATE by construction), $p_1$ clearly differs from $p_0$ in significant respects, yielding $\psi$ significantly greater than zero. In Scenario 1, a sample is generated with $X \sim \text{Uniform[0,1]}$ and $A \sim \text{Bernoulli(0.5)}$, and we construct all the nuisance estimators in an independent, separate sample with the same size. In Scenario 2, we set $r=0.25$ when constructing the nuisance estimators. Confidence intervals for the estimators $\widehat{\psi}^{cd}_h$, $\widehat{\psi}^{sm}_s$, $\widehat{\psi}^{mc}$ are obtained using Algorithm \ref{algorithm:density-effect-cd2}, the formula \eqref{eqn:CI-sm-conservative}, Algorithm \ref{algorithm:density-effect-mc}, respectively. For $n \in \{500, 1000, 5000\}$, we repeat the simulation $500$ times and compute the coverage. Results are given in Table \ref{tbl:simulation-density-effect-coverage}. As expected, coverage was generally near the nominal level (99\%) in large samples, while the intervals from Algorithm \ref{algorithm:density-effect-cd2} and the formula \eqref{eqn:CI-sm-conservative} produce slightly lower- and higher-than-nominal coverages, respectively. 

\begin{table}[!t]
    \centering
    \begin{tabular}{cccccccccc}
    \toprule
         n&  \multicolumn{3}{c}{500}&  \multicolumn{3}{c}{1000}&  \multicolumn{3}{c}{5000}\\
         Estimator & $\widehat{\psi}^{cd}_h$ & $\widehat{\psi}^{sm}_s$ & $\widehat{\psi}^{mc}$ & $\widehat{\psi}^{cd}_h$  & $\widehat{\psi}^{sm}_s$ & $\widehat{\psi}^{mc}$ & $\widehat{\psi}^{cd}_h$  & $\widehat{\psi}^{sm}_s$ & $\widehat{\psi}^{mc}$ \\
     \midrule \midrule
         Scenario 1&  90.6\% &  92.8\%&  93.4\%&  92.2\%&  99.4\%&  96.8\%&  97.2\%&  99.8\%& 98.8\%\\
         Scenario 2&  92.6\%&  95.8\%&  95.2\%&  96.6\%&  99.8\%&  98.4\%&  98.8\%&  100\%& 99.2\%\\
    \bottomrule
    \end{tabular}
    \caption{Coverage of the proposed 99\% confidence intervals across 500 simulations.} 
    \label{tbl:simulation-density-effect-coverage}
\end{table}

We also test the null hypothesis of no distributional effect using the confidence intervals in Algorithm \ref{algorithm:density-effect-cd1} and the formula \eqref{algorithm:density-effect-sm}, as discussed in Section \ref{subsubsec:testing-no-effect}. We compute rejection percentage per each scenario at level $\alpha = 0.01$ across 500 simulations, and present results in Table \ref{tbl:no-effect-test}. The results show that both tests tend to correctly reject the null for sufficiently large $n$.

\begin{table}[!t]
    \centering
    \begin{tabular}{ccccccc} 
    \toprule
      n   &  \multicolumn{2}{c}{500}&  \multicolumn{2}{c}{1000}&  \multicolumn{2}{c}{5000}\\ 
      Estimator    &  $\widehat{\psi}^{cd}_h$ & $\widehat{\psi}^{sm}_s$ &  $\widehat{\psi}^{cd}_h$ &  $\widehat{\psi}^{sm}_s$ &  $\widehat{\psi}^{cd}_h$  & $\widehat{\psi}^{sm}_s$ \\ 
    \midrule \midrule         
         Scenario 1& 87.6\%& 85.2\%& 95.0\%& 93.2\%& 98.8\%& 99.4\%\\ 
         Scenario 2& 92.8\%& 89.2\%& 95.6\%& 95.8\%& 99.6\%& 99.8\%\\ 
    \bottomrule          
    \end{tabular}
    \caption{Rejection percentage of the proposed no effect tests across 500 simulations.} 
    \label{tbl:no-effect-test}
\end{table}

\begin{figure}[!t]
\centering
\includegraphics[width=.95\linewidth, trim=2 2 2 2,clip]{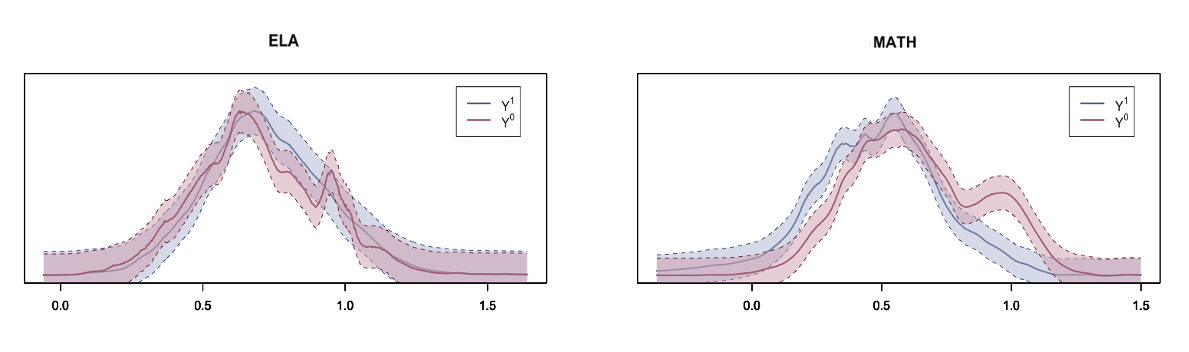}
\caption{Estimated counterfactual densities with $99\%$ confidence bands for the achievement gap in Math and ELA subjects between White and non-White students.}
\label{fig:achievement_gap_density}
\end{figure}

Lastly, we illustrate the use of the proposed counterfactual density estimator \eqref{eqn:doubly-robust} with a real-world dataset where we analyze the effect of free lunch on achievement gap. Many public schools in the US provide free lunch for qualifying students, one aim of which is to equalize academic performance for disadvantaged groups. Using data from the Stanford Education Data Archive (SEDA), we attempt to investigate the effect of providing free lunch on the improvement on the achievement gap among students of different ethnic groups. We collected the test score gaps and percent free lunch provided, as well as other socioeconomic and demographic characteristics of geographical school districts during 2009-2012. We consider a school district treated ($A=1$) if it is providing above-average free lunch to students. Our outcome is the achievement gap in Math and ELA test scores between White and non-White (specifically Hispanic and Black) students. Results are presented in Figures \ref{fig:achievement_gap_density}. The free lunch program tends to narrow the achievement gap. The distributional shift appears to be more marked for the gap in Math test scores; when we use Algorithm \ref{algorithm:density-effect-cd1}, we reject the null of no effect at $\alpha=0.99$ for the gap in Math test scores, but not for the ELA test scores; however, both are rejected at $\alpha=0.95$. Interestingly, the small right peak found in the density curve for $Y^0$ does not appear in the density curve for $Y^1$. This suggests that the free lunch program may be exerting a significant impact on a subset of districts that would otherwise experience considerable achievement gaps in the absence of the program.

\section{Discussion}
Characterizing causal effects using standard summary statistics, such as averages, may often be insufficient to uncover important insights into how a treatment works. In this study, we pursued a more nuanced approach to probing causal effects based on counterfactual distributions. We developed a doubly-robust estimator for counterfactual densities and analyzed its asymptotic properties and risk bounds, which are conjectured to be minimax optimal under appropriate nonparametric conditions. We also defined a causal effect using the non-smooth $L_1$ distance between counterfactual distributions and discussed three approaches for efficient estimation. We presented bootstrap algorithms for constructing the relevant confidence bands and intervals. Based on our findings, we also discussed testing the null hypothesis of no distributional effect. We remain agnostic regarding which approach in Section \ref{sec:densiti-effects-estimator} should be preferred in practice, as each has its own strengths and limitations in terms of estimation and inference.

Lastly, a few remarks are set out as follows. Our proposed method can always be used jointly with ATE estimators, as a first step in assessing whether there is evidence of treatment effects beyond a mean shift; for instance, it is possible that the ATE is nearly zero but the proposed distributional effect is large. In a similar vein, even when we are interested in other types of functionals, such as quantiles and CDFs, our results could be useful for testing hypotheses with respect to those quantities. Also, as noted in Remark \ref{remark:alternative-bootstrap}, alternative bootstrap approaches may be considered, which could potentially eliminate the need to reserve a portion of the sample solely for nuisance estimation. There are other interesting avenues for future work. These include adaptation of variable bandwidth in our bootstrap algorithms for the proposed kernel-smoothed estimators, computation of the minimax lower bounds for counterfactual density estimation, as well as extensions to other causal inference setups, when the treatment is continuous, or time-varying, or afflicted by unmeasured confounding.


\newpage

\bibliography{bibliography}

\appendix

\input{appendix}

\end{document}

%% file: appendix.tex
\newpage
\section*{\centerline{Supplementary Materials}}
\vspace*{.1in}
\setcounter{equation}{0}
\renewcommand{\theequation}{A.\arabic{equation}}


\section{Empirical processes} \label{sec:appendix-empirical-processes}

Here, we present some of the modern empirical process tools and techniques that we apply in the proofs for the results in Section \ref{sec:confidence_interval}.
Let $X_{1},\ldots,X_{n}$ be an independent and identically distributed
random sample taking values in the measure space $(\mathcal{X},P)$.
For a measurable function $f:\mathcal{X}\to\mathbb{R}$, we denote
$Pf=\int fdP$ and $P_{n}f=\int fdP_{n}=\frac{1}{n}\sum_{i=1}^{n}f(X_{i})$.

The definition of the Donsker class below is taken from Section~19.2 in \citet{vanderVaart2000}
or Section~2.1 in \citet{Kosorok2008}.

\begin{definition}

A class $\mathcal{F}$ of measurable functions $f:\mathbb{X}\to\mathbb{R}$
is called $P$-Donsker if the process $\{\sqrt{n}(P_{n}f-Pf)\}_{f\in\mathcal{F}}$
converges in distribution to a limit process in the space $\ell^{\infty}(\mathcal{F})$.
The limit process is a Gaussian process $\mathbb{G}$ with zero mean
and covariance function $Cov\left(\mathbb{G}f,\mathbb{G}g\right)\coloneqq\mathbb{E}_{P}\left[fg\right]-\mathbb{E}_{P}\left[f\right]\mathbb{E}_{P}\left[g\right]$;
this process is known as a Brownian Bridge.

\end{definition}

The following proposition is taken from Theorem~19.5 in \citet{vanderVaart2000} (or
Theorem~2.3 in \citet{Kosorok2008}).

\begin{proposition}

Let $\mathcal{F}$ be a class of measurable functions with $\int_{0}^{1}\sqrt{\log\mathcal{N}_{[]}\left(\mathcal{F},L_{2}(P),\epsilon\right)}d\epsilon<\infty$.
Then $\mathcal{F}$ is $P$-Donsker.

\end{proposition}

One sufficient condition for Donsker class is to assume a bound on the
covering number: fix a probability measure $Q$. A set ${\cal C}=\{f_{1},\ldots,f_{N}\}$
is an $\epsilon$-cover of ${\cal F}$ if, for every $f\in{\cal F}$
there exists a $f_{j}\in{\cal C}$ such that $\left\Vert f-f_{j}\right\Vert _{L_{2}(Q)}<\epsilon$.
The size of the smallest $\epsilon$-cover is called the covering
number and is denoted by $\mathcal{N}({\cal F},L_{2}(Q),\epsilon)$.
For a function class $\mathcal{F}$, we say a function $F$ is an
envelope of $\mathcal{F}$ if for all $f\in\mathcal{F}$ and for all
$x\in\mathbb{X}$, $f(x)\leq F(x)$. The following proposition is Theorem~19.14
in \citet{vanderVaart2000} or Theorem~2.5 in \citet{Kosorok2008}.

We use the following notation
for the integral of the covering number: 
\[
J(\mathcal{F},F,\delta)\coloneqq\int_{0}^{\delta}\sqrt{\log\sup_{Q}\mathcal{N}(\mathcal{F},L_{2}(Q),\tau\left\Vert F\right\Vert _{Q,2})}d\tau.
\]

\begin{proposition}

\label{prop:empirical_donsker_covering}

Let $\mathcal{F}$ be an appropriately measurable class of measurable
functions, and let $F$ be an envelop of $\mathcal{F}$. If $PF^{2}<\infty$
and 
$J(\mathcal{F},F,1)<\infty$, 
then $\mathcal{F}$ is $P$-Donsker.

\end{proposition}

Let $P_{n}^{*}f=\frac{1}{n}\sum_{i=1}^{n}f(X_{i}^{*})$ where $\{X_{1}^{*},\ldots,X_{n}^{*}\}$
is a bootstrap sample from $P_{n}$. the measure that puts mass $1/n$
on each element of the sample $\{X_{1},\ldots,X_{n}\}$. The following proposition
is Theorem~23.7 in \citet{vanderVaart2000} or a combination of Theorem~2.7
and 2.8 in \citet{Kosorok2008}.

\begin{proposition}

\label{prop:empirical_donsker_bootstrap}

$\mathcal{F}$ is $P$-Donsker and $F$ is an envelope of $\mathcal{F}$
with $PF^{2}<\infty$ if and only if $\{\sqrt{n}(P_{n}^{*}f-P_{n}f)\}_{f\in\mathcal{F}}$
converges in distribution to $\mathbb{G}$ in $\ell^{\infty}(\mathcal{F})$
a.s..

\end{proposition}

A trivial corollary is that by considering a single function class,
$\sqrt{n}(P_{n}f-Pf)$ and $\sqrt{n}(P_{n}^{*}f-P_{n}f)$ have the
same normal distribution as their convergence limits.

\section{Notations and Useful Lemmas}

We present some notation used throughout our proofs.
Fix a kernel function $K:\mathbb{R}^{d}\to\mathbb{R}$. For a bandwidth
$h>0$ and any $y\in\mathcal{Y}$, let $K_{h,y}$ be the normalized
kernel function at $y$, i.e., $K_{h,y}(y')=\frac{1}{h^{d}}K\left(\frac{y-y'}{h}\right)$.
For $h>0$, $a\in A$ and $y\in\mathcal{Y}$, define $f_{h,y}^{a}:\mathcal{X}\times\mathcal{A}\times\mathcal{Y}\to\mathbb{R}$
and $\hat{f}_{h,y}^{a}:\mathcal{X}\times\mathcal{A}\times\mathcal{Y}\to\mathbb{R}$
by 
\begin{align*}
f_{h,y}^{a} \equiv  f_{h,y}^{a}(Z;\eta) & =\frac{\mathbbm{1}(A=a)}{{\pi}_{a}(X)}\left(K_{h,y}(Y)-{\mu}_{A,y}(X)\right)+{\mu}_{a,y}(X),\\
\hat{f}_{h,y}^{a} \equiv f_{h,y}^{a}(Z;\hat{\eta}) & =\frac{\mathbbm{1}(A=a)}{\widehat{\pi}_{a}(X)}\left(K_{h,y}(Y)-\hat{\mu}_{A,y}(X)\right)+\hat{\mu}_{a,y}(X).
\end{align*}
Then our estimator $\hat{p}_{a,h}(y)$ in \eqref{eqn:doubly-robust}
becomes 
\[
\hat{p}_{a,h}(y)=\mathbb{P}_{n}\hat{f}_{h,y}^{a}.
\]

For a bandwidth $h>0$, let $\mathcal{Y}_{h}$ be 
\begin{equation}
\mathcal{Y}_{h}:=\left\{ y\in\mathbb{R}^{d}:\, \exists u\in\mathbb{R}^{d}\text{ with }\left\Vert y-u\right\Vert \leq R_{K}h\right\},\label{eq:notation_support_y}
\end{equation}
where $R_{K}$ is from Assumption \ref{assumption:A1}. When $h$ is fixed, boundedness on $\mathcal{Y}_{h}$ and $\mathcal{Y}$ are equivalent.
We also denote the collection of functions $f_{h,y}^{a}(\cdot;\bar{\eta})$
as 
\begin{equation}
\mathcal{F}_{\bar{\eta},h}\coloneqq\left\{ f_{h,y}^{a}(\cdot;\bar{\eta}):\,y\in\mathcal{Y}_{h},a\in\mathcal{A}\right\} .\label{eq:notation_doubly_robust_class_general}
\end{equation}

In particular, we denote the collection
of functions $f_{h,y}^{a}$, $\hat{f}_{h,y}^{a}$ as 
\begin{equation}
\mathcal{F}_{h}\coloneqq\mathcal{F}_{\eta,h}\left\{ f_{h,y}^{a}:\,y\in\mathcal{Y}_{h},a\in\mathcal{A}\right\} ,\qquad\hat{\mathcal{F}}_{h}\coloneqq\mathcal{F}_{\hat{\eta},h}\left\{ \hat{f}_{h,y}^{a}:\,y\in\mathcal{Y}_{h},a\in\mathcal{A}\right\} .\label{eq:notation_doubly_robust_class}
\end{equation}

Furthermore, we denote the collection of difference of $\hat{\mathcal{F}}_{h}$
and $\mathcal{F}_{h}$ as 
\begin{equation}
\bar{\mathcal{F}}_{h}\coloneqq\left\{ \hat{f}_{h,y}^{a}-f_{h,y}^{a}:\,y\in\mathcal{Y}_{h},a\in\mathcal{A}\right\} .\label{eq:notation_doubly_robust_class_diff}
\end{equation}


Below are some helpful lemmas that we repeatedly employ in proofs.

\begin{lemma} \label{lem:distance_distribution_triangle}

For probability densities $p_{1}$, $p_{2}$, $p_{3}$, $p_{4}$, and the associated distance function $D$, we have

\[
\left|D(p_{1},p_{2})-D(p_{3},p_{4})\right|\leq D(p_{1},p_{3})+D(p_{2},p_{4}).
\]

\end{lemma}

\begin{proof}

Since $D$ is distance measure, by triangle inequality it follows
\begin{align*}
 & D(p_{1},p_{2})\leq D(p_{1},p_{3})+D(p_{3},p_{4})+D(p_{4},p_{2}),\\
 & D(p_{3},p_{4})\leq D(p_{3},p_{1})+D(p_{1},p_{2})+D(p_{2},p_{4}),
\end{align*}
and consequently we obtain 
\[
\left|D(p_{1},p_{2})-D(p_{3},p_{4})\right|\leq D(p_{1},p_{3})+D(p_{2},p_{4}).
\]

\end{proof}

\begin{lemma} \label{lem:unbiasedness} 
Recall that $p_{a,h}(y) = \E \left\{ \E \left[ K_{h,y}(Y) \Big\vert X, A=a  \right] \right\}$ as defined in (\ref{eqn:psi_h}). Then under the causal assumptions \ref{assumption:c1} - \ref{assumption:c3}, we have
\begin{equation} \label{eqn:appdx.1}
\begin{aligned}
p_{a,h}(y) &= \E \left\{ \mu_{a,y}(X) \right\} = \E\left\{ f_{h,y}^{a}(Z) \right\} 
\end{aligned}
\end{equation}
\end{lemma}

\begin{proof}
The first equality in (\ref{eqn:appdx.1}) immediately comes from the definition. Now by the law of total expectation, we have
\begin{align*}
\E & \left\{ \frac{\mathbbm{1}(A=a)}{{\pi}_a(X)}\Big(K_{h,y}(Y) - {\mu}_{a,y}(X) \Big) +{\mu}_{a,y}(X) \right\} \\
& = \E \left\{ \E \left[ \frac{\mathbbm{1}(A=a)}{{\pi}_a(X)}\Big(K_{h,y}(Y) - {\mu}_{a,y}(X) \Big) +{\mu}_{a,y}(X) \Big\vert A,X \right] \right\} \\
&= \E \left\{  \frac{\mathbbm{1}(A=a)}{{\pi}_a(X)}\Big(\E [K_{h,y}(Y)|A,X] - {\mu}_{a,y}(X) \Big) +{\mu}_{a,y}(X) \Big\vert \right\} \\
&= \E \left\{  \frac{\mathbbm{1}(A=a)}{{\pi}_a(X)}\Big({\mu}_{a,y}(X) - {\mu}_{a,y}(X) \Big) +{\mu}_{a,y}(X) \Big\vert \right\} \\
&= \E \left\{ \mu_{a,y}(X) \right\},
\end{align*}
and the second equality follows.
\end{proof}

\begin{lemma}[Sample-splitting with $L_q$-norm]
\label{lem:empirical_bound_l2}
Let $\mathbb{P}_{n}$ denote the empirical measure over a set $D_1^n=(Z_{1},\ldots,Z_{n})$,
which is i.i.d. from $\mathbb{P}$. Let $\hat{f}$ be a sample operator (e.g., estimator) constructed in a separate, independent sample set $D_2^{m}$ with $m$ observations. Then for any $q \geq 2$, we have
\[
\begin{aligned}
\left\Vert (\mathbb{P}_{n}-\mathbb{P})\hat{f} \right\Vert_q^q \lesssim n^{1-q} \Vert \hat{f} - \Pb \hat{f} \Vert_q^q + n^{-\frac{q}{2}} \left\{\var(\hat{f})\right\}^{\frac{q}{2}},
\end{aligned}
\]
where $\left\Vert \hat{f}\right\Vert$ is a function of $m$.
\end{lemma}

\begin{proof}
First by Rosenthal’s inequality, it follows that
\begin{align*}
\Pb\left\{\left\vert(\mathbb{P}_{n}-\mathbb{P})\widehat{f}\right\vert^{q}\right\} & =\Pb\left\vert \frac{1}{n}\sum_{i=1}^{n}\left(\hat{f}(Z_{i})-\Pb\left\{\hat{f}(Z_{i})\right\}\right)\right\vert^{q}\\
 & \leq \frac{1}{n^{q}} C_q \left\{ \sum_{i=1}^{n}\Pb \left\vert \hat{f}(Z_{i})-\Pb\left\{\hat{f}(Z_{i})\right\}\right\vert^{q} + \left( \sum_{i=1}^{n} \Pb \left[ \hat{f}(Z_{i})-\Pb\left\{\hat{f}(Z_{i})\right\} \right]^2 \right)^{\frac{q}{2}} \right\},
\end{align*}
for some constant $C_q > 0$ depending only on $q$.

Hence, since 
\begin{align*}
    \Pb \left[ \hat{f}(Z_{i})-\Pb\left\{\hat{f}(Z_{i})\right\} \right]^2 = \var(\hat{f}(Z_{i})), 
\end{align*}
we obtain the desired result that
\begin{align*}
    \Pb\left\{\left\vert(\mathbb{P}_{n}-\mathbb{P})\widehat{f}\right\vert^{q}\right\} & \leq \frac{C_q}{n^{q}} \left\{ 2^q n \Vert \hat{f} - \Pb \hat{f} \Vert_q^q + \left( n \var(\hat{f}) \right)^{\frac{q}{2}} \right\} \\
    & \lesssim n^{1-q} \Vert \hat{f} - \Pb \hat{f} \Vert_q^q + n^{-\frac{q}{2}} \left\{\var(\hat{f})\right\}^{\frac{q}{2}}.
\end{align*}
\end{proof}

\begin{remark} \label{rmk:empirical_bound-1}
One may use a more succinct (but somewhat looser) form of Lemma \ref{lem:empirical_bound_l2} by upper bounding each term in the inequality with the same term. Note that 
\begin{align*}
    \Pb\left\vert \hat{f}(Z_{i})-\Pb\left\{\hat{f}(Z_{i})\right\} \right\vert^{q} &\leq 2^{q-1} \left\{ \Pb\left\vert \hat{f}(Z_{i}) \right\vert^q + \left\vert \Pb\hat{f}(Z_{i}) \right\vert^q \right\}\\
    &\leq 2^q \Pb\left\vert \hat{f}(Z_{i}) \right\vert^q,
\end{align*}
where the last inequality follows by Jensen's inequality, and that
\[
\var(\hat{f}(Z_{i})) \leq \Pb\left\{\hat{f}(Z_{i})\right\}^2 \leq \Vert \hat{f} \Vert_q^2.
\]

Now, since $-\frac{q}{2} \geq 1 - q$ for all $q \geq 2$, the following result is an immediate consequence of Lemma \ref{lem:empirical_bound_l2}:
\begin{align*}
    \left\Vert (\mathbb{P}_{n}-\mathbb{P})\hat{f} \right\Vert_q^q \lesssim n^{-\frac{q}{2}} \Vert \hat{f} \Vert_q^q.
\end{align*}
Therefore, the sample splitting lemma \citep[][Lemma 2]{kennedy2020sharp} is a special case of Lemma \ref{lem:empirical_bound_l2} when $q=2$.
\end{remark}

\begin{remark} \label{rmk:empirical_bound-2}
Note that the result of Lemma \ref{lem:empirical_bound_l2} also holds for a fixed, real-valued function $f$ without necessity of conditioning on the extra sample.
\end{remark}

The following lemma extends Lemma 2 of \citet{kennedy2020sharp} to a bootstrapped sample.

\begin{lemma}[Sample splitting with bootstrapped sample] \label{lem:sample-splitting-bootstrapped-sample}
Let $\mathbb{P}_{n}$ denote the empirical measure over a set $D_1^n=(Z_{1},\ldots,Z_{n})$,
which is i.i.d. from $\mathbb{P}$. Let $\mathbb{P}_{n}^{*}=\frac{1}{n}\sum_{i=1}^n \delta_{Z_i^*}$ be the bootstrap empirical distribution from the bootstrapped sample $Z_1^*, \ldots, Z_n^*$.
Let $\hat{f}$ be a sample operator constructed in a separate sample $D_2^m$ with size $m$ independent of $D_1^n$. Then, we have
\begin{align*}
    \left(\mathbb{P}_{n}^{*} - \mathbb{P}_{n} \right)\left(\hat{f} - f \right) = O_\Pb\left( \frac{\Vert \hat{f} - f \Vert}{\sqrt{n}} \right).
\end{align*}
\end{lemma}
\begin{proof}
Note that
\begin{align*}
    \left(\mathbb{P}_{n}^{*} - \mathbb{P}_{n} \right)\left(\hat{f} - f \right) = \frac{1}{n} \sum_{i=1}^n (M_i - 1)\left(\hat{f}(Z_i) - f(Z_i) \right)
\end{align*}
where $(M_1, \ldots, M_n)$ are multinomially distributed with parameters $n$ and probabilities $\left(\frac{1}{n}, \ldots ,\frac{1}{n}\right)$, independent of $Z_1, \ldots, Z_n$. Let $g=\hat{f} - f$. Then
\begin{align*}
  \Pb\left[ \left\{ \left(\mathbb{P}_{n}^{*} - \mathbb{P}_{n} \right)g(Z) \right\}^2\right]  &= \Pb\left[ \frac{1}{n^2}\sum_{i=1}^n (M_i - 1)^2 g^2(Z_i) +  \frac{1}{n^2}\sum_{1 \leq i,j \leq n}(M_i - 1)(M_j - 1)g(Z_i)g(Z_j) \right] \\
  &=  \Pb\left[ \frac{1}{n^2}\sum_{i=1}^n \E\left\{ (M_i - 1)^2 \mid D_1^n \right\} g^2(Z_i)\right]\\
  & \quad + \Pb\left[ \frac{1}{n^2}\sum_{1 \leq i,j \leq n}\E\left\{(M_i - 1)(M_j - 1)\mid D_1^n \right\}g(Z_i)g(Z_j) \right] \\
  &= \frac{n-1}{n^2} \Pb\left\{ g^2(Z) \right\} - \frac{n(n-1)}{n^3}\Pb\left\{g(Z_1)g(Z_2)\right\} \\
  & \leq \frac{2(n-1)}{n^2}\Vert g(Z)\Vert^2  \\
  & \leq  \frac{2}{n}\Vert g(Z)\Vert^2.
\end{align*}
Hence, using Markov inequality we have
\begin{align*}
    \Pb\left\{ \frac{\left\vert \left(\mathbb{P}_{n}^{*} - \mathbb{P}_{n} \right)\left(\hat{f} - f \right) \right\vert}{\left\Vert \hat{f} - f \right\Vert / \sqrt{n}} \geq t \right\} &= \E\left[ \Pb\left\{\frac{\left\vert \left(\mathbb{P}_{n}^{*} - \mathbb{P}_{n} \right)\left(\hat{f} - f \right) \right\vert}{\left\Vert \hat{f} - f \right\Vert / \sqrt{n}} \geq t \Biggm\vert  D_2^m \right\} \right] \\
    &\leq \frac{1}{t} \frac{\left\Vert \left(\mathbb{P}_{n}^{*} - \mathbb{P}_{n} \right)\left(\hat{f} - f \right) \right\Vert}{\left\Vert \hat{f} - f \right\Vert / \sqrt{n}} \\
    & \leq \frac{\sqrt{2}}{t},
\end{align*}
which follows by the iterated expectation and the Cauchy Schwarz inequality, thus yields the result.
\end{proof}

Next, we utilize the results from Theorem 3.5.1 and Theorem 3.5.4 in \cite{gine2021mathematical} in a simplified form adapted to our framework and formally present them as Theorem \ref{thm:empirical_bound_envelope}. We then extend Theorem \ref{thm:empirical_bound_envelope} to apply to a bootstrapped sample, as detailed in Theorem \ref{thm:empirical_bound_envelope_bootstrap}.

\begin{theorem}

\label{thm:empirical_bound_envelope}

Let $\mathbb{P}_{n}$ denote the empirical measure over a set $D_{1}^{n}=(Z_{1},\ldots,Z_{n})$,
which is i.i.d. from $\mathbb{P}$. Let $\mathcal{F}$ be a class
of measurable functions, and let $F$ be an envelop of $\mathcal{F}$.
If $\mathbb{P}F^{2}<\infty$ and $J(\mathcal{F},F,1)<\infty$, then

\[
\mathbb{E}\left[\sup_{f\in\mathcal{F}}\left|\sqrt{n}(\mathbb{P}_{n}-\mathbb{P})f\right|\right]\leq\left(8\sqrt{2}J(\mathcal{F},F,1)+2\right)\left\Vert F\right\Vert _{L^{2}(\mathbb{P})}.
\]

\end{theorem}

\begin{proof} Let $Z_{1}^{\prime},\ldots,Z_{n}^{\prime}$ be independent
copy of $Z_{1},\ldots,Z_{n}$, and let $\epsilon_{1},\ldots,\epsilon_{n}$
be Rademacher random variables independent of $Z_{1},\ldots,Z_{n},Z_{1}^{\prime},\ldots,Z_{n}^{\prime}$.
By Jensen's inequality, $\mathbb{E}\left[\left.\sup_{f\in\mathcal{F}}\left|(\mathbb{P}_{n}^{*}-\mathbb{P}_{n})f\right|\right|Z\right]$
is bounded as

\begin{align*}
\mathbb{E}\left[\sup_{f\in\mathcal{F}}\left|\sum_{i=1}^{n}f(Z_{i})-\mathbb{E}\left[f(Z)\right]\right|\right] & =\mathbb{E}\left[\sup_{f\in\mathcal{F}}\left|\sum_{i=1}^{n}f(Z_{i})-\mathbb{E}\left[\left.\sum_{i=1}^{n}f(Z_{i}^{\prime})\right|Z\right]\right|\right]\\
 & =\mathbb{E}\left[\sup_{f\in\mathcal{F}}\left|\mathbb{E}\left[\left.\sum_{i=1}^{n}f(Z_{i})-\sum_{i=1}^{n}f(Z_{i}^{\prime})\right|Z\right]\right|\right]\\
 & \leq\mathbb{E}\left[\sup_{f\in\mathcal{F}}\left|\sum_{i=1}^{n}\left(f(Z_{i})-f(Z_{i}^{\prime})\right)\right|\right].
\end{align*}
Then since $f(Z_{i})-f(Z_{i}^{\prime})$ has the same distribution
as $\epsilon_{i}\left(f(Z_{i})-f(Z_{i}^{\prime})\right)$, this is
further expanded and bounded as 
\begin{align}
\mathbb{E}\left[\sup_{f\in\mathcal{F}}\left|\sum_{i=1}^{n}f(Z_{i})-\mathbb{E}\left[f(Z)\right]\right|\right] & \leq\mathbb{E}\left[\sup_{f\in\mathcal{F}}\left|\sum_{i=1}^{n}\left(f(Z_{i})-f(Z_{i}^{\prime})\right)\right|\right]\nonumber \\
 & =\mathbb{E}\left[\sup_{f\in\mathcal{F}}\left|\sum_{i=1}^{n}\epsilon_{i}\left(f(Z_{i})-f(Z_{i}^{\prime})\right)\right|\right]\nonumber \\
 & \leq2\mathbb{E}\left[\sup_{f\in\mathcal{F}}\left|\sum_{i=1}^{n}\epsilon_{i}f(Z_{i})\right|\right].\label{eq:empirical_bound_envelope_rademacher}
\end{align}
Since the process $\sum_{i=1}^{n}a_{i}\epsilon_{i}$ for any fixed
$(a_{1},\ldots,a_{n})\in\mathbb{R}^{n}$ is seperable for Euclidean
distance, and is sub-Gaussian for this distance, the process $\left\{ \frac{1}{\sqrt{n}}\sum_{i=1}^{n}\epsilon_{i}f(Z_{i})\right\} _{f\in\mathcal{F}}$
is sub-Gaussian with respect to the $L^{2}(\mathbb{P}_{n})$ pseudo-distance
on $\mathcal{F}$ conditioned on the variables $Z_{i}$, and Theorem
2.3.7 of \cite{gine2021mathematical} can be applied. Set $\sigma_{n}^{2}\coloneqq\frac{1}{n}\sup_{f\in\mathcal{F}}\sum_{i=1}^{n}f^{2}(Z_{i})$,
then the diameter of $\mathcal{F}$ (with respect to $L^{2}(\mathbb{P}_{n})$
random pseudo-norm) is dominated by $2\sigma_{n}$. Thus, by noting
that 
\[
\frac{1}{n}\mathbb{E}\left[\left.\left(\sum_{i=1}^{n}\epsilon_{i}(f(Z_{i})-g(Z_{i}))\right)^{2}\right|Z\right]=\frac{1}{n}\sum_{i=1}^{n}(f-g)^{2}(Z_{i})=\left\Vert f-g\right\Vert _{L^{2}(\mathbb{P}_{n})}^{2},
\]
the entropy bound from Theorem 2.3.7 (b) of \cite{gine2021mathematical}
gives that, for any $f_{0}\in\mathcal{F}$, the RHS of \eqref{eq:empirical_bound_envelope_rademacher}
conditioned on $Z$ can be upper bounded as 
\[
\frac{1}{\sqrt{n}}\mathbb{E}\left[\left.\sup_{f\in\mathcal{F}}\left|\sum_{i=1}^{n}\epsilon_{i}f(Z_{i})\right|\right|Z\right]\leq\frac{1}{\sqrt{n}}\mathbb{E}\left[\left.\left|\sum_{i=1}^{n}\epsilon_{i}f_{0}(Z_{i})\right|\right|Z\right]+4\sqrt{2}\int_{0}^{\sigma_{n}}\sqrt{\log\mathcal{N}(\mathcal{F},L^{2}(\mathbb{P}_{n}),\tau)}d\tau,
\]
which implies that 
\begin{equation}
\frac{1}{\sqrt{n}}\mathbb{E}\left[\sup_{f\in\mathcal{F}}\left|\sum_{i=1}^{n}\epsilon_{i}f(Z_{i})\right|\right]\leq\frac{1}{\sqrt{n}}\mathbb{E}\left[\left|\sum_{i=1}^{n}\epsilon_{i}f_{0}(Z_{i})\right|\right]+4\sqrt{2}\mathbb{E}\left[\int_{0}^{\sigma_{n}}\sqrt{\log\mathcal{N}(\mathcal{F},L^{2}(\mathbb{P}_{n}),\tau)}d\tau\right].\label{eq:empirical_bound_envelope_integral}
\end{equation}
Now since $f_{0}\leq F$, the first term of \eqref{eq:empirical_bound_envelope_integral}
can be bounded as 
\begin{align}
\frac{1}{\sqrt{n}}\mathbb{E}\left[\left|\sum_{i=1}^{n}\epsilon_{i}f_{0}(Z_{i})\right|\right] & \leq\sqrt{\frac{1}{n}\mathbb{E}\left[\left(\sum_{i=1}^{n}\epsilon_{i}f_{0}(Z_{i})\right)^{2}\right]}=\left\Vert f_{0}\right\Vert _{L^{2}(\mathbb{P}_{n})}\nonumber \\
 & \leq\left\Vert F\right\Vert _{L^{2}(\mathbb{P}_{n})}.\label{eq:empirical_bound_envelope_integral_first}
\end{align}
And in the second term of \eqref{eq:empirical_bound_envelope_integral},
let us write $J(t)$ for $J(\mathcal{F},F,t)$ for convenience, then
the integral can be bounded as 
\begin{align*}
\int_{0}^{\sigma_{n}}\sqrt{\log2\mathcal{N}(\mathcal{F},L^{2}(\mathbb{P}_{n}),\tau)}d\tau & =\left\Vert F\right\Vert _{L^{2}(\mathbb{P}_{n})}\int_{0}^{\sigma_{n}/\left\Vert F\right\Vert _{L^{2}(\mathbb{P}_{n})}}\sqrt{\log\mathcal{N}(\mathcal{F},L^{2}(\mathbb{P}_{n}),\tau\left\Vert F\right\Vert _{L^{2}(\mathbb{P}_{n})})}d\tau\\
 & \leq\left\Vert F\right\Vert _{L^{2}(\mathbb{P}_{n})}J\left(\sigma_{n}/\left\Vert F\right\Vert _{L^{2}(\mathbb{P}_{n})}\right).
\end{align*}
Then by Fubini's theorem and the concavity of $(x,t)\mapsto\sqrt{t}J(\sqrt{x/t})$,
i.e., Lemma 3.5.3 (c) of \cite{gine2021mathematical}, we have 
\begin{align*}
\mathbb{E}\left[\int_{0}^{\sigma_{n}}\sqrt{\log2\mathcal{N}(\mathcal{F},L^{2}(\mathbb{P}_{n}),\tau)}d\tau\right] & \leq\mathbb{E}\left[\left\Vert F\right\Vert _{L^{2}(\mathbb{P}_{n})}J\left(\sigma_{n}/\left\Vert F\right\Vert _{L^{2}(\mathbb{P}_{n})}\right)\right]\\
 & \leq\mathbb{E}\left[\left\Vert F\right\Vert _{L^{2}(\mathbb{P}_{n})}\right]J\left(\mathbb{E}\left[\sigma_{n}\right]/\mathbb{E}\left[\left\Vert F\right\Vert _{L^{2}(\mathbb{P}_{n})})\right]\right)\\
 & =\left\Vert F\right\Vert _{L^{2}(\mathbb{P})}J\left(\left\Vert \sigma_{n}\right\Vert _{L^{2}(\mathbb{P})}/\left\Vert F\right\Vert _{L^{2}(\mathbb{P})}\right).
\end{align*}
And since $\left\Vert \sigma_{n}\right\Vert _{L^{2}(\mathbb{P})}\leq\left\Vert F\right\Vert _{L^{2}(\mathbb{P})}$
and $J$ is nondecreasing, 
\begin{align}
\mathbb{E}\left[\int_{0}^{\sigma_{n}}\sqrt{\log2\mathcal{N}(\mathcal{F},L^{2}(\mathbb{P}_{n}),\tau)}d\tau\right] & \leq\left\Vert F\right\Vert _{L^{2}(\mathbb{P})}J\left(\left\Vert \sigma_{n}\right\Vert _{L^{2}(\mathbb{P})}/\left\Vert F\right\Vert _{L^{2}(\mathbb{P})}\right)\nonumber \\
 & \leq\left\Vert F\right\Vert _{L^{2}(\mathbb{P}_{n})}J(1).\label{eq:empirical_bound_envelope_integral_second}
\end{align}
Hence applying \eqref{eq:empirical_bound_envelope_integral_first}
and \eqref{eq:empirical_bound_envelope_integral_second} to \eqref{eq:empirical_bound_envelope_integral}
give 
\begin{align*}
\frac{1}{\sqrt{n}}\mathbb{E}\left[\sup_{f\in\mathcal{F}}\left|\sum_{i=1}^{n}\epsilon_{i}f(Z_{i})\right|\right] & \leq\frac{1}{\sqrt{n}}\mathbb{E}\left[\left|\sum_{i=1}^{n}\epsilon_{i}f_{0}(Z_{i})\right|\right]+4\sqrt{2}\mathbb{E}\left[\int_{0}^{\sigma_{n}}\sqrt{\log\mathcal{N}(\mathcal{F},L^{2}(\mathbb{P}_{n}),\tau)}d\tau\right]\\
 & \leq\left(4\sqrt{2}J(1)+1\right)\left\Vert F\right\Vert _{L^{2}(\mathbb{P}_{n})}.
\end{align*}
And applying this back to \eqref{eq:empirical_bound_envelope_rademacher}
gives the desired upper bound for $\mathbb{E}\left[\sup_{f\in\mathcal{F}}\left|\sqrt{n}(\mathbb{P}_{n}-\mathbb{P})f\right|\right]$
as  
\begin{align*}
\mathbb{E}\left[\left.\sup_{f\in\mathcal{F}}\left|\sqrt{n}(\mathbb{P}_{n}-\mathbb{P})f\right|\right|Z\right] & \leq\frac{2}{\sqrt{n}}\mathbb{E}\left[\sup_{f\in\mathcal{F}}\left|\sum_{i=1}^{n}\epsilon_{i}f(Z_{i})\right|\right]\\
 & \leq\left(8\sqrt{2}J(\mathcal{F},F,1)+2\right)\left\Vert F\right\Vert _{L^{2}(\mathbb{P})}.
\end{align*}

\end{proof} 

\begin{theorem}
\label{thm:empirical_bound_envelope_bootstrap}
Let $\mathbb{P}_{n}$ denote the empirical measure over a set $D_{1}^{n}=(Z_{1},\ldots,Z_{n})$,
which is i.i.d. from $\mathbb{P}$. Let $\mathbb{P}_{n}^{*}=\frac{1}{n}\sum_{i=1}^{n}\delta_{Z_{i}^{*}}$
be the bootstrap empirical distribution from the bootstrapped sample
$Z_{1}^{*},\ldots,Z_{n}^{*}$. Let $\mathcal{F}$ be a class of measurable
functions, and let $F$ be an envelop of $\mathcal{F}$. If $\mathbb{P}F^{2}<\infty$
and $J(\mathcal{F},F,1)<\infty$, then 

\[
\mathbb{E}\left[\left.\sup_{f\in\mathcal{F}}\left|\sqrt{n}(\mathbb{P}_{n}^{*}-\mathbb{P}_{n})f\right|\right|Z\right]\leq\left(8\sqrt{2}J(\mathcal{F},F,1)+2\right)\left\Vert F\right\Vert _{L^{2}(\mathbb{P}_{n})}.
\]

\end{theorem}

\begin{proof}
Let $Z_{1}^{*\prime},\ldots,Z_{n}^{*\prime}$ be independent copy
of $Z_{1}^{*},\ldots,Z_{n}^{*}$, and let $\epsilon_{1},\ldots,\epsilon_{n}$
be Rademacher random variables independent of $Z_{1}^{*},\ldots,Z_{n}^{*},Z_{1}^{*\prime},\ldots,Z_{n}^{*\prime}$.
By Jensen's inequality,

\begin{align*}
\mathbb{E}\left[\left.\sup_{f\in\mathcal{F}}\left|\sum_{i=1}^{n}f(Z_{i}^{*})-\sum_{i=1}^{n}f(Z_{i})\right|\right|Z\right] & =\mathbb{E}\left[\left.\sup_{f\in\mathcal{F}}\left|\sum_{i=1}^{n}f(Z_{i}^{*})-\mathbb{E}\left[\sum_{i=1}^{n}f(Z_{i}^{*\prime})|Z^{*}\right]\right|\right|Z\right]\\
 & =\mathbb{E}\left[\left.\sup_{f\in\mathcal{F}}\left|\mathbb{E}\left[\sum_{i=1}^{n}f(Z_{i}^{*})-\sum_{i=1}^{n}f(Z_{i}^{*\prime})|Z^{*}\right]\right|\right|Z\right]\\
 & \leq\mathbb{E}\left[\left.\sup_{f\in\mathcal{F}}\left|\sum_{i=1}^{n}\left(f(Z_{i}^{*})-f(Z_{i}^{*\prime})\right)\right|\right|Z\right].
\end{align*}
Then since $f(Z_{i}^{*})-f(Z_{i}^{*\prime})$ has the same distribution
as $\epsilon_{i}\left(f(Z_{i}^{*})-f(Z_{i}^{*\prime})\right)$, this
is further expanded and bounded as 
\begin{align}
\mathbb{E}\left[\left.\sup_{f\in\mathcal{F}}\left|\sum_{i=1}^{n}f(Z_{i}^{*})-\sum_{i=1}^{n}f(Z_{i})\right|\right|Z\right] & \leq\mathbb{E}\left[\left.\sup_{f\in\mathcal{F}}\left|\sum_{i=1}^{n}\left(f(Z_{i}^{*})-f(Z_{i}^{*\prime})\right)\right|\right|Z\right]\nonumber \\
 & =\mathbb{E}\left[\left.\sup_{f\in\mathcal{F}}\left|\sum_{i=1}^{n}\epsilon_{i}\left(f(Z_{i}^{*})-f(Z_{i}^{*\prime})\right)\right|\right|Z\right]\nonumber \\
 & \leq2\mathbb{E}\left[\left.\sup_{f\in\mathcal{F}}\left|\sum_{i=1}^{n}\epsilon_{i}f(Z_{i}^{*})\right|\right|Z\right].\label{eq:empirical_bound_envelope_rademacher_bootstrap}
\end{align}
Since the process $\sum_{i=1}^{n}a_{i}\epsilon_{i}$ for any fixed
$(a_{1},\ldots,a_{n})\in\mathbb{R}^{n}$ is seperable for Euclidean
distance, and is sub-Gaussian for this distance, the process $\left\{ \frac{1}{\sqrt{n}}\sum_{i=1}^{n}\epsilon_{i}f(Z_{i}^{*})\right\} _{f\in\mathcal{F}}$
is sub-Gaussian with respect to the $L^{2}(\mathbb{P}_{n}^{*})$ pseudo-distance
on $\mathcal{F}$ conditioned on the variables $Z_{i}^{*}$, and Theorem
2.3.7 of \cite{gine2021mathematical} can be applied. Set $\sigma_{n}^{2}\coloneqq\frac{1}{n}\sup_{f\in\mathcal{F}}\sum_{i=1}^{n}f^{2}(Z_{i}^{*})$,
then the diameter of $\mathcal{F}$ (with respect to $L^{2}(\mathbb{P}_{n}^{*})$
random pseudo-norm) is dominated by $2\sigma_{n}$. Thus, by noting
that 
\[
\frac{1}{n}\mathbb{E}\left[\left.\left(\sum_{i=1}^{n}\epsilon_{i}(f(Z_{i}^{*})-g(Z_{i}^{*}))\right)^{2}\right|Z^{*}\right]=\frac{1}{n}\sum_{i=1}^{n}(f-g)^{2}(Z_{i}^{*})=\left\Vert f-g\right\Vert _{L^{2}(\mathbb{P}_{n}^{*})}^{2},
\]
the entropy bound from Theorem 2.3.7 (b) of \cite{gine2021mathematical}
gives that, for any $f_{0}\in\mathcal{F}$, the RHS of \eqref{eq:empirical_bound_envelope_rademacher_bootstrap}
conditioned on $Z$ can be upper bounded as
\[
\frac{1}{\sqrt{n}}\mathbb{E}\left[\left.\sup_{f\in\mathcal{F}}\left|\sum_{i=1}^{n}\epsilon_{i}f(Z_{i}^{*})\right|\right|Z^{*}\right]\leq\frac{1}{\sqrt{n}}\mathbb{E}\left[\left.\left|\sum_{i=1}^{n}\epsilon_{i}f_{0}(Z_{i}^{*})\right|\right|Z^{*}\right]+4\sqrt{2}\int_{0}^{\sigma_{n}}\sqrt{\log\mathcal{N}(\mathcal{F},L^{2}(\mathbb{P}_{n}^{*}),\tau)}d\tau,
\]
which implies that 
\begin{equation}
\frac{1}{\sqrt{n}}\mathbb{E}\left[\left.\sup_{f\in\mathcal{F}}\left|\sum_{i=1}^{n}\epsilon_{i}f(Z_{i}^{*})\right|\right|Z\right]\leq\frac{1}{\sqrt{n}}\mathbb{E}\left[\left.\left|\sum_{i=1}^{n}\epsilon_{i}f_{0}(Z_{i}^{*})\right|\right|Z\right]+4\sqrt{2}\mathbb{E}\left[\left.\int_{0}^{\sigma_{n}}\sqrt{\log\mathcal{N}(\mathcal{F},L^{2}(\mathbb{P}_{n}^{*}),\tau)}d\tau\right|Z\right].\label{eq:empirical_bound_envelope_integral_bootstrap}
\end{equation}
Now since $f_{0}\leq F$, the first term of \eqref{eq:empirical_bound_envelope_integral_bootstrap}
can be bounded as 
\begin{align}
\frac{1}{\sqrt{n}}\mathbb{E}\left[\left.\left|\sum_{i=1}^{n}\epsilon_{i}f_{0}(Z_{i}^{*})\right|\right|Z\right] & \leq\sqrt{\frac{1}{n}\mathbb{E}\left[\left.\left(\sum_{i=1}^{n}\epsilon_{i}f_{0}(Z_{i}^{*})\right)^{2}\right|Z\right]}=\left\Vert f_{0}\right\Vert _{L^{2}(\mathbb{P}_{n})}\nonumber \\
 & \leq\left\Vert F\right\Vert _{L^{2}(\mathbb{P}_{n})}.\label{eq:empirical_bound_envelope_integral_bootstrap_first}
\end{align}
And in the second term of \eqref{eq:empirical_bound_envelope_integral_bootstrap},
let us write $J(t)$ for $J(\mathcal{F},F,t)$ for convenience, then
the integral can be bounded as 
\begin{align*}
\int_{0}^{\sigma_{n}}\sqrt{\log2\mathcal{N}(\mathcal{F},L^{2}(\mathbb{P}_{n}^{*}),\tau)}d\tau & =\left\Vert F\right\Vert _{L^{2}(\mathbb{P}_{n}^{*})}\int_{0}^{\sigma_{n}/\left\Vert F\right\Vert _{L^{2}(\mathbb{P}_{n}^{*})}}\sqrt{\log\mathcal{N}(\mathcal{F},L^{2}(\mathbb{P}_{n}^{*}),\tau\left\Vert F\right\Vert _{L^{2}(\mathbb{P}_{n}^{*})})}d\tau\\
 & \leq\left\Vert F\right\Vert _{L^{2}(\mathbb{P}_{n}^{*})}J\left(\sigma_{n}/\left\Vert F\right\Vert _{L^{2}(\mathbb{P}_{n}^{*})}\right).
\end{align*}
Then by Fubini's theorem and the concavity of $(x,t)\mapsto\sqrt{t}J(\sqrt{x/t})$,
i.e., Lemma 3.5.3 (c) of \cite{gine2021mathematical}, we have 
\begin{align*}
\mathbb{E}\left[\left.\int_{0}^{\sigma_{n}}\sqrt{\log2\mathcal{N}(\mathcal{F},L^{2}(\mathbb{P}_{n}^{*}),\tau)}d\tau\right|Z\right] & \leq\mathbb{E}\left[\left.\left\Vert F\right\Vert _{L^{2}(\mathbb{P}_{n}^{*})}J\left(\sigma_{n}/\left\Vert F\right\Vert _{L^{2}(\mathbb{P}_{n}^{*})}\right)\right|Z\right]\\
 & \leq\mathbb{E}\left[\left.\left\Vert F\right\Vert _{L^{2}(\mathbb{P}_{n}^{*})}\right|Z\right]J\left(\mathbb{E}\left[\left.\sigma_{n}\right|Z\right]/\mathbb{E}\left[\left.\left\Vert F\right\Vert _{L^{2}(\mathbb{P}_{n}^{*})}\right|Z\right]\right)\\
 & \leq\left\Vert F\right\Vert _{L^{2}(\mathbb{P}_{n})}J\left(\left\Vert \sigma_{n}\right\Vert _{L^{2}(\mathbb{P}_{n})}/\left\Vert F\right\Vert _{L^{2}(\mathbb{P}_{n})}\right).
\end{align*}
And since $\left\Vert \sigma_{n}\right\Vert _{L^{2}(\mathbb{P}_{n})}\leq\left\Vert F\right\Vert _{L^{2}(\mathbb{P}_{n})}$
and $J$ is nondecreasing, 
\begin{align}
\mathbb{E}\left[\left.\int_{0}^{\sigma_{n}}\sqrt{\log2\mathcal{N}(\mathcal{F},L^{2}(\mathbb{P}_{n}^{*}),\tau)}d\tau\right|Z\right] & \leq\left\Vert F\right\Vert _{L^{2}(\mathbb{P}_{n})}J\left(\left\Vert \sigma_{n}\right\Vert _{L^{2}(\mathbb{P}_{n})}/\left\Vert F\right\Vert _{L^{2}(\mathbb{P}_{n})}\right)\nonumber \\
 & \leq\left\Vert F\right\Vert _{L^{2}(\mathbb{P}_{n})}J(1).\label{eq:empirical_bound_envelope_integral_bootstrap_second}
\end{align}
Hence applying \eqref{eq:empirical_bound_envelope_integral_bootstrap_first}
and \eqref{eq:empirical_bound_envelope_integral_bootstrap_second}
to \eqref{eq:empirical_bound_envelope_integral_bootstrap} give 
\begin{align*}
\frac{1}{\sqrt{n}}\mathbb{E}\left[\left.\sup_{f\in\mathcal{F}}\left|\sum_{i=1}^{n}\epsilon_{i}f(Z_{i}^{*})\right|\right|Z\right] & \leq\frac{1}{\sqrt{n}}\mathbb{E}\left[\left.\left|\sum_{i=1}^{n}\epsilon_{i}f_{0}(Z_{i}^{*})\right|\right|Z\right]+4\sqrt{2}\mathbb{E}\left[\left.\int_{0}^{\sigma_{n}}\sqrt{\log\mathcal{N}(\mathcal{F},L^{2}(\mathbb{P}_{n}^{*}),\tau)}d\tau\right|Z\right]\\
 & \leq\left(4\sqrt{2}J(1)+1\right)\left\Vert F\right\Vert _{L^{2}(\mathbb{P}_{n})},
\end{align*}
And applying this back to \eqref{eq:empirical_bound_envelope_rademacher_bootstrap}
gives the desired upper bound for $\mathbb{E}\left[\sup_{f\in\mathcal{F}}\left|\sqrt{n}(\mathbb{P}_{n}-\mathbb{P})f\right|\right]$
as 
\begin{align*}
\mathbb{E}\left[\left.\sup_{f\in\mathcal{F}}\left|\sqrt{n}(\mathbb{P}_{n}^{*}-\mathbb{P}_{n})f\right|\right|Z\right] & \leq\frac{2}{\sqrt{n}}\mathbb{E}\left[\left.\sup_{f\in\mathcal{F}}\left|\sum_{i=1}^{n}\epsilon_{i}f(Z_{i}^{*})\right|\right|Z\right]\\
 & \leq\left(8\sqrt{2}J(\mathcal{F},F,1)+2\right)\left\Vert F\right\Vert _{L^{2}(\mathbb{P}_{n})}.
\end{align*}

\end{proof}

\section{Proofs for Section \ref{sec:counterfactual-density-estimation}}

\subsection{Proof of Theorem \ref{thm:asymptotics-pa-hat}} 


We give the following lemma before proving Theorem \ref{thm:asymptotics-pa-hat}.

\begin{lemma} \label{lem:app-variance-bound-f-a_hy}
\textbf{(a)} It follows that
\[
\left\Vert \frac{\mathbbm{1}(A=a)}{{\pi}_{a}(X)}\left[K_{h,y}(Y)-{\mu}_{A,y}(X)\right]\right\Vert^{2}\leq\frac{1}{\varepsilon}\var\left(K_{h,y}(Y^{a})\right),
\]
and 
\[
\var\left({f}_{h,y}^{a}\right)\leq\left(\frac{1}{\varepsilon}+1\right)\var\left(K_{h,y}(Y^{a})\right).
\]

\textbf{(b)} Suppose $\left\Vert K\right\Vert _{\infty}<\infty$, then 
\[
\var\left(K_{h,y}(Y^{a})\right)\leq\frac{\left\Vert K\right\Vert _{\infty}^{2}}{h^{2d}},
\]
and 
\[
\var\left({f}_{h,y}^{a}\right)\leq\left\Vert K\right\Vert _{\infty}^{2}\left(\frac{1}{\varepsilon}+1\right)\frac{1}{h^{2d}},
\]

\textbf{(c)} 
Suppose $p_a(y) \leq p_{\max}$ for some $p_{\max} < \infty$, $\forall y \in \mathcal{Y}$. Then under Assumptions \ref{assumption:A1},
\[
\mathbb{E}\left|K_{h,y}(Y^{a})\right|^{q}\leq\frac{p_{\max}}{h^{d(q-1)}}\int|K(u)|^{q}du.
\]
In particular, 
\[
\var\left(K_{h,y}(Y^{a})\right)\leq\frac{c'_{a,1}}{h^{d}},
\]
and 
\begin{align*}
    \var\left({f}_{h,y}^{a} \right) \leq c'_{a,1}\left(\frac{1}{\varepsilon} + 1 \right)\frac{1}{h^d},
\end{align*}
for some positive constants $c'_{a,1}$ that depend only on $p_{\max}, \Vert K \Vert_{\infty}$.

\textbf{(d)} For every positive integer $q$,
\[
\mathbb{E}\left\vert {\mu}_{a,y}(X)\right\vert ^{q}\leq\mathbb{E}\left\{ \left\vert K_{h,y}(Y^{a})\right\vert ^{q}\right\} .
\]
\end{lemma}

\begin{proof}
(a) 
First, we note that 
\begin{align} \label{eqn:appendix-thm-3.1-1}
    \Pb\left\{ \frac{\mathbbm{1}(A=a)}{{\pi}^2_{a}(X)}\left(K_{h,y}(Y)-{\mu}_{A,y}(X)\right)^2\right\} &= \Pb\left\{ \frac{1}{\pi_a(X)} \Pb\left[ \left(K_{h,y}(Y)-{\mu}_{a,y}(X)\right)^2 \mid X,A=a\right] \right\} \nonumber \\
    &= \Pb\left\{ \frac{1}{\pi_a(X)} \var\left(K_{h,y}(Y)\mid X,A=a \right) \right\} \nonumber \\
    &= \Pb\left\{ \frac{1}{\pi_a(X)} \var\left(K_{h,y}(Y^a)\mid X \right) \right\} \nonumber \\
    & \leq\frac{1}{\varepsilon}\Pb\left\{ \var\left(K_{h,y}(Y^{a})\mid X\right)\right\}
    \nonumber \\
    & \leq\frac{1}{\varepsilon}\var\left(K_{h,y}(Y^{a})\right), 
\end{align}
where the second equality follows by the law of total expectation, the third by Assumptions \ref{assumption:c1}, \ref{assumption:c2}, and the fourth inequality by \ref{assumption:c3}.

Next, by Lemma \ref{lem:unbiasedness},
\begin{align}
    \var\left({f}_{h,y}^{a} \right) & = \Pb\left\{ \frac{\mathbbm{1}(A=a)}{{\pi}_{a}(X)}\left(K_{h,y}(Y)-{\mu}_{A,y}(X)\right)+{\mu}_{a,y}(X) - p_{a,h}(y) \right\}^2\nonumber \\
    &= \Pb\left\{ \frac{\mathbbm{1}(A=a)}{{\pi}^2_{a}(X)}\left(K_{h,y}(Y)-{\mu}_{A,y}(X)\right)^2 \right\} + \Pb\left\{\left({\mu}_{a,y}(X) - p_{a,h}(y)\right)^2 \right\} \label{eq:app-variance-bound-f-a_hy_expand}
\end{align}
which follows by the fact that $\E[\mu_{a,y}(X)] - p_{a,h} = 0$. 
The first term in \eqref{eq:app-variance-bound-f-a_hy_expand} is bounded by \eqref{eqn:appendix-thm-3.1-1}. For the second term, we get
\begin{align}
    \Pb\left\{ \left({\mu}_{a,y}(X) - p_{a,h}(y)\right)^2 \right\} &= \var\left(\E[K_{h,y}(Y) \mid X, A=a] \right)\nonumber \\
    &= \var\left(\E[K_{h,y}(Y^a) \mid X] \right)\nonumber \\
    & = \var\left(K_{h,y}(Y^a)\right).\label{eq:app-variance-bound-f-a_hy_second}
\end{align}
Hence we obtain
\[
\var\left({f}_{h,y}^{a}\right)\leq\left(\frac{1}{\varepsilon}+1\right)\var\left(K_{h,y}(Y^{a})\right).
\]

(b) When $\left\Vert K\right\Vert _{\infty}<\infty$, then we obtain
the following bound for $\var\left(K_{h,y}(Y^{a})\right)$:
\[
\var\left(K_{h,y}(Y^{a})\right)\leq\mathbb{E}\left[K_{h,y}^{2}(Y^{a})\right]\leq\frac{\left\Vert K\right\Vert _{\infty}^{2}}{h^{2d}}.
\]
Hence, applying this to (a) gives  
\[
\var\left({f}_{h,y}^{a}\right)\leq\left\Vert K\right\Vert _{\infty}^{2}\left(\frac{1}{\varepsilon}+1\right)\frac{1}{h^{2d}}.
\]

(c) By definition, we note that 
\begin{align*}
    \E\left\vert K_{h,y}(Y^a) \right\vert^q &= \frac{1}{h^{dq}} \int \left\vert K\left(\frac{z - y}{h}\right) \right\vert^q p_a(z) dz \\
    & \leq \frac{p_{\max}}{h^{d(q-1)}} \int \left\vert K(u)\right\vert^q du,
\end{align*}
which is finite under Assumption \ref{assumption:A1}. 

Also, since $p_{a}(y)\leq p_{\max}$ for all $y\in\mathcal{Y}$, under Assumption \ref{assumption:A1}, Proposition 1.1 of \cite{Tsybakov10} implies that 
\[
\var\left(K_{h,y}(Y^{a})\right)\leq\frac{c'_{a,1}}{h^{d}},
\]
where $c'_{a,1} = p_{\text{max}} \int K(u)^2 du$.
Applying this to (a) yields
\[
\var\left({f}_{h,y}^{a}\right)\leq c'_{a,1}\left(\frac{1}{\varepsilon}+1\right)\frac{1}{h^{d}}.
\]

(d) By Assumption \ref{assumption:c2} and Jensen's inequality, it follows that
\begin{align*}
    \E\left\vert {\mu}_{a,y}(X) \right\vert^q &= \E \left[ \left\vert \E\left\{ K_{h,y}(Y^a) \mid X \right\} \right\vert^q \right] \\ 
    & \leq \E \left[ \E\left\{ \left\vert K_{h,y}(Y^a) \right\vert^q \mid X \right\}  \right] \\
    & = \E \left\{ \left\vert K_{h,y}(Y^a) \right\vert^q \right\} \\
    & \leq \frac{p_{\max}}{h^{d(q-1)}} \int \left\vert K(u)\right\vert^q du.
\end{align*}

\end{proof}


The following claim aims to simplify the proof of Theorem \ref{thm:asymptotics-pa-hat}.
\begin{claim}

\label{claim:pa-hat_expand}

\textbf{(a)} The following decompositions hold:
\[
\widehat{p}_{a,h}(y)-p_{a,h}(y)=(\mathbb{P}_{n}-\mathbb{P})({f}_{h,y}^{a})+(\mathbb{P}_{n}-\mathbb{P})(\hat{f}_{h,y}^{a}-{f}_{h,y}^{a})+\mathbb{P}(\hat{f}_{h,y}^{a}-f_{h,y}^{a}),
\]
\[
\widehat{p}_{a,h}(y)-p_{a}(y)=(\mathbb{P}_{n}-\mathbb{P})({f}_{h,y}^{a})+(\mathbb{P}_{n}-\mathbb{P})(\hat{f}_{h,y}^{a}-{f}_{h,y}^{a})+\mathbb{P}(\hat{f}_{h,y}^{a}-f_{h,y}^{a})+p_{a,h}(y)-p_{a}(y).
\]

\textbf{(b)} $\hat{f}_{h,y}^{a}-{f}_{h,y}^{a}$ is further expanded as 
\begin{align*}
\hat{f}_{h,y}^{a}-{f}_{h,y}^{a} & =\left(\frac{{\pi}_{a}(X)-\widehat{\pi}_{a}(X)}{\widehat{\pi}_{a}(X)}\right)\frac{\mathbbm{1}(A=a)}{{\pi}_{a}(X)}\left\{ K_{h,y}(Y)-{\mu}_{A,y}(X)\right\} \\
 & -\mathbbm{1}(A=a)\frac{\widehat{\mu}_{A,y}(X)-{\mu}_{A,y}(X)}{\widehat{\pi}_{a}(X)}+\widehat{\mu}_{a,y}(X)-{\mu}_{a,y}(X),
\end{align*}
and can be bounded as 
\begin{align*}
 & \left|\hat{f}_{h,y}^{a}-{f}_{h,y}^{a}\right|\\
 & \leq\left|{\pi}_{a}(X)-\widehat{\pi}_{a}(X)\right|\left\Vert \frac{1}{\widehat{\pi}_{a}}\right\Vert _{\infty}\left|\frac{\mathbbm{1}(A=a)}{{\pi}_{a}(X)}\left\{ K_{h,y}(Y)-{\mu}_{A,y}(X)\right\} \right|+\left(1+\left\Vert \frac{1}{\widehat{\pi}_{a}}\right\Vert _{\infty}\right)\left|\widehat{\mu}_{a,y}(X)-{\mu}_{a,y}(X)\right|.
\end{align*}

\textbf{(c)} $\mathbb{P}(\hat{f}_{h,y}^{a}-f_{h,y}^{a})$ is bounded as 
\[
\left|\mathbb{P}(\hat{f}_{h,y}^{a}-f_{h,y}^{a})\right|\leq\left\Vert \frac{1}{\widehat{\pi}_{a}(X)}\right\Vert _{\infty}\left\Vert \widehat{\mu}_{a,y}(X)-{\mu}_{a,y}(X)\right\Vert \left\Vert \widehat{\pi}_{a}(X)-{\pi}_{a}(X)\right\Vert .
\]
\end{claim}

\begin{proof}
(a) It suffices to analyze $\widehat{p}_{a,h}(y)-p_{a,h}(y)$. By adding and subtracting terms, 
for all $a\in\mathcal{A}$ and $y\in\mathcal{Y}$, it is immediate to see
\begin{align*}
\widehat{p}_{a,h}(y)-p_{a,h}(y) & =\mathbb{P}_{n}\hat{f}_{h,y}^{a}-\mathbb{P}f_{h,y}^{a}\\
 & =(\mathbb{P}_{n}-\mathbb{P})({f}_{h,y}^{a})+(\mathbb{P}_{n}-\mathbb{P})(\hat{f}_{h,y}^{a}-{f}_{h,y}^{a})+\mathbb{P}(\hat{f}_{h,y}^{a}-f_{h,y}^{a}).
\end{align*}

(b) By definition, we have that
\begin{align*}
    & \hat{f}_{h,y}^{a}-{f}_{h,y}^{a} \\ 
    &= \left( \frac{{\pi}_{a}(X) - \widehat{\pi}_{a}(X)}{\widehat{\pi}_{a}(X)} \right)\left(\frac{\mathbbm{1}(A=a)}{{\pi}_{a}(X)}K_{h,y}(Y) \right) + \mathbbm{1}(A=a)\left\{ \frac{{\mu}_{A,y}(X)}{{\pi}_{a}(X)} - \frac{\widehat{\mu}_{A,y}(X)}{\widehat{\pi}_{a}(X)}\right\} + \widehat{\mu}_{a,y}(X) - {\mu}_{a,y}(X)\\
    &= \left( \frac{{\pi}_{a}(X) - \widehat{\pi}_{a}(X)}{\widehat{\pi}_{a}(X)} \right)\left(\frac{\mathbbm{1}(A=a)}{{\pi}_{a}(X)}K_{h,y}(Y) \right) \\
    & \quad + \mathbbm{1}(A=a)\left\{  \frac{{\mu}_{A,y}(X)}{\widehat{\pi}_{a}(X){\pi}_{a}(X)}(\widehat{\pi}_{a}(X)-{\pi}_{a}(X)) - \frac{\widehat{\mu}_{A,y}(X)-{\mu}_{A,y}(X)}{\widehat{\pi}_{a}(X)}\right\} + \widehat{\mu}_{a,y}(X) - {\mu}_{a,y}(X)\\
    &= \left( \frac{{\pi}_{a}(X) - \widehat{\pi}_{a}(X)}{\widehat{\pi}_{a}(X)} \right)\frac{\mathbbm{1}(A=a)}{{\pi}_{a}(X)} \left\{ K_{h,y}(Y) - {\mu}_{A,y}(X) \right\}   - \mathbbm{1}(A=a)\frac{\widehat{\mu}_{A,y}(X)-{\mu}_{A,y}(X)}{\widehat{\pi}_{a}(X)} + \widehat{\mu}_{a,y}(X) - {\mu}_{a,y}(X).
\end{align*}

From the above, it is immediate to obtain the bound:
\begin{align*}
 & \left|\hat{f}_{h,y}^{a}-{f}_{h,y}^{a}\right|\\
 & \leq\left|{\pi}_{a}(X)-\widehat{\pi}_{a}(X)\right|\left\Vert \frac{1}{\widehat{\pi}_{a}}\right\Vert _{\infty}\left|\frac{\mathbbm{1}(A=a)}{{\pi}_{a}(X)}\left\{ K_{h,y}(Y)-{\mu}_{A,y}(X)\right\} \right|+\left(1+\left\Vert \frac{1}{\widehat{\pi}_{a}}\right\Vert _{\infty}\right)\left|\widehat{\mu}_{a,y}(X)-{\mu}_{a,y}(X)\right|.
\end{align*}

(c) By adding and subtracting terms and the law of total expectation, it follows that
\begin{align}
& \Pb(\hat{f}_{h,y}^{a}-f_{h,y}^{a}) \nonumber \\
&= \Pb\left[\frac{\mathbbm{1}(A=a)}{\widehat{\pi}_{a}(X)}\left(K_{h,y}(Y)-\widehat{\mu}_{A,y}(X)\right)+\widehat{\mu}_{a,y}(X)-\frac{\mathbbm{1}(A=a)}{{\pi}_{a}(X)}\left(K_{h,y}(Y)-{\mu}_{A,y}(X)\right)-{\mu}_{a,y}(X)\right] \nonumber\\
 & =\Pb\left[\frac{\mathbbm{1}(A=a)K_{h,y}(Y)}{\widehat{\pi}_{a}(X){\pi}_{a}(X)}\left({\pi}_{a}(X)-\widehat{\pi}_{a}(X)\right) -\frac{\mathbbm{1}(A=a)}{\widehat{\pi}_{a}(X)}(\widehat{\mu}_{A,y}(X)-{\mu}_{A,y}(X)) \right. \nonumber\\
 &\qquad \ \ \left. -{\mu}_{A,y}(X)\mathbbm{1}(A=a)\frac{{\pi}_{a}(X)-\widehat{\pi}_{a}(X)}{\widehat{\pi}_{a}(X){\pi}_{a}(X)}+\widehat{\mu}_{a,y}(X)-{\mu}_{a,y}(X)\right] \nonumber\\
 & =\Pb\left[\frac{\mathbbm{1}(A=a)}{\widehat{\pi}_{a}(X){\pi}_{a}(X)}\left({\pi}_{a}(X)-\widehat{\pi}_{a}(X)\right)\left(K_{h,y}(Y)-{\mu}_{A,y}(X)\right)+(\widehat{\mu}_{a,y}(X)-{\mu}_{a,y}(X))\left(1-\frac{{\pi}_{a}(X)}{\widehat{\pi}_{a}(X)}\right)\right] \nonumber\\
 & =\Pb\left[\frac{\mathbbm{1}(A=a)}{\widehat{\pi}_{a}(X){\pi}_{a}(X)}\left({\pi}_{a}(X)-\widehat{\pi}_{a}(X)\right)\left({\mu}_{A,y}(X)-{\mu}_{A,y}(X)\right)+(\widehat{\mu}_{a,y}(X)-{\mu}_{a,y}(X))\frac{\left(\widehat{\pi}_{a}(X)-{\pi}_{a}(X)\right)}{\widehat{\pi}_{a}(X)}\right] \nonumber\\
 & =\Pb\left[(\widehat{\mu}_{a,y}(X)-{\mu}_{a,y}(X))\frac{\left(\widehat{\pi}_{a}(X)-{\pi}_{a}(X)\right)}{\widehat{\pi}_{a}(X)}\right] \nonumber\\
 & \leq \left\Vert \frac{1}{\widehat{\pi}_{a}(X)}\right\Vert _{\infty}\Vert\widehat{\mu}_{a,y}(X)-{\mu}_{a,y}(X)\Vert\Vert\widehat{\pi}_{a}(X)-{\pi}_{a}(X)\Vert, \label{eqn:asymptotics-pa-hat-ffhatdiff}
\end{align}
where the last inequality follows by the H\"{o}lder and Cauchy Schwarz inequality.
\end{proof}

Now we turn to the proof of Theorem \ref{thm:asymptotics-pa-hat}.
\begin{proof}
From Claim~\ref{claim:pa-hat_expand}~(a), for all $a\in\mathcal{A}$ and $y\in\mathcal{Y}$, $\widehat{p}_{a,h}(y)-p_{a,h}(y)$ is expanded as
\begin{equation}\label{eqn:app-pointwise_density_decompose}
\begin{aligned}
\widehat{p}_{a,h}(y)-p_{a,h}(y)&=\mathbb{P}_{n}\hat{f}_{h,y}^{a}-\mathbb{P}f_{h,y}^{a} \\
&= \underbrace{(\mathbb{P}_{n}-\mathbb{P})({f}_{h,y}^{a})}_{\text{(i)}} +\underbrace{ (\mathbb{P}_{n}-\mathbb{P})(\hat{f}_{h,y}^{a}-{f}_{h,y}^{a})}_{\text{ (ii)}}+\underbrace{\mathbb{P}(\hat{f}_{h,y}^{a}-f_{h,y}^{a})}_{\text{(iii)}}
\end{aligned}
\end{equation}
The term (i) in \eqref{eqn:app-pointwise_density_decompose} is a simple sample average of a fixed function and so will be asymptotically Gaussian by the central limit theorem. The term (iii) can be bounded using the result in Claim~\ref{claim:pa-hat_expand}~(c). In what follows, we analyze the term (ii).

By Claim~\ref{claim:pa-hat_expand}~(b), it follows that
\begin{align*}
\left\Vert \hat{f}_{h,y}^{a}-{f}_{h,y}^{a}\right\Vert  & \leq\left\Vert \left(\frac{{\pi}_{a}(X)-\widehat{\pi}_{a}(X)}{\widehat{\pi}_{a}(X)}\right)\frac{\mathbbm{1}(A=a)}{{\pi}_{a}(X)}\left\{ K_{h,y}(Y)-{\mu}_{A,y}(X)\right\} \right\Vert \\
 & \qquad+\left\Vert -\mathbbm{1}(A=a)\frac{\widehat{\mu}_{A,y}(X)-{\mu}_{A,y}(X)}{\widehat{\pi}_{a}(X)}+\widehat{\mu}_{a,y}(X)-{\mu}_{a,y}(X)\right\Vert.
\end{align*}
For the first term in the last display, we have
\begin{align*}
    & \left\Vert \left( \frac{{\pi}_{a}(X) - \widehat{\pi}_{a}(X)}{\widehat{\pi}_{a}(X)} \right)\frac{\mathbbm{1}(A=a)}{{\pi}_{a}(X)} \left[ K_{h,y}(Y) - {\mu}_{A,y}(X) \right]  \right\Vert \label{eqn:proof-thm3.1-1} \\
    &\leq  \left\Vert \frac{{\pi}_{a}(X) - \widehat{\pi}_{a}(X)}{\widehat{\pi}_{a}(X)} \right\Vert_\infty \left\Vert \frac{\mathbbm{1}(A=a)}{{\pi}_{a}(X)} \left[ K_{h,y}(Y) - {\mu}_{A,y}(X) \right]  \right\Vert \\
    &\leq \left\Vert {\pi}_{a}(X) - \widehat{\pi}_{a}(X) \right\Vert_\infty \left\Vert \frac{1}{\widehat{\pi}_{a}(X)} \right\Vert_\infty \sqrt{    
    \frac{1}{\varepsilon}\var\left(K_{h,y}(Y^{a})\right)
    }, 
\end{align*}
where the third inequality follows by Lemma~\ref{lem:app-variance-bound-f-a_hy}~(a). Consequently, we obtain that
\begin{align*}
 \left\Vert \hat{f}_{h,y}^{a}-{f}_{h,y}^{a}\right\Vert
 &
 \leq
 \left\Vert {\pi}_{a}(X) - \widehat{\pi}_{a}(X) \right\Vert_\infty\left\Vert \frac{1}{\widehat{\pi}_{a}(X)} \right\Vert_\infty \sqrt{
 \frac{1}{\varepsilon}\var\left(K_{h,y}(Y^{a})\right)
 } \\
 & \qquad+ \left(1 + \left\Vert \frac{1}{\widehat{\pi}_{a}(X)} \right\Vert_\infty\right) \left\Vert \widehat{\mu}_{a,y}(X)-{\mu}_{a,y}(X) \right\Vert.
\end{align*}

First consider the varying bandwidth case. By Claim~\ref{lem:app-variance-bound-f-a_hy}~(b), we have that  
\begin{align*}
 &
 \left\Vert \hat{f}_{h,y}^{a}-{f}_{h,y}^{a}\right\Vert
 \\
 &
 \leq
  \left\Vert {\pi}_{a}(X) - \widehat{\pi}_{a}(X) \right\Vert_\infty\left\Vert \frac{1}{\widehat{\pi}_{a}(X)} \right\Vert_\infty \sqrt{\frac{c'_{a,1}}{\varepsilon h^d}} + \left(1 + \left\Vert \frac{1}{\widehat{\pi}_{a}(X)} \right\Vert_\infty\right) \left\Vert \widehat{\mu}_{a,y}(X)-{\mu}_{a,y}(X) \right\Vert.
\end{align*}

Hence, by Lemma \ref{lem:empirical_bound_l2}, we conclude
\begin{align*}
    (\mathbb{P}_{n}-\mathbb{P})(\hat{f}_{h,y}^{a}-{f}_{h,y}^{a}) = O_{\Pb}\left(\frac{\left\Vert {\pi}_{a}(X) - \widehat{\pi}_{a}(X) \right\Vert_\infty}{\sqrt{nh^d}} + \frac{\left\Vert \widehat{\mu}_{a,y}(X)-{\mu}_{a,y}(X) \right\Vert}{\sqrt{n}}\right).
\end{align*}
Next, for the fixed-bandwidth case, $h$ is regarded as a constant. Hence, given that 
$\left\Vert K\right\Vert _{\infty}<\infty$, by Claim~\ref{claim:pa-hat_expand}~(b) it follows that
\begin{align*}
 & 
 \left\Vert \hat{f}_{h,y}^{a}-{f}_{h,y}^{a}\right\Vert
 \\
 &
 \leq
 \left\Vert {\pi}_{a}(X) - \widehat{\pi}_{a}(X) \right\Vert_\infty\left\Vert \frac{1}{\widehat{\pi}_{a}(X)} \right\Vert_\infty 
 \frac{\left\Vert K\right\Vert _{\infty}}{\sqrt{\epsilon}h^{d}}
 + \left(1 + \left\Vert \frac{1}{\widehat{\pi}_{a}(X)} \right\Vert_\infty\right) \left\Vert \widehat{\mu}_{a,y}(X)-{\mu}_{a,y}(X) \right\Vert,
\end{align*}
which gives
\begin{align*}
    (\mathbb{P}_{n}-\mathbb{P})(\hat{f}_{h,y}^{a}-{f}_{h,y}^{a}) = O_{\Pb}\left(\frac{\left(\left\Vert {\pi}_{a}(X) - \widehat{\pi}_{a}(X) \right\Vert + \left\Vert \widehat{\mu}_{a,y}(X)-{\mu}_{a,y}(X) \right\Vert \right)}{\sqrt{n}}\right).
\end{align*}

Then the desired results can be obtained by combining all the pieces into \eqref{eqn:app-pointwise_density_decompose}. 
\end{proof}

\subsection{Proof of Theorem \ref{thm:Lq-risk-pointwise}}

First, we give the following claim to simplify the proof of Theorem \ref{thm:Lq-risk-pointwise}.
\begin{claim}
\label{claim:Lq-risk_expand}
\textbf{(a)} For all $a\in\mathcal{A}$ and $y\in\mathcal{Y}$ and $q\geq1$,
$\left\Vert \widehat{p}_{a,h}(y)-p_{a}(y)\right\Vert _{q}$ and its
$q$-th power are bounded as 
\begin{align*}
&\left\Vert \widehat{p}_{a,h}(y)-p_{a}(y)\right\Vert _{q}\\
&\leq\left\Vert (\Pn-\Pb){f}_{h,y}^{a}\right\Vert _{q}+\left\Vert (\Pn-\Pb)(\hat{f}_{h,y}^{a}-{f}_{h,y}^{a})\right\Vert _{q}+\left\vert \Pb(\hat{f}_{h,y}^{a}-f_{h,y}^{a})\right\vert +\left\vert p_{h,a}(y)-p_{a}(y)\right\vert,
\end{align*}
and
\begin{align*}
& \left\Vert \widehat{p}_{a,h}(y)-p_{a}(y)\right\Vert _{q}^{q} \\ 
& \leq 4^{q-1}\left(\left\Vert (\Pn-\Pb){f}_{h,y}^{a}\right\Vert_{q}^{q}+\left\Vert (\Pn-\Pb)(\hat{f}_{h,y}^{a}-{f}_{h,y}^{a})\right\Vert _{q}^{q}+\left\vert \Pb(\hat{f}_{h,y}^{a}-f_{h,y}^{a})\right\vert ^{q}+\left\vert p_{h,a}(y)-p_{a}(y)\right\vert ^{q}\right),
\end{align*}
respectively.

\textbf{(b)} For all $a\in\mathcal{A}$ and $y\in\mathcal{Y}$ and $q\geq1$,
\[
\left\Vert \frac{\mathbbm{1}(A=a)}{{\pi}_{a}^{q}(X)}\left\{ K_{h,y}(Y)-{\mu}_{A,y}(X)\right\} \right\Vert _{q}^{q}\leq\left(\frac{2}{\epsilon}\right)^{q-1}\mathbb{E}\left\vert K_{h,y}(Y^{a})\right\vert ^{q}.
\]

\textbf{(c)} For all $a\in\mathcal{A}$ and $y\in\mathcal{Y}$ and $q\geq2$, 

\[
\left\Vert (\Pn-\Pb){f}_{h,y}^{a}\right\Vert _{q}^{q}\lesssim n^{1-q}\mathbb{E}\left[\left\vert K_{h,y}(Y^{a})\right\vert ^{q}\right]+n^{-\frac{q}{2}}\left\{ \var({f}_{h,y}^{a})\right\} ^{\frac{q}{2}},
\]
where all the constants depend only on $q\geq2$.
\end{claim}

\begin{proof}
(a) Claim~\ref{claim:pa-hat_expand}~(a) immediately gives
\[
\left\Vert \widehat{p}_{a,h}(y)-p_{a}(y)\right\Vert _{q}\leq\left\Vert (\Pn-\Pb){f}_{h,y}^{a}\right\Vert _{q}+\left\Vert (\Pn-\Pb)(\hat{f}_{h,y}^{a}-{f}_{h,y}^{a})\right\Vert _{q}+\left\vert \Pb(\hat{f}_{h,y}^{a}-f_{h,y}^{a})\right\vert +\left\vert p_{h,a}(y)-p_{a}(y)\right\vert .
\]
Then by H{\"o}lder's inequality, we have that 
\begin{align*}
& \left\Vert \widehat{p}_{a,h}(y)-p_{a}(y)\right\Vert _{q}^{q}\\
& \leq\left(\left\Vert (\Pn-\Pb){f}_{h,y}^{a}\right\Vert _{q}+\left\Vert (\Pn-\Pb)(\hat{f}_{h,y}^{a}-{f}_{h,y}^{a})\right\Vert _{q}+\left\vert \Pb(\hat{f}_{h,y}^{a}-f_{h,y}^{a})\right\vert +\left\vert p_{h,a}(y)-p_{a}(y)\right\vert \right)^{q}\\
 & \leq4^{q-1}\left(\left\Vert (\Pn-\Pb){f}_{h,y}^{a}\right\Vert _{q}^{q}+\left\Vert (\Pn-\Pb)(\hat{f}_{h,y}^{a}-{f}_{h,y}^{a})\right\Vert _{q}^{q}+\left\vert \Pb(\hat{f}_{h,y}^{a}-f_{h,y}^{a})\right\vert ^{q}+\left\vert p_{h,a}(y)-p_{a}(y)\right\vert ^{q}\right).
\end{align*}

(b) Note that
\begin{align*}
\left\Vert \frac{\mathbbm{1}(A=a)}{{\pi}_{a}^{q}(X)}\left\{ K_{h,y}(Y)-{\mu}_{A,y}(X)\right\} \right\Vert _{q}^{q} & =\E\left[\frac{\mathbbm{1}(A=a)}{{\pi}_{a}^{q}(X)}\left\vert K_{h,y}(Y)-{\mu}_{A,y}(X)\right\vert ^{q}\right]\\
 & =\E\left[\frac{1}{{\pi}_{a}^{q-1}(X)}\E\left\{ \left\vert K_{h,y}(Y)-{\mu}_{a,y}(X)\right\vert ^{q}\mid X,A=a\right\} \right]\\
 & \leq\frac{1}{\varepsilon^{q-1}}\E\left\{ \left\vert K_{h,y}(Y^{a})-{\mu}_{a,y}(X)\right\vert ^{q}\right\} \\
 & \leq\frac{2^{q}}{\varepsilon^{q-1}}\E\left\vert K_{h,y}(Y^{a})\right\vert ^{q},
\end{align*}
which follows by the iterated expectation, Minkowski inequality, and Lemma~\ref{lem:app-variance-bound-f-a_hy}~(d).

(c) By Lemma \ref{lem:empirical_bound_l2} (see Remark \ref{rmk:empirical_bound-2}), it follows for all $a\in\mathcal{A}$ and $y\in\mathcal{Y}$ that
\begin{align*}
\left\Vert (\Pn - \Pb){f}_{h,y}^{a} \right\Vert_q^q \lesssim  n^{1-q} \Vert {f}_{h,y}^{a} \Vert_q^q + n^{-\frac{q}{2}} \left\{\var({f}_{h,y}^{a})\right\}^{\frac{q}{2}},
\end{align*}
where all the constants depend only on $q \geq 2$. Then for the first term of the above inequality, we have
\begin{align*}
    \left\Vert {f}_{h,y}^{a} \right\Vert_q^q &= \E \left\vert \frac{\mathbbm{1}(A=a)}{{\pi}_{a}(X)}\left\{K_{h,y}(Y)-{\mu}_{A,y}(X)\right\}+{\mu}_{a,y}(X) \right\vert^q \\
    &         
    \leq 2^{q-1}    
    \left(\E \left[\frac{\mathbbm{1}(A=a)}{{\pi}_{a}^q(X)}\left\vert K_{h,y}(Y)-{\mu}_{A,y}(X)\right\vert^q \right] + \E\left\vert {\mu}_{a,y}(X) \right\vert^q\right),
\end{align*}
which follows by Minkowski inequality. 

Hence by combining (i) and Lemma~\ref{lem:app-variance-bound-f-a_hy}~(d), one may get
\[
    \left\Vert {f}_{h,y}^{a} \right\Vert_q^q     
    \leq\left(\frac{2^{2q-1}}{\varepsilon^{q-1}}+2^{q-1}\right)\mathbb{E}\left[\left\vert K_{h,y}(Y^{a})\right\vert ^{q}\right].    
\]
Then $\left\Vert (\Pn-\Pb){f}_{h,y}^{a}\right\Vert _{q}^{q}$ is correspondingly bounded as 
\[
\left\Vert (\Pn-\Pb){f}_{h,y}^{a}\right\Vert _{q}^{q}\lesssim n^{1-q}\mathbb{E}\left[\left\vert K_{h,y}(Y^{a})\right\vert ^{q}\right]+n^{-\frac{q}{2}}\left\{ \var({f}_{h,y}^{a})\right\} ^{\frac{q}{2}},
\]
where all the constants depend only on $q\geq2$.
\end{proof}

Based upon the above claim, we give the proof of Theorem \ref{thm:Lq-risk-pointwise} as below.
\begin{proof}

From 
Claim~\ref{claim:Lq-risk_expand}~(a), for all $a\in\mathcal{A}$ and $y \in \mathcal{Y}$ we have the following bound:
\begin{align*}
    & \left\Vert \widehat{p}_{a,h}(y)-p_{a}(y) \right\Vert_q^q \\
    &\lesssim \underbrace{\left\Vert (\Pn - \Pb){f}_{h,y}^{a} \right\Vert_q^q}_{(i)} + \underbrace{\left\Vert (\Pn - \Pb)(\hat{f}_{h,y}^{a}-{f}_{h,y}^{a}) \right\Vert_q^q}_{(ii)} + \underbrace{\left\vert \Pb(\hat{f}_{h,y}^{a}-f_{h,y}^{a}) \right\vert^q}_{(iii)} + \underbrace{\left\vert p_{h,a}(y) - p_{a}(y) \right\vert^q}_{(iv)}.
\end{align*}

In what follows, each term will be analyzed in turn.

\textbf{(i)} By 
Claim~\ref{claim:Lq-risk_expand}~(c), for all $a\in\mathcal{A}$ and $y\in\mathcal{Y}$,
it follows that 
\begin{align*}
\left\Vert (\Pn - \Pb){f}_{h,y}^{a} \right\Vert_q^q \lesssim  n^{1-q} 
\mathbb{E}\left[\left\vert K_{h,y}(Y^{a})\right\vert ^{q}\right]
+ n^{-\frac{q}{2}} \left\{\var({f}_{h,y}^{a})\right\}^{\frac{q}{2}}, 
\end{align*}
where all the constants depend only on $q \geq 2$. Then 
under Assumption~\ref{assumption:A1}
and the condition that $p_{a}(y)\leq p_{\max}$ for all $y\in\mathcal{Y}$, we have that
\begin{align*}
    \E\left\vert K_{h,y}(Y^a) \right\vert^q \leq \frac{p_{\max}}{h^{d(q-1)}} \int \left\vert K(u)\right\vert^q du.
\end{align*}

Further, by Lemma \ref{lem:app-variance-bound-f-a_hy} (c) we have that
$$
    \var\left({f}_{h,y}^{a} \right) \lesssim \frac{1}{h^d},
$$
where the constants depend on $\epsilon, p_{\max}, \int K(u)^2 du$. Hence,
\begin{align*}
    \left\{\var({f}_{h,y}^{a})\right\}^{\frac{q}{2}} \lesssim h^{-\frac{dq}{2}}.
\end{align*}

Putting the above pieces together, finally we have that  for all $a\in\mathcal{A}$ and $y\in\mathcal{Y}$,
\begin{align*}
    \left\Vert (\Pn - \Pb){f}_{h,y}^{a} \right\Vert_q^q \lesssim   n^{-(q-1)}h^{-d(q-1)} + n^{-\frac{q}{2}}h^{-\frac{dq}{2}}, \quad \forall y.
\end{align*}

\textbf{(ii)} From 
Claim~\ref{claim:pa-hat_expand}~(b), for all $a\in\mathcal{A}$ and $y\in\mathcal{Y}$, $\hat{f}_{h,y}^{a}-{f}_{h,y}^{a}$ can be expanded as 
\begin{align*}
    \hat{f}_{h,y}^{a}-{f}_{h,y}^{a} 
    &= \left( \frac{{\pi}_{a}(X) - \widehat{\pi}_{a}(X)}{\widehat{\pi}_{a}(X)} \right)\frac{\mathbbm{1}(A=a)}{{\pi}_{a}(X)} \left\{ K_{h,y}(Y) - {\mu}_{A,y}(X) \right\} \\
    & \qquad - \mathbbm{1}(A=a)\frac{\widehat{\mu}_{A,y}(X)-{\mu}_{A,y}(X)}{{\pi}_{a}(X)} + \widehat{\mu}_{a,y}(X) - {\mu}_{a,y}(X).
\end{align*}

Also, by Claim~\ref{claim:Lq-risk_expand}~(a) and Lemma~\ref{lem:app-variance-bound-f-a_hy}~(c),
under Assumption~\ref{assumption:A1} and the condition that $p_{a}(y)\leq p_{\max}$
for all $y\in\mathcal{Y}$, it follows that
\begin{align*}
    & \left\Vert \left( \frac{{\pi}_{a}(X) - \widehat{\pi}_{a}(X)}{\widehat{\pi}_{a}(X)} \right)\frac{\mathbbm{1}(A=a)}{{\pi}_{a}(X)} \left\{ K_{h,y}(Y) - {\mu}_{A,y}(X) \right\} \right\Vert_q^q \\
    & \leq \left\Vert \frac{1}{\widehat{\pi}_{a}(X)} \right\Vert_\infty^q \left\Vert \frac{\mathbbm{1}(A=a)}{{\pi}_{a}(X)} \left\{ K_{h,y}(Y) - {\mu}_{A,y}(X) \right\} \right\Vert_q^q \\
    & \leq\left(\frac{2}{\varepsilon}\right)^{q-1}\left\Vert \frac{1}{\widehat{\pi}_{a}(X)}\right\Vert _{\infty}^{q}\E\left\vert K_{h,y}(Y^{a})\right\vert ^{q}\\
    &     
    \lesssim \frac{1}{h^{(q-1)d}},    
\end{align*}
and that
\begin{align*}
    \var\left\{\left( \frac{{\pi}_{a}(X) - \widehat{\pi}_{a}(X)}{\widehat{\pi}_{a}(X)} \right)\frac{\mathbbm{1}(A=a)}{{\pi}_{a}(X)} \left\{ K_{h,y}(Y) - {\mu}_{A,y}(X) \right\}\right\} &\leq \left\Vert \frac{1}{\widehat{\pi}_{a}(X)} \right\Vert_\infty^2 \var\left\{ {f}_{h,y}^{a} \right\} \\ 
    & \lesssim \frac{1}{h^d},
\end{align*}
where all the constants only depend on $q, \epsilon, p_{\max}, \int K(u)^q du, \int K(u)^2 du, \left\Vert \frac{1}{\widehat{\pi}_{a}} \right\Vert_\infty^q$.
 
Hence, by Lemma \ref{lem:empirical_bound_l2}, we have that
\begin{align*}
    \left\Vert (\Pn - \Pb) \left[ \left( \frac{{\pi}_{a}(X) - \widehat{\pi}_{a}(X)}{\widehat{\pi}_{a}(X)} \right)\frac{\mathbbm{1}(A=a)}{{\pi}_{a}(X)} \left\{ K_{h,y}(Y) - {\mu}_{A,y}(X) \right\} \right] \right\Vert_q^q \lesssim   n^{-(q-1)}h^{-d(q-1)} + n^{-\frac{q}{2}}h^{-\frac{dq}{2}}, \quad \forall y,
\end{align*}
and that
\begin{align*}
    \left\Vert (\Pn - \Pb) \left[ \mathbbm{1}(A=a)\frac{\widehat{\mu}_{A,y}(X)-{\mu}_{A,y}(X)}{{\pi}_{a}(X)} - \left\{\widehat{\mu}_{a,y}(X) - {\mu}_{a,y}(X)\right\} \right] \right\Vert_q^q & \lesssim n^{-\frac{q}{2}} \left\Vert \widehat{\mu}_{a,y}(X) - {\mu}_{a,y}(X) \right\Vert_q^q, \quad \forall y.
\end{align*}

Thus we obtain that
\begin{align*}
    \left\Vert (\Pn - \Pb)(\hat{f}_{h,y}^{a}-{f}_{h,y}^{a}) \right\Vert_q^q \lesssim n^{-(q-1)}h^{-d(q-1)} + n^{-\frac{q}{2}}h^{-\frac{dq}{2}} + n^{-\frac{q}{2}} \left\Vert \widehat{\mu}_{a,y}(X) - {\mu}_{a,y}(X) \right\Vert_q^q, \quad \forall y.
\end{align*}

\textbf{(iii)} From Claim~\ref{claim:pa-hat_expand}~(c), it immediately follows that
\begin{align*}
    \left\vert \Pb(\hat{f}_{h,y}^{a}-f_{h,y}^{a}) \right\vert^q \lesssim \left\Vert\widehat{\mu}_{a,y}(X)-{\mu}_{a,y}(X)\right\Vert^q \left\Vert\widehat{\pi}_{a}(X)-{\pi}_{a}(X)\right\Vert^q, \quad \forall y.
\end{align*}

\textbf{(iv)} This part mimics the proof of Proposition 1.2 of \cite{Tsybakov10}. Since $p_a \in \mathcal{P}_\Sigma(\beta, L)$ and $K$ is a kernel of order $l$, using the multi-index notation, by Taylor's theorem one would obtain that $\forall y\in \mathcal{Y}$, 
\begin{align*}
    \left\vert p_{h,a}(y) - p_{a}(y) \right\vert  & \leq \int \vert K(u) \vert \underset{\substack{\vert \alpha \vert = l \\ \alpha \in {\mathbb{Z}_0^+}^d}}{\sum} \frac{l}{\alpha!} \Vert uh \Vert_1^\alpha \int_0^1 (1-\tau)^{(l-1)} \left\vert D^\alpha p_a(y + \tau uh) - D^\alpha p_a(y) \right\vert d\tau du \\
    & \leq \int \vert K(u) \vert \underset{\substack{\vert \alpha \vert = l \\ \alpha \in {\mathbb{Z}_0^+}^d}}{\sum} \frac{l}{\alpha!} \Vert uh \Vert_1^l \left\{\int_0^1 (1-\tau)^{(l-1)} L \Vert \tau uh \Vert_1^{\beta-l}d\tau \right\} du \\
    & \leq h^\beta \left(\underset{\substack{\vert \alpha \vert = l \\ \alpha \in {\mathbb{Z}_0^+}^d}}{\sum} \frac{L l}{\alpha!} \int \vert K(u) \vert \Vert u \Vert_1^\beta du \int_0^1 (1-\tau)^{(l-1)} \tau^{\beta-l} d\tau \right) 
\end{align*}
Hence, provided that $\int \vert K(u) \vert \Vert u \Vert_1^\beta du < \infty$, we have $\left\vert p_{h,a}(y) - p_{a}(y) \right\vert^q \lesssim h^{q\beta}$.

Putting together the above four pieces, we finally obtain that for all $a\in\mathcal{A}$ and $y\in\mathcal{Y}$, 
\begin{align*}
    & \left\Vert \widehat{p}_{a,h}(y)-p_{a}(y) \right\Vert_q^q \\
    & \lesssim n^{-(q-1)}h^{-d(q-1)} + n^{-\frac{q}{2}}h^{-\frac{dq}{2}} + h^{q\beta} + n^{-\frac{q}{2}} \left\Vert \widehat{\mu}_{a,y} - {\mu}_{a,y} \right\Vert_q^q + \left\Vert\widehat{\mu}_{a,y}-{\mu}_{a,y}\right\Vert^q \left\Vert\widehat{\pi}_{a}-{\pi}_{a}\right\Vert^q.
\end{align*}

\end{proof}


\begin{remark}[Remark \ref{rmk:new-boundedness-condition-on-kernel}] \label{rmk:thm3.2-p_max-representation} %
   When the assumptions of Theorem \ref{thm:Lq-risk-pointwise} are satisfied, we no longer require the condition $p_{\max} < \infty$ as it is implicitly derived by $\int \vert K(u) \vert \Vert u \Vert_1^\beta du < \infty$. Consider any kernel function $K$ satisfying $\int \vert K(u) \vert \Vert u \Vert_1^\beta du < \infty$.
   Then from the Taylor expansion, by letting $h=1$ we have 
   \begin{align*}
    p_a(y) &\leq \underset{\substack{\vert \alpha \vert = l \\ \alpha \in {\mathbb{Z}_0^+}^d}}{\sum} \frac{L l}{\alpha!} \int \vert K(u) \vert \Vert u \Vert_1^\beta du \int_0^1 (1-\tau)^{(l-1)} \tau^{\beta-l} d\tau + \int \left\vert K(z - y) \right\vert p_a(z) dz\\
    & \leq \underset{\substack{\vert \alpha \vert = l \\ \alpha \in {\mathbb{Z}_0^+}^d}}{\sum} \frac{\floor{\beta}{}L}{\alpha!} \int \vert K(u) \vert \Vert u \Vert_1^\beta du + \sup_u \left\vert K(u) \right\vert, 
   \end{align*}
   where the terms in the last display depend only on $L, \beta, K(\cdot)$. Hence, the constant $p_{\max}$ such that
   \begin{align*}
    \underset{y \in \R^d}{\sup} \underset{p_a \in \mathcal{P}_\Sigma(\beta,L)}{\sup} p_a(y) \leq p_{\max} < \infty
   \end{align*}
   can be characterized as
   \begin{align*}
       p_{\max} \equiv \underset{\substack{\vert \alpha \vert = l \\ \alpha \in {\mathbb{Z}_0^+}^d}}{\sum} \frac{\floor{\beta}{}L}{\alpha!} \int \vert K(u) \vert \Vert u \Vert_1^\beta du + \left\Vert K \right\Vert_\infty
   \end{align*}
   provided that $\int \vert K(u) \vert \Vert u \Vert_1^\beta du < \infty$.
\end{remark}

\subsection{Proof of Theorem \ref{thm:Lq-risk-integrated}}
\begin{proof}

By Fubini's Theorem and Claim~\ref{claim:Lq-risk_expand}~(a), for all $a\in\mathcal{A}$, we have
the following bound: 
\begin{align*}
    & \Pb\left\{ \int \left\vert \widehat{p}_{a,h}(y)-p_{a}(y) \right\vert^q dy \right\} \\
    &= \int \left\Vert \widehat{p}_{a,h}(y)-p_{a}(y) \right\Vert_q^q dy \\
    &         
    \leq 4^q    
    \int \left\{ \left\Vert (\Pn - \Pb){f}_{h,y}^{a} \right\Vert_q^q + \left\Vert (\Pn - \Pb)(\hat{f}_{h,y}^{a}-{f}_{h,y}^{a}) \right\Vert_q^q + \left\vert \Pb(\hat{f}_{h,y}^{a}-f_{h,y}^{a}) \right\vert^q + \left\vert p_{h,a}(y) - p_{a}(y) \right\vert^q \right\} dy.
\end{align*}



Parts (i) and (ii) of the proof of Theorem \ref{thm:Lq-risk-pointwise} indicate that for the first two terms, it suffices to show that
\begin{align*}
    \int \E\left\vert K_{h,y}(Y^a) \right\vert^q dy &= \frac{1}{h^{dq}} \int \int \left\vert K\left(\frac{z - y}{h}\right) \right\vert^q p_a(z) dz dy \\
    & = \frac{1}{h^{dq}} \int p_a(z)  \left[ \int \left\vert K\left(\frac{z - y}{h}\right) \right\vert^q dy \right] dz \\
    & \leq \frac{1}{h^{d(q-1)}} \int \left\vert K(u)\right\vert^q du,
\end{align*}
\begin{align*}
    \int \E\left\vert \mu_{a,y}(X) \right\vert^q dy & \leq \int \E\left\vert K_{h,y}(Y^a) \right\vert^q dy \\
    & \leq \frac{1}{h^{d(q-1)}} \int \left\vert K(u)\right\vert^q du,
\end{align*}
and that
\begin{align*}
    \int \left\{\var({f}_{h,y}^{a})\right\}^{\frac{q}{2}} dy &\leq \int \left\{\E[({f}_{h,y}^{a})^2]\right\}^{\frac{q}{2}} dy \\
    & \lesssim \int \left\{\E[( K_{h,y}(Y^a))^2]\right\}^{\frac{q}{2}} dy \\
    &= \int \left\{ \int \frac{1}{h^{2d}} K\left(\frac{z-y}{h}\right)^2 p_a(z) dz \right\}^{\frac{q}{2}} dy \\
    & = \int \left\{ \int \frac{1}{h^{d}} K\left(u\right)^2 p_a(y+hu) du \right\}^{\frac{q}{2}} dy \\
    & \leq \frac{1}{h^{\frac{dq}{2}}} \left( \int \left\{  \int \left\vert K\left(u\right)\right\vert^q p_a^{\frac{q}{2}}(y+hu) dy \right\}^{\frac{2}{q}} du \right)^{\frac{q}{2}} \\
    & = \frac{1}{h^{\frac{dq}{2}}} \left( \left\{ \int \left\vert K\left(u\right)\right\vert^2 du \right\} \left\{ \int p_a^{\frac{q}{2}}(v) dv \right\}^{\frac{2}{q}} \right)^{\frac{q}{2}} \\
    & \leq \frac{\Vert p_a \Vert_{L_\frac{q}{2}}^{\frac{q}{2}}}{h^{\frac{dq}{2}}} \left(\int \left\vert K\left(u\right)\right\vert^2 du\right)^{\frac{q}{2}},
\end{align*}
which eventually leads to 
\begin{align*}
    \int \left\Vert (\Pn - \Pb){f}_{h,y}^{a} \right\Vert_q^q dy  & \lesssim n^{-(q-1)} h^{-d(q-1)} + n^{-\frac{q}{2}} h^{-\frac{dq}{2}}, \\
    \int \left\Vert (\Pn - \Pb)(\hat{f}_{h,y}^{a}-{f}_{h,y}^{a}) \right\Vert_q^q dy & \lesssim n^{-(q-1)} h^{-d(q-1)} + n^{-\frac{q}{2}} h^{-\frac{dq}{2}} + n^{-\frac{q}{2}} \int \left\Vert \widehat{\mu}_{a,y}(X) - {\mu}_{a,y}(X) \right\Vert_q^q dy,
\end{align*}
provided that $\Vert p_a \Vert_{L_\frac{q}{2}} < \infty$.

For the third term, by Claim~\ref{claim:pa-hat_expand}~(c) we have that
\begin{align*}
    \int \left\vert \Pb(\hat{f}_{h,y}^{a}-f_{h,y}^{a}) \right\vert^q dy \lesssim \left\Vert\widehat{\pi}_{a}(X)-{\pi}_{a}(X)\right\Vert^q \int \left\Vert\widehat{\mu}_{a,y}(X)-{\mu}_{a,y}(X)\right\Vert^q dy .
\end{align*}

For the last term, write the Taylor expansion of $p_a$ at $y \in \mathcal{Y}$
\begin{align*}
    p_a(y + uh) &= \underset{\substack{\vert \alpha \vert \leq l-1 \\ \alpha \in {\mathbb{Z}_0^+}^d}}{\sum} \frac{D^\alpha p_a(y)}{\alpha!}(uh)^\alpha \\
    & \quad + \underset{\substack{\vert \alpha \vert = l \\ \alpha \in {\mathbb{Z}_0^+}^d}}{\sum} \frac{l}{\alpha!} (uh)^\alpha \left[ \int_0^1 (1-\tau)^{(l-1)} \left\{D^\alpha p_a(y + \tau uh) - D^\alpha p_a(y) \right\} d\tau \right].
\end{align*}

Since $p_a \in \mathcal{P}_{\mathcal{H}}(q,\beta, L)$ and $K$ is a kernel of order $l$, similarly as in the part (iv) of the proof of Theorem \ref{thm:Lq-risk-pointwise}, we obtain
\begin{align*}
    & \int \left\vert p_{h,a}(y) - p_{a}(y) \right\vert^q dy \\
    & \leq \int \left(\int \vert K(u) \vert l \underset{\substack{\vert \alpha \vert = l \\ \alpha \in {\mathbb{Z}_0^+}^d}}{\sum} \frac{\Vert uh \Vert_1^\alpha}{\alpha!} \int_0^1 (1-\tau)^{(l-1)} \left\vert D^\alpha p_a(y + \tau uh) - D^\alpha p_a(y) \right\vert d\tau du  \right)^q dy \\
    &  \leq \left[h^\beta \left(\underset{\substack{\vert \alpha \vert = l \\ \alpha \in {\mathbb{Z}_0^+}^d}}{\sum} \frac{L l}{\alpha!} \int \vert K(u) \vert \Vert u \Vert_1^\beta du \int_0^1 (1-\tau)^{(l-1)} \tau^{\beta-l} d\tau \right) \right]^q \\
    & \leq \left( \underset{\substack{\vert \alpha \vert = l \\ \alpha \in {\mathbb{Z}_0^+}^d}}{\sum} \frac{L \floor{\beta}{}}{\alpha!} \int \vert K(u) \vert \Vert u \Vert_1^\beta du \right)^q h^{q\beta},
\end{align*}
which follows by applying twice the Minkowski's integral inequality and using the fact that
\[
\left[ \int \left\vert D^\alpha p_a(y + \tau uh) - D^\alpha p_a(y) \right\vert^q dy \right]^{1/q} \leq L \Vert \tau uh \Vert_1^{\beta-l}, \quad \text{for }\alpha \text{ such that } \vert\alpha\vert=l.
\]

Hence, provided that $\int \vert K(u) \vert \Vert u \Vert_1^\beta du < \infty$,
\[
 \int \left\vert p_{h,a}(y) - p_{a}(y) \right\vert^q dy \lesssim h^{q\beta}.
\]

Putting this together, we finally obtain that
\begin{align*}
    \Pb\left\{ \int \left\vert \widehat{p}_{a,h}(y)-p_{a}(y) \right\vert^q dy \right\} & \lesssim n^{-(q-1)} h^{-d(q-1)} + n^{-\frac{q}{2}} h^{-\frac{dq}{2}} +  h^{q\beta} + n^{-\frac{q}{2}} \int \left\Vert \widehat{\mu}_{a,y}(X) - {\mu}_{a,y}(X) \right\Vert_q^q dy \\
    & \quad + \left\Vert\widehat{\pi}_{a}(X)-{\pi}_{a}(X)\right\Vert^q \int \left\Vert\widehat{\mu}_{a,y}(X)-{\mu}_{a,y}(X)\right\Vert^q dy.
\end{align*}

\end{proof}

\section{Proofs for Section \ref{sec:densiti-effects-estimator}}

\subsection{Proof of Theorem \ref{thm:psi-sm-counterfactual-density}}

\begin{proof}
By Lemma \ref{lem:distance_distribution_triangle},
\[
\left( D\left(\widehat{p}_{1,h},\widehat{p}_{0,h}\right)-D\left({p}_{1},{p}_{0}\right) \right)^2 \leq\left(D(\widehat{p}_{1,h},{p}_{1})+ D(\widehat{p}_{0,h},{p}_{0})\right)^2.
\]
and thus
\begin{align*}
\left\Vert D\left(\widehat{p}_{1,h},\widehat{p}_{0,h}\right)-D\left({p}_{1},{p}_{0}\right) \right\Vert &\leq \left\Vert D(\widehat{p}_{1,h},{p}_{1})+ D(\widehat{p}_{0,h},{p}_{0})\right\Vert \\
&\leq \left\Vert D(\widehat{p}_{1,h},{p}_{1}) \right\Vert+ \left\Vert D(\widehat{p}_{0,h},{p}_{0})\right\Vert.
\end{align*}

By Minkowski's integral inequality and 
Claim~\ref{claim:Lq-risk_expand}~(a), 
$\forall a$ it follows that 
\begin{align*}
    \left\Vert D(\widehat{p}_{a,h}, p_a) \right\Vert &= \left\{ \int \left( \int \left\vert \widehat{p}_{a,h}(y) - p_a(y) \right\vert dy \right)^2 d\Pb \right\}^{\frac{1}{2}} \\
    & \leq  \int \left\Vert \widehat{p}_{a,h}(y) - p_a(y) \right\Vert dy \\
    & \leq \int \left\{ \left\Vert (\Pn - \Pb){f}_{h,y}^{a} \right\Vert + \left\Vert (\Pn - \Pb)(\hat{f}_{h,y}^{a}-{f}_{h,y}^{a}) \right\Vert + \left\vert \Pb(\hat{f}_{h,y}^{a}-f_{h,y}^{a}) \right\vert + \left\vert p_{h,a}(y) - p_{a}(y) \right\vert \right\} dy.
\end{align*}

It suffices to analyze the first three terms.
Similarly as in the proof of Theorem \ref{thm:Lq-risk-integrated}, one may show that
\begin{align*}
    \int \left\{\E\left\vert K_{h,y}(Y^a) \right\vert^2\right\}^{\frac{1}{2}} dy &= \int \left[\frac{1}{h^{2d}}  \int \left\vert K\left(\frac{z - y}{h}\right) \right\vert^2 p_a(z) dz\right]^{\frac{1}{2}} dy \\
    &= h^{-\frac{d}{2}}\int \left[  \int \left\vert K\left(u\right) \right\vert^2 p_a(y + uh) du\right]^{\frac{1}{2}} dy \\
    &\leq h^{-\frac{d}{2}} \left(\int \left[  \int \left\vert K\left(u\right) \right\vert p_a^{\frac{1}{2}}(y + uh) dy\right]^2 du \right)^{\frac{1}{2}}\\
    &= h^{-\frac{d}{2}} \left(\int \left\vert K\left(u\right) \right\vert^2 \left[ \int  p_a^{\frac{1}{2}}(v) dv\right]^2 du \right)^{\frac{1}{2}}\\
    & = h^{-\frac{d}{2}} \left(\int \left\vert K\left(u\right) \right\vert^2 du \right)^{\frac{1}{2}} \int  p_a^{\frac{1}{2}}(v) dv \\
    & \leq h^{-\frac{d}{2}} \left(\int \left\vert K\left(u\right) \right\vert^2 du \right)^{\frac{1}{2}}  \\
    & \lesssim h^{-\frac{d}{2}},
\end{align*} 
and that
\begin{align*}
    \int \left\{\var({f}_{h,y}^{a})\right\}^{\frac{1}{2}} dy &\lesssim h^{-\frac{d}{2}}.
\end{align*}
Hence, by applying Lemma \ref{lem:empirical_bound_l2} (with $q=2$) twice, for the first two terms we obtain
\begin{align*}
    & \int \left\Vert (\Pn - \Pb){f}_{h,y}^{a} \right\Vert dy  \lesssim n^{-\frac{1}{2}} h^{-\frac{d}{2}}, \quad \text{and} \\
    & \int \left\Vert (\Pn - \Pb)(\hat{f}_{h,y}^{a}-{f}_{h,y}^{a}) \right\Vert dy \lesssim n^{-\frac{1}{2}} h^{-\frac{d}{2}} + n^{-\frac{1}{2}} \int \left\Vert \widehat{\mu}_{a,y}(X) - {\mu}_{a,y}(X) \right\Vert dy.
\end{align*}

For the third term, it is straightforward to see
\begin{align*}
    \int \left\vert \Pb(\hat{f}_{h,y}^{a}-f_{h,y}^{a}) \right\vert dy \lesssim \left\Vert\widehat{\pi}_{a}(X)-{\pi}_{a}(X)\right\Vert \int \left\Vert\widehat{\mu}_{a,y}(X)-{\mu}_{a,y}(X)\right\Vert dy.
\end{align*} 

Hence the result follows.    
    
\end{proof}




\subsection{Proof of Theorem \ref{thm:error-bound-psi-sm}}
\label{app:proof-error-psi-sm}

We first state several technical results.
To ease notation, here we suppress the superscript `$sm$' in $\widehat{\eta}^{sm},{\eta}^{sm}$. 




\begin{claim}

\label{claim:error-psi-sm-expand}

Let $\varphi_{s}^{sm}(\cdot;\eta)$ be the uncentered influence function of $\psi^{sm}_s$ such that $\phi_{s}^{sm}(\cdot;\eta)=\varphi_{s}^{sm}(\cdot;\eta)-\psi^{sm}_s$.
Then for two probability measures $\Pb$ and $\bar{\Pb}$ on $\mathcal{X}\times\mathcal{A}\times\mathcal{Y}$,
we have 
\begin{align*}
\bar{\Pb}\phi_{s}^{sm}(\cdot;\bar{\eta})-\Pb\phi_{s}^{sm}(\cdot;\eta) & =(\bar{\Pb}-\Pb)\varphi_{s}^{sm}(\cdot;\eta) + \bar{\Pb}\left\{ \varphi_{s}^{sm}(\cdot;\bar{\eta})-\varphi_{s}^{sm}(\cdot;\eta)\right\} \\
 & \qquad - \int_{\mathcal{Y}}h_{s}\left(\bar{p}_{1}(y)-\bar{p}_{0}(y)\right)dy+\int_{\mathcal{Y}}h_{s}\left(p_{1}(y)-p_{0}(y)\right)dy.
\end{align*}

\end{claim}

\begin{proof}
The result immediately follows by adding and subtracting terms:
\begin{align*}
 & \bar{\Pb}\phi_{s}^{sm}(\cdot;\bar{\eta})-\Pb\phi_{s}^{sm}(\cdot;\eta)\\
 & =\bar{\Pb}\varphi_{s}^{sm}(\cdot;\bar{\eta})-\int_{\mathcal{Y}}h_{s}\left(\bar{p}_{1}(y)-\bar{p}_{0}(y)\right)dy-\Pb\varphi_{s}^{sm}(\cdot;\eta)+\int_{\mathcal{Y}}h_{s}\left(p_{1}(y)-p_{0}(y)\right)dy\\
 & =(\bar{\Pb}-\Pb)\varphi_{s}^{sm}(\cdot;\eta)+\bar{\Pb}\left\{ \varphi_{s}^{sm}(\cdot;\bar{\eta})-\varphi_{s}^{sm}(\cdot;\eta)\right\} \\
 & \qquad -\int_{\mathcal{Y}}h_{s}\left(\bar{p}_{1}(y)-\bar{p}_{0}(y)\right)dy+\int_{\mathcal{Y}}h_{s}\left(p_{1}(y)-p_{0}(y)\right)dy.
\end{align*}

\end{proof}



\begin{lemma} \label{lem:sm-empirical-difference-measure} Let $\phi_{s}^{sm}(\cdot;{\eta})$ and $\phi_{s}^{sm}(\cdot;\bar{\eta})$ be the efficient influence functions for $\psi_{s}^{sm}$ with respect to different probability measures $\Pb$ and $\bar{\Pb}$ on $\mathcal{X}\times\mathcal{A}\times\mathcal{Y}$, respectively. Then, we have
\begin{align*}
\mathbb{P}_{n}\left\{ \phi_{s}^{sm}(Z;\bar{\eta})-\phi_{s}^{sm}(Z;\eta)\right\} &=O_\Pb\Bigg( \sum_a \left\{ \Vert h'_s\Vert_{\infty}\Vert \bar{\pi}_a - \pi_a \Vert \Vert \bar{\nu}_a - \nu_a \Vert + \Vert h''_s\Vert_{\infty}\Vert\bar{p}_a - p_a \Vert^2 \right\} \\
& \qquad \quad + \Vert h''_s\Vert_{\infty}\Vert\bar{p}_0 - p_0 \Vert \Vert\bar{p}_1 - p_1 \Vert  \\
& \qquad \quad + \frac{\Vert h'_s\Vert_{\infty}}{\sqrt{n}} \max_a\left\{ \left\Vert\frac{1}{\bar{\pi}_a}\right\Vert_{\infty} \right\} \left( \Vert\bar{\pi}_1 - \pi_1 \Vert + \max_a \left\{\Vert \bar{\nu}_a - \nu_a \Vert \right\} \right)\Bigg) \\
&\quad + \int_{\mathcal{Y}} h_s\left(p_1(y)-p_0(y)\right) dy  - \int_{\mathcal{Y}} h_s\left(\bar{p}_1(y)-\bar{p}_0(y)\right) dy.
\end{align*}
\end{lemma}
\begin{proof} Consider the following decomposition:
    \begin{align} \label{appeqn:sm-empirical-decomposition-1}
        \Pn\left\{ \phi_{s}^{sm}(Z;\bar{\eta})-\phi_{s}^{sm}(Z;\eta)\right\} &= (\Pn-\Pb)\left\{ \phi^{sm}_s(Z;\bar{\eta})  - \phi^{sm}_s(Z;\eta)  \right\} + \Pb\left\{ \phi^{sm}_s(Z;\bar{\eta}) - \phi_{s}^{sm}(Z;\eta) \right\}         
    \end{align}
For the first term in \eqref{appeqn:sm-empirical-decomposition-1}, we may show that
\begin{align*}
    \Vert \phi^{sm}_s(Z;\bar{\eta})  - \phi^{sm}_s(Z;\eta) \Vert & \lesssim \sum_a \left\{\left\Vert\frac{1}{\bar{\pi}_a}\right\Vert_{\infty} \Vert h'_s \Vert_{\infty}\left( \Vert \bar{\pi}_a - \pi_a \Vert + \Vert \bar{\nu}_a - \nu_a \Vert \right) + \Vert h'_s \Vert_{\infty}\Vert \bar{\nu}_a - \nu_a \Vert \right\}\\
    & \lesssim \Vert h'_s\Vert_{\infty} \max_a\left\{ \left\Vert\frac{1}{\bar{\pi}_a}\right\Vert_{\infty} \right\} \left( \Vert\bar{\pi}_1 - \pi_1 \Vert + \max_a \left\{\Vert \bar{\nu}_a - \nu_a \Vert \right\} \right),    
\end{align*}
which basically follows by the triangle and Cauchy-Schwarz inequalities. Hence, by Lemma \ref{lem:empirical_bound_l2}, we have that
\begin{align*}
    & (\Pn-\Pb)\left\{ \phi^{sm}_s(Z;\bar{\eta})  - \phi^{sm}_s(Z;\eta)  \right\} \\
    & = O_\Pb\left( \frac{\Vert h'_s\Vert_{\infty}}{\sqrt{n}} \max_a\left\{ \left\Vert\frac{1}{\bar{\pi}_a}\right\Vert_{\infty} \right\} \left( \Vert\bar{\pi}_1 - \pi_1 \Vert + \max_a \left\{\Vert \bar{\nu}_a - \nu_a \Vert \right\} \right) \right).
\end{align*}
Next, consider the von Mises expansion
\begin{align*}
    \psi^{sm}_s(\bar{\Pb}) - \psi^{sm}_s({\Pb}) = \int \phi^{sm}_s(Z;\bar{\Pb})d(\Pb-\bar{\Pb}) + R_2(\bar{\Pb}, \Pb).
\end{align*}
By \citet[][Lemma 1]{kennedy2021semiparametric}, $R_2(\bar{\Pb}, \Pb)$ is given by
\begin{align*}
    R_2(\bar{\Pb}, \Pb) &= \sum_a(2a-1)\Bigg\{\Pb\left[ \int h'_s\left(\bar{p}_1(y)-\bar{p}_0(y)\right)\left\{ \frac{\pi_a(X)}{\bar{\pi}_a(X)} - 1 \right\} \left\{ \nu_a(y\mid X) - \bar{\nu}_a(y\mid X) \right\}dy \right] \\
    & \qquad \qquad \quad \qquad + \int h'_s\left(\bar{p}_1(y)-\bar{p}_0(y)\right) \left\{{p}_a(y)-\bar{p}_a(y) \right\} dy \Bigg\} \\
    & \quad + \int \left\{ h_s\left(\bar{p}_1(y)-\bar{p}_0(y)\right) - h_s\left({p}_1(y)-{p}_0(y)\right)  \right\} dy \\
    &= \sum_a(2a-1)\Pb\left[ \int h'_s\left(\bar{p}_1(y)-\bar{p}_0(y)\right)\left\{ \frac{\pi_a(X)}{\bar{\pi}_a(X)} - 1 \right\} \left\{ \nu_a(y\mid X) - \bar{\nu}_a(y\mid X) \right\}dy \right] \\
    & \quad + \frac{1}{2}\int h''_s\left(\tilde{p}^*(y)\right)\left\{ \bar{p}_1(y) - p_1(y) - \bar{p}_0(y) + p_0(y) \right\}^2 dy,
\end{align*}
where the last line follows by a Taylor expansion with $\tilde{p}^*(y)$ lying between $\bar{p}_1(y) - \bar{p}_0(y)$ and $p_1(y)-p_0(y)$. Hence, for the second term in \eqref{appeqn:sm-empirical-decomposition-1} we obtain   
\begin{align*}
    & \Pb\left\{ \phi^{sm}_s(Z;\bar{\eta}) - \phi_{s}^{sm}(Z;\eta) \right\} = 
    \Pb\left\{ \phi^{sm}_s(Z;\bar{\eta}) \right\} \\
    & = \int_{\mathcal{Y}} h_s\left(p_1(y)-p_0(y)\right) dy  - \int_{\mathcal{Y}} h_s\left(\bar{p}_1(y)-\bar{p}_0(y)\right) dy\\
    & \quad + \sum_a(2a-1)\Pb\left[ \int h'_s\left(\bar{p}_1(y)-\bar{p}_0(y)\right)\left\{ \frac{\pi_a(X)}{\bar{\pi}_a(X)} - 1 \right\} \left\{ \nu_a(y\mid X) - \bar{\nu}_a(y\mid X) \right\}dy \right] \\
    & \quad + \frac{1}{2}\int h''_s\left(\tilde{p}^*(y)\right)\left[ \sum_a \left\{\bar{p}_a(y) - p_a(y) \right\}\right]^2 dy \\
    & = \int_{\mathcal{Y}} h_s\left(p_1(y)-p_0(y)\right) dy  - \int_{\mathcal{Y}} h_s\left(\bar{p}_1(y)-\bar{p}_0(y)\right) dy  \\
    & \quad + O_\Pb\Bigg( \sum_a \left\{ \Vert h'_s\Vert_{\infty}\Vert \bar{\pi}_a - \pi_a \Vert \Vert \bar{\nu}_a - \nu_a \Vert + \Vert h''_s\Vert_{\infty}\Vert\bar{p}_a - p_a \Vert^2 \right\} \\
    & \qquad \qquad + \Vert h''_s\Vert_{\infty}\Vert\bar{p}_0 - p_0 \Vert \Vert\bar{p}_1 - p_1 \Vert  \Bigg),
\end{align*}
where the first equality follows by definition and the fact that $\Pb\left\{\phi_{s}^{sm}(Z;\eta)\right\} = 0$, and the second by the Cauchy-Schwarz inequality.
Hence, the result follows.
\end{proof}

The following corollary immediately follows by Lemmas \ref{lem:sample-splitting-bootstrapped-sample} and \ref{lem:sm-empirical-difference-measure}.

\begin{corollary} \label{cor:sm-empirical-difference-measure}
Under the same conditions of Lemma \ref{lem:sm-empirical-difference-measure}, we have
\begin{align*}
\mathbb{P}_{n}^{*}\left\{ \phi_{s}^{sm}(Z;\bar{\eta})-\phi_{s}^{sm}(Z;\eta)\right\} &=O_\Pb\Bigg( \sum_a \left\{ \Vert h'_s\Vert_{\infty}\Vert \bar{\pi}_a - \pi_a \Vert \Vert \bar{\nu}_a - \nu_a \Vert + \Vert h''_s\Vert_{\infty}\Vert\bar{p}_a - p_a \Vert^2 \right\} \\
& \qquad \quad + \Vert h''_s\Vert_{\infty}\Vert\bar{p}_0 - p_0 \Vert \Vert\bar{p}_1 - p_1 \Vert  \\
& \qquad \quad + \frac{\Vert h'_s\Vert_{\infty}}{\sqrt{n}} \max_a\left\{ \left\Vert\frac{1}{\bar{\pi}_a}\right\Vert_{\infty} \right\} \left( \Vert\bar{\pi}_1 - \pi_1 \Vert + \max_a \left\{\Vert \bar{\nu}_a - \nu_a \Vert \right\} \right)\Bigg) \\
&\quad + \int_{\mathcal{Y}} h_s\left(p_1(y)-p_0(y)\right) dy  - \int_{\mathcal{Y}} h_s\left(\bar{p}_1(y)-\bar{p}_0(y)\right) dy.
\end{align*}
\end{corollary}

Now we are ready to prove Theorem \ref{thm:error-bound-psi-sm}.

\begin{proof}[Proof of Theorem \ref{thm:error-bound-psi-sm}]
By adding and subtracting terms, we have the following expansion:
\begin{align*}
    \widehat{\psi}^{sm}_s - \psi^{sm}_s  & = \Pn\{\phi^{sm}_s(Z;\widehat{\eta})\} + \int_{\mathcal{Y}} h_s\left(\widehat{p}_1(y)-\widehat{p}_0(y)\right) dy - \int_{\mathcal{Y}} h_s\left(p_1(y)-p_0(y)\right) dy \\
    & = (\Pn-\Pb)\phi^{sm}_s(Z;\eta) + \underbrace{(\Pn-\Pb)\left\{ \phi^{sm}_s(Z;\widehat{\eta})  - \phi^{sm}_s(Z;\eta)  \right\}}_{(i)} \\
    & \quad + \underbrace{\Pb\left\{ \phi^{sm}_s(Z;\widehat{\eta}) - \phi^{sm}_s(Z;\eta) \right\}  + \int_{\mathcal{Y}} h_s\left(\widehat{p}_1(y)-\widehat{p}_0(y)\right) dy - \int_{\mathcal{Y}} h_s\left(p_1(y)-p_0(y)\right) dy}_{(ii)}.
\end{align*}
By Lemma \ref{lem:sm-empirical-difference-measure} and Assumption \ref{assumption:A5}, we have that
\begin{align*}
   (i) + (ii) &= O_\Pb\Bigg( \sum_a \left\{ \Vert h'_s\Vert_{\infty}\Vert \bar{\pi}_a - \pi_a \Vert \Vert \bar{\nu}_a - \nu_a \Vert + \Vert h''_s\Vert_{\infty}\Vert\bar{p}_a - p_a \Vert^2 \right\} \\
& \qquad \quad + \Vert h''_s\Vert_{\infty}\Vert\bar{p}_0 - p_0 \Vert \Vert\bar{p}_1 - p_1 \Vert  \\
& \qquad \quad + \frac{\Vert h'_s\Vert_{\infty}}{\sqrt{n}} \max_a\left\{ \left\Vert\frac{1}{\widehat{\pi}_a}\right\Vert_{\infty} \right\} \left( \Vert\widehat{\pi}_1 - \pi_1 \Vert + \max_a \left\{\Vert \widehat{\nu}_a - \nu_a \Vert \right\} \right)\Bigg) \\
&= O_\Pb\Bigg( \sum_a \left\{ \Vert h'_s\Vert_{\infty}\Vert \bar{\pi}_a - \pi_a \Vert \Vert \bar{\nu}_a - \nu_a \Vert + \Vert h''_s\Vert_{\infty}\Vert\bar{p}_a - p_a \Vert^2 \right\} \\
& \qquad \quad + \Vert h''_s\Vert_{\infty}\Vert\bar{p}_0 - p_0 \Vert \Vert\bar{p}_1 - p_1 \Vert + o_\Pb\left( \frac{1}{\sqrt{n}} \right) \Bigg).
\end{align*}
This proves the first part of the theorem.
Next, from \eqref{eqn:EIF-psi-sm},
\begin{align*}
    \left\Vert \phi^{sm}_s(Z;\eta^{sm}) \right\Vert &\leq \left\Vert \frac{(2A-1)}{\pi_A(X)}\left\{h'_s(p_1(Y)-p_0(Y)) - \int h'_s(p_1(y)-p_0(y))\nu_A(y \mid X)dy\right\} \right\Vert  \\
    & \quad + \left\Vert \int h'_s(p_1(y)-p_0(y))\left\{\nu_1(y \mid X) - \nu_0(y \mid X) \right\}dy \right\Vert  \\
    & \quad \qquad + \left\vert \int\int h'_s(p_1(y)-p_0(y))\left\{\nu_1(y \mid x) - \nu_0(y \mid x) \right\}dy d\Pb(x) \right\vert \\
    & \lesssim \Vert h'_s \Vert_\infty.
\end{align*}
Hence, by Lemma \ref{lem:empirical_bound_l2} we have
\[
\left\Vert(\Pn-\Pb)\phi^{sm}_s(Z;\eta) \right\Vert = O\left( \frac{\Vert h'_s \Vert_\infty}{\sqrt{n}} \right).
\]


And consequently,
\begin{align*}
    \left\Vert  \widehat{\psi}^{sm}_s - \psi \right\Vert &= O_\Pb\Bigg(g(s) + \frac{\Vert h'_s \Vert_\infty}{\sqrt{n}} + \sum_a\left\{ \Vert h'_s\Vert_{\infty}\Vert \widehat{\pi}_a - \pi_a \Vert \Vert \widehat{\nu}_a - \nu_a \Vert + \Vert h''_s\Vert_{\infty}\Vert\widehat{p}_a - p_a \Vert^2 \right\} \\
    & \qquad \quad + \Vert h''_s\Vert_{\infty}\Vert\bar{p}_0 - p_0 \Vert \Vert\bar{p}_1 - p_1 \Vert+ o_\Pb\left( \frac{1}{\sqrt{n}} \right) \Bigg),
\end{align*}
which gives the second part of the result.
\end{proof}

\subsection{Proof of Theorem \ref{thm:error-bound-psi-mc}}
\label{app:proof-error-psi-mc}
We first give several technical results.
In the proof, for simplicity we suppress the superscript `$mc$' in $\widehat{\eta}^{mc},{\eta}^{mc}$, and let
\[
\psi^{+}=\int\left(p_{1}(y)-p_{0}(y)\right)\gamma^{+}dy,
\]
and 
\[
\widehat{\psi}^{+}=\mathbb{P}_{n}\{\underline{\varphi}(Z;\widehat{\gamma}^{+},\widehat{\eta})\}.
\]

\begin{claim}
\label{claim:error-psi-mc-expand}
Let $\underline{\varphi}(\cdot;\gamma,\eta)$ be the influence function defined in \eqref{eqn:etf-mc-component}, $\forall \gamma \in \{\gamma^+, \gamma^-\}$, and $\Pb$ and $\bar{\Pb}$ denote two different probability measures on $\mathcal{X}\times\mathcal{A}\times\mathcal{Y}$. Then, we have
\begin{align*}
\bar{\Pb}\underline{\varphi}(\cdot;\bar{\gamma},\bar{\eta})-\Pb\underline{\varphi}(\cdot;\gamma,\eta) & =(\bar{\Pb}-\Pb)\underline{\varphi}(\cdot;\gamma,\eta) + \bar{\Pb}\left\{ \underline{\varphi}(\cdot;\bar{\gamma},\bar{\eta})-\underline{\varphi}(\cdot;\gamma,\bar{\eta})\right\} \\
 & \quad + \bar{\Pb}\left\{ \underline{\varphi}(\cdot;\gamma,\bar{\eta})-\underline{\varphi}(\cdot;\gamma,\eta)\right\},
\end{align*}
and correspondingly, 
\begin{align*}
\bar{\Pb}\varphi^{mc}(\cdot;\bar{\gamma},\bar{\eta})-P\varphi^{mc}(\cdot;\gamma,\eta) & =(\bar{\Pb}-\Pb)\varphi^{mc}(\cdot;\gamma,\eta) + \bar{\Pb}\left\{ \varphi^{mc}(\cdot;\bar{\gamma},\bar{\eta})-\varphi^{mc}(\cdot;\gamma,\bar{\eta})\right\}\\
 & \quad + \bar{\Pb}\left\{ \varphi^{mc}(\cdot;\gamma,\bar{\eta})-\varphi^{mc}(\cdot;\gamma,\eta)\right\}.
\end{align*}

\end{claim}

\begin{proof}

It suffices to show, by adding and subtracting terms, that
\begin{align*}
 & \bar{\Pb}\underline{\varphi}(\cdot;\bar{\gamma},\bar{\eta})-\Pb\underline{\varphi}(\cdot;\gamma,\eta)\\
 & =(\bar{\Pb}-\Pb)\underline{\varphi}(\cdot;\gamma,\eta)+\bar{\Pb}\left\{ \underline{\varphi}(\cdot;\bar{\gamma},\bar{\eta})-\underline{\varphi}(\cdot;\gamma,\eta)\right\} \\
 & =(\bar{\Pb}-\Pb)\underline{\varphi}(\cdot;\gamma,\eta)+\bar{\Pb}\left\{ \underline{\varphi}(\cdot;\bar{\gamma},\bar{\eta})-\underline{\varphi}(\cdot;\gamma,\bar{\eta})\right\} +\bar{\Pb}\left\{ \underline{\varphi}(\cdot;\gamma,\bar{\eta})-\underline{\varphi}(\cdot;\gamma,\eta)\right\} .
\end{align*}

\end{proof}

\begin{lemma}\label{lem:mc-empirical-difference-measure-eta} For a fixed function $\gamma$, let $\varphi^{mc}(\cdot;\gamma,\eta)$ and $\varphi^{mc}(\cdot;\gamma,\bar{\eta})$ be the uncentered efficient influence functions for $\psi_{s}^{mc}$ with respect to different probability measures $\Pb$ and $\bar{\Pb}$, respectively. Then, we have
\begin{align*}
    \Pb\left\{\underline{\varphi}(Z;\gamma,\bar{\eta}) - \underline{\varphi}(Z;\gamma,\eta) \right\} \lesssim \sum_a \Vert \bar{\pi}_a - {\pi}_a \Vert \Vert \bar{\nu}_a - {\nu}_a \Vert.
\end{align*}    
\end{lemma}
\begin{proof}
We have
\begin{align*}
    & \Pb\left\{\underline{\varphi}(Z;\gamma,\bar{\eta}) - \underline{\varphi}(Z;\gamma,\eta) \right\} \\
    & = \sum_a \Bigg\{ \int \frac{\pi_a(x)}{\bar{\pi}_a(x)} \left[\int \gamma(y) \left(\nu_a(y \mid x) - \bar{\nu}_a(y \mid x) \right)dy \right] d\Pb(x) \\
    & \qquad \qquad + \int\int \gamma(y)\left( \bar{\nu}_a(y \mid x) - \nu_a(y \mid x)\right)dy d\Pb(x) \Bigg\} \\
    & = \sum_a \int \frac{\left\{ {\pi}_a(x) - \bar{\pi}_a(x)\right\}}{\bar{\pi}_a(x)} \left[\int \gamma(y)\left( \bar{\nu}_a(y \mid x) - \nu_a(y \mid x)\right)dy \right]  d\Pb(x) \\
    & \lesssim \sum_a \Vert \bar{\pi}_a - {\pi}_a \Vert \Vert \bar{\nu}_a - {\nu}_a \Vert,
\end{align*}
which follows by the iterated expectation and the fact that
\[
\Pb\left[ \frac{\mathbbm{1}(A=a)}{\pi_a(X)} \left( \gamma(Y) - \int \gamma(y)\nu_a(y \mid X)dy \right) \right] = 0.
\]
\end{proof}

\begin{lemma}\label{lem:mc-empirical-difference-measure-gamma} For any $\gamma \in \{\gamma^+, \gamma^-\}$, let $\varphi^{mc}(\cdot;\gamma,\bar{\eta})$ and $\varphi^{mc}(\cdot;\bar{\gamma},\bar{\eta})$ be the uncentered efficient influence functions for $\psi_{s}^{mc}$ where nuisance components $\gamma$ and $\{\bar{\gamma},\bar{\eta}\}$ are constructed on samples from different probability measures $\Pb$ and $\bar{\Pb}$, respectively. Then, we have
\begin{align*}
    & \Pb\left\{ \underline{\varphi}(Z;\bar{\gamma},\bar{\eta}) - \underline{\varphi}(Z;\gamma,\bar{\eta}) \right\} \\
    &\lesssim \sum_a \Vert \bar{\pi}_a - {\pi}_a \Vert \Vert \bar{\nu}_a - {\nu}_a \Vert + \left ( \Vert \bar{p}_1 - p_1 \Vert_{\infty} + \Vert \bar{p}_0 - p_0 \Vert_{\infty} \right)^{\alpha+1}.
\end{align*}
\end{lemma}
\begin{proof}
By the law of iterated expectation, it follows
\begin{align}
    & \Pb\left\{ \underline{\varphi}(Z;\bar{\gamma},\bar{\eta}) - \underline{\varphi}(Z;\gamma,\bar{\eta}) \right\} \nonumber \\
    &\leq \Pb\Bigg[ \frac{\pi_1(X)}{\bar{\pi}_1(X)}\left\{\int \bar{\gamma}(y) \nu_1(y \mid X)dy - \int \bar{\gamma}(y) \bar{\nu}_1(y \mid X)dy\right\}  \nonumber \\
    & \, \qquad -\frac{\pi_0(X)}{\bar{\pi}_0(X)}\left\{\int \bar{\gamma}(y) \nu_0(y \mid X)dy - \int \bar{\gamma}(y) \bar{\nu}_0(y \mid X)dy\right\}  \nonumber \\
    & \, \qquad + \int \left\{\bar{\gamma}(y) - \gamma(y) \right\} \nu_1(y \mid X)dy - \int \left\{\bar{\gamma}(y) - \gamma(y) \right\} \nu_0(y \mid X)dy \Bigg] \nonumber \\
    &\leq \Pb\Bigg[ \left\{\frac{\pi_1(X) - \bar{\pi}_1(X)}{\bar{\pi}_1(X)}\right\}\left\{\int \bar{\gamma}(y) \nu_1(y \mid X)dy - \int \bar{\gamma}(y) \bar{\nu}_1(y \mid X)dy\right\}  \nonumber\\
    & \, \qquad -\left\{\frac{\pi_0(X) - \bar{\pi}_0(X)}{\bar{\pi}_0(X)}\right\}\left\{\int \bar{\gamma}(y) \nu_0(y \mid X)dy - \int \bar{\gamma}(y) \bar{\nu}_0(y \mid X)dy\right\} \Bigg]  \nonumber \\
    & \quad + \Pb\Bigg[ \int \left\{\bar{\gamma}(y) - \gamma(y) \right\} \left\{\nu_1(y \mid X) - \nu_0(y \mid X) \right\}dy \Bigg]. \label{eqn:lemma:C3-proof-1}
\end{align}
Letting $\xi(y) = p_1(y) - p_0(y)$ and $p(y)$ be the density function of $Y$, the last term of the above display equals
\begin{align*}
    \int \xi(y) \left[\mathbbm{1}\left\{\bar{\xi}(y) >0 \right\} - \mathbbm{1}\left\{\xi(y) >0 \right\}\right] dy &\leq \int \left\vert \xi(y) \right\vert \mathbbm{1}\left\{\left\vert \xi(y) \right\vert \leq \left\vert \bar{\xi}(y) - \xi(y) \right\vert\right\} dy \\
    &\leq \int \left\vert \bar{\xi}(y) - \xi(y) \right\vert \mathbbm{1}\left\{\left\vert \xi(y) \right\vert \leq \left\vert \bar{\xi}(y) - \xi(y) \right\vert\right\} dy \\
    &\leq \frac{1}{p_{\min}} \int \left\vert \bar{\xi}(y) - \xi(y) \right\vert \mathbbm{1}\left\{\left\vert \xi(y) \right\vert \leq \left\vert \bar{\xi}(y) - \xi(y) \right\vert\right\} p(y) dy \\
    &\leq \frac{\Vert \bar{\xi} - \xi \Vert_{\infty}}{p_{\min}} \Pb\left\{ \left\vert \bar{\xi}(Y) \right\vert \leq \left\vert \bar{\xi}(Y) - \xi(Y) \right\vert \right\}\\
    &\lesssim \Vert \bar{\xi} - \xi \Vert_{\infty}^{\alpha+1},
\end{align*}
where the first inequality follows by \citet[][Lemma 1]{kennedy2020sharp}, the third by Cauchy-Schwarz along with our assumption that $\underset{y \in \mathcal{Y}}{\inf} \, p(y) \geq p_{\min} > 0$, and the last by the margin condition. Hence, the result follows.
\end{proof}

The proof of Theorem \ref{thm:error-bound-psi-mc} is provided below.
\begin{proof}[Proof of Theorem \ref{thm:error-bound-psi-mc}]
It suffices to consider the parameter
\[
\psi^+ = \int \left( p_1(y) - p_0(y) \right) \gamma^+(y) dy,
\]
and the estimator
\[
\widehat{\psi}^+ = \Pn\{\underline{\varphi}(Z;\widehat{\gamma}^{+},\widehat{\eta})\}.
\]
We first aim to show that 
\begin{align} \label{eqn:appendex-proof-thm-psi-mc-sub}
    \widehat{\psi}^+ - \psi^+  & = (\Pn-\Pb)\underline{\varphi}(Z;\gamma^{+},\eta) \nonumber \\
    & \quad + O_{\Pb}\left( \sum_a \left\{ \Vert \widehat{\pi}_a - \pi_a \Vert \Vert \widehat{\nu}_a - \nu_a \Vert  \right\} + \left(\Vert \widehat{p}_1 - p_1 \Vert_{\infty} + \Vert \widehat{p}_0 - p_0 \Vert_{\infty}\right)^{\alpha} +  o_{\Pb}\left(\frac{1}{\sqrt{n}}\right) \right).
\end{align}

By adding and subtracting terms, we have
\begin{align} \label{eqn:appendex-proof-thm-psi-mc-eq1}
    \widehat{\varphi}^+ - \psi^+  & =\Pn\{\underline{\varphi}(Z;\widehat{\gamma}^{+},\widehat{\eta})\} - \Pb\{\underline{\varphi}(Z;\gamma^{+},\eta)\} \nonumber \\
    &= (\Pn-\Pb)\underline{\varphi}(Z;\gamma^{+},\eta) + \underbrace{(\Pn-\Pb)\left\{\underline{\varphi}(Z;\widehat{\gamma}^{+},\widehat{\eta}) - \underline{\varphi}(Z;\gamma^{+},\widehat{\eta}) \right\}}_{(i)} \nonumber \\
    & \qquad + \underbrace{(\Pn-\Pb)\left\{\underline{\varphi}(Z;\gamma^{+},\widehat{\eta}) - \underline{\varphi}(Z;\gamma^{+},\eta) \right\}}_{(ii)} \nonumber \\
    & \qquad + \underbrace{\Pb\left\{ \underline{\varphi}(Z;\widehat{\gamma}^{+},\widehat{\eta}) - \underline{\varphi}(Z;\gamma^{+},\widehat{\eta}) \right\}}_{(iii)} + \underbrace{\Pb\left\{\underline{\varphi}(Z;\gamma^{+},\widehat{\eta}) - \underline{\varphi}(Z;\gamma^{+},\eta) \right\}}_{(iv)}.
\end{align}

Out of the five terms in \eqref{eqn:appendex-proof-thm-psi-mc-eq1}, the first term will be asymptotically Normal by the central limit theorem. The second and third terms ($i$ and $ii$) are empirical process terms, and will be order $o_{\Pb}(1/\sqrt{n})$ under some weak regularity conditions. 

By Lemmas \ref{lem:mc-empirical-difference-measure-gamma} and \ref{lem:mc-empirical-difference-measure-eta},
the terms $(iii)$ and $(iv)$ are bounded as follows:
\begin{align*}
    & \Pb\left\{ \underline{\varphi}(Z;\widehat{\gamma}^{+},\widehat{\eta}) - \underline{\varphi}(Z;\gamma^{+},\widehat{\eta}) \right\} \\
    &\lesssim \sum_a \Vert \widehat{\pi}_a - {\pi}_a \Vert \Vert \widehat{\nu}_a - {\nu}_a \Vert + \left ( \Vert \widehat{p}_1 - p_1 \Vert_{\infty} + \Vert \widehat{p}_0 - p_0 \Vert_{\infty} \right)^{\alpha+1}, \\
    & \Pb\left\{\underline{\varphi}(Z;\gamma^{+},\widehat{\eta}) - \underline{\varphi}(Z;\gamma^{+},\eta) \right\} 
    \lesssim \sum_a \Vert \widehat{\pi}_a - {\pi}_a \Vert \Vert \widehat{\nu}_a - {\nu}_a \Vert.
\end{align*}

Now we analyze the terms $(i)$ and $(ii)$.

\textbf{(i)} 
We will show that $\Vert \underline{\varphi}(Z;\widehat{\gamma}^{+},\widehat{\eta}) - \underline{\varphi}(Z;\gamma^{+},\widehat{\eta}) \Vert = o_{\Pb}(1)$. To this end, it suffices to show that $\forall a \in \mathcal{A}$
\begin{align*}
\left\Vert \frac{1}{\widehat{\pi}_a(X)}\left\{ \widehat{\gamma}^+(Y) - \gamma^+(Y) \right\} \right\Vert = o_{\Pb}(1), \, \text{and}
\end{align*}
\begin{align*}
\left\Vert \int \left\{\widehat{\gamma}^+(y) - \gamma^+(y) \right\} \widehat{\nu}_a(y \mid X) dy \right\Vert = o_{\Pb}(1).
\end{align*}

Letting $\xi(y) = p_1(y) - p_0(y)$, we have that
\begin{align*}
 \left\Vert \frac{1}{\widehat{\pi}_a(X)}\left\{ \widehat{\gamma}^+(Y) - \gamma^+(Y) \right\} \right\Vert &\leq  \left\Vert \frac{1}{\widehat{\pi}_a(X)} \right\Vert_\infty \sqrt{ \Pb \left\vert \widehat{\gamma}^+(Y) - \gamma^+(Y) \right\vert }, \\
& \lesssim \sqrt{ \Pb\left[  \mathbbm{1}\left\{\left\vert \xi(Y) \right\vert \leq \left\vert \widehat{\xi}(Y) - \xi(Y) \right\vert\right\} \right] } \\
& \lesssim \left( \Vert \widehat{p}_1 - p_1 \Vert_{\infty} + \Vert \widehat{p}_0 - p_0 \Vert_{\infty} \right)^{\frac{\alpha}{2}} 
\end{align*}
where follows by the fact that and that $\widehat{\gamma}^+, {\gamma}^+$ are indicator functions, \citet[][Lemma 1]{kennedy2020sharp}, and the margin condition. Letting $p(y)$ be the density of $Y$, we also have that
\begin{align*}
& \left( \int \left[\int \left\{\widehat{\gamma}^+(y) - \gamma^+(y) \right\} \widehat{\nu}_a(y \mid x) dy \right]^2 d\Pb(x) \right)^{\frac{1}{2}} \\
& \leq \left( \int \left[\int \left\{\widehat{\gamma}^+(y) - \gamma^+(y) \right\} {\nu}_a(y \mid x) dy \right]^2 d\Pb(x) \right)^{\frac{1}{2}} \\
& \quad + \left( \int \left[\int \left\{\widehat{\gamma}^+(y) - \gamma^+(y) \right\} \left\{\widehat{\nu}_a(y \mid x) - {\nu}_a(y \mid x) \right\} dy \right]^2 d\Pb(x) \right)^{\frac{1}{2}} \\
& \leq \int  \left[ \int \left\vert \left\{\widehat{\gamma}^+(y) - \gamma^+(y) \right\} {\nu}_a(y \mid x) \right\vert^2 d\Pb(x) \right]^{\frac{1}{2}} dy \\
& \quad + \int  \left[ \int \left\vert \left\{\widehat{\gamma}^+(y) - \gamma^+(y) \right\} \left\{\widehat{\nu}_a(y \mid x) - {\nu}_a(y \mid x) \right\} \right\vert^2 d\Pb(x) \right]^{\frac{1}{2}} dy \\
& \leq \int \left\vert \widehat{\gamma}^+(y) - \gamma^+(y) \right\vert  \left\{ \int  {\nu}_a^2(y \mid x) d\Pb(x) \right\}^{\frac{1}{2}} dy \\
& \quad + \int \left\vert \widehat{\gamma}^+(y) - \gamma^+(y) \right\vert \left[ \int \left\vert \widehat{\nu}_a(y \mid x) - {\nu}_a(y \mid x) \right\vert^2 d\Pb(x) \right]^{\frac{1}{2}} dy \\
& \leq \left[\frac{1}{p_{\min}}\int \mathbbm{1}\left\{\left\vert \xi(y) \right\vert \leq \left\vert \widehat{\xi}(y) - \xi(y) \right\vert\right\} p(y) dy\right]^{\frac{1}{2}} \\
& \, \quad \times \left[ \underset{y}{\sup} \left\{ \int  {\nu}_a^2(y \mid x) d\Pb(x) \right\}^{\frac{1}{2}} + \sqrt{\int \int \left\{ \widehat{\nu}_a(y \mid x) - {\nu}_a(y \mid x) \right\}^2 \Pb(x) dy } \right] \\
& \lesssim \left( \Vert \widehat{p}_1 - p_1 \Vert_{\infty} + \Vert \widehat{p}_0 - p_0 \Vert_{\infty} \right)^{\frac{\alpha}{2}} \left(\underset{y \in \mathcal{Y}}{\sup}\Vert {\nu}_a(y \mid X) \Vert + \Vert \widehat{\nu}_a - {\nu}_a \Vert \right) 
\end{align*}
where the first and second inequalities follows by the triangle and Minkowski's Integral Inequality, respectively, the fourth by the Cauchy–Schwarz inequality along with the lower bound condition $p(y) > p_{\min} > 0$, and the last by the margin condition.

Under the consistency condition in Assumption \ref{assumption:A6}, therefore, we obtain $\Vert \underline{\varphi}(Z;\widehat{\gamma}^{+},\widehat{\eta}) - \underline{\varphi}(Z;\gamma^{+},\widehat{\eta}) \Vert^2 = o_{\Pb}(1)$. Hence, the term $(i)$ is $o_{\Pb}(1/\sqrt{n})$ by Lemma \ref{lem:empirical_bound_l2}.

\textbf{(ii)} It follows that
\begin{align*}
    \Vert \underline{\varphi}(Z;\gamma^{+},\widehat{\eta}) - \underline{\varphi}(Z;\gamma^{+},\eta) \Vert \lesssim \sum_a \left( \Vert \widehat{\pi}_a - {\pi}_a \Vert + \Vert \widehat{\nu}_a - {\nu}_a \Vert \right),
\end{align*}
which is $o_{\Pb}(1)$ under Assumption \ref{assumption:A6}. Thus, the term $(ii)$ is $o_{\Pb}(1/\sqrt{n})$ by Lemma \ref{lem:empirical_bound_l2}.

Putting the four pieces together, we obtain the desired result in \eqref{eqn:appendex-proof-thm-psi-mc-sub}. 

Next, by Lemma \ref{lem:empirical_bound_l2} it immediately follows that 
\[
\Vert (\Pn-\Pb)\underline{\varphi}(Z;\gamma^{+},\eta) \Vert = O\left(\frac{1}{\sqrt{n}}\right).
\]

Moreover, since both $\widehat{\gamma}^{+},\widehat{\eta}$ are constructed in a separate sample from $\Pn$ and we know that
\begin{align*}
    & \left\Vert \underline{\varphi}(Z;\widehat{\gamma}^{+},\widehat{\eta}) - \underline{\varphi}(Z;\gamma^{+},\widehat{\eta}) \right\Vert = o_{\Pb}(1) \\
    & \left\Vert \underline{\varphi}(Z;\gamma^{+},\widehat{\eta}) - \underline{\varphi}(Z;\gamma^{+},\eta) \right\Vert = o_{\Pb}(1),
\end{align*}
by Lemma \ref{lem:empirical_bound_l2} we have
\begin{align*}
    \Vert (\Pn-\Pb)\left\{\underline{\varphi}(Z;\widehat{\gamma}^{+},\widehat{\eta}) - \underline{\varphi}(Z;\gamma^{+},\widehat{\eta}) \right\} \Vert & = o_\Pb\left( \frac{1}{\sqrt{n}} \right),\\
    \Vert (\Pn-\Pb)\left\{\underline{\varphi}(Z;\gamma^{+},\widehat{\eta}) - \underline{\varphi}(Z;\gamma^{+},\eta) \right\} \Vert & = o_\Pb\left( \frac{1}{\sqrt{n}} \right).
\end{align*}
Hence from \eqref{eqn:appendex-proof-thm-psi-mc-eq1}, we obtain the $L_2$ risk bound as
\begin{align*}
    \Vert \widehat{\psi}^+ - \psi^+  \Vert = 
    O_{\Pb}\left( \sum_a \left\{ \Vert \widehat{\pi}_a - \pi_a \Vert \Vert \widehat{\nu}_a - \nu_a \Vert  \right\} + \left(\Vert \widehat{p}_1 - p_1 \Vert_{\infty} + \Vert \widehat{p}_0 - p_0 \Vert_{\infty}\right)^{\alpha+1} + \frac{1}{\sqrt{n}} \right).
\end{align*}
\end{proof}

The following results are immediate consequences of Lemmas \ref{lem:mc-empirical-difference-measure-eta}, \ref{lem:mc-empirical-difference-measure-gamma}, and Theorem \ref{thm:error-bound-psi-mc}.
\begin{corollary}\label{cor:error-psi-mc-empirical-1}
    For a fixed function $\gamma$, let $\varphi^{mc}(\cdot;\gamma,\eta)$ and $\varphi^{mc}(\cdot;\gamma,\bar{\eta})$ be the uncentered efficient influence functions for $\psi_{s}^{mc}$ with respect to different probability measures $\Pb$ and $\bar{\Pb}$, respectively. Then, under the same conditions of Theorem \ref{thm:error-bound-psi-mc}, we have
\begin{align*}
    \Pn\left\{\underline{\varphi}(Z;\gamma,\bar{\eta}) - \underline{\varphi}(Z;\gamma,\eta) \right\} = O_\Pb\left(\sum_a \Vert \bar{\pi}_a - {\pi}_a \Vert \Vert \bar{\nu}_a - {\nu}_a \Vert\right) + o_{\Pb}\left(\frac{1}{\sqrt{n}}\right),
\end{align*}    
and
\begin{align*}
    \Pn^*\left\{\underline{\varphi}(Z;\gamma,\bar{\eta}) - \underline{\varphi}(Z;\gamma,\eta) \right\} = O_\Pb\left(\sum_a \Vert \bar{\pi}_a - {\pi}_a \Vert \Vert \bar{\nu}_a - {\nu}_a \Vert\right) + o_{\Pb}\left(\frac{1}{\sqrt{n}}\right).
\end{align*}    
\end{corollary}

\begin{corollary}\label{cor:error-psi-mc-empirical-2}
For any $\gamma \in \{\gamma^+, \gamma^-\}$, let $\varphi^{mc}(\cdot;\gamma,\bar{\eta})$ and $\varphi^{mc}(\cdot;\bar{\gamma},\bar{\eta})$ be the uncentered efficient influence functions for $\psi_{s}^{mc}$ where nuisance components $\gamma$ and $\{\bar{\gamma},\bar{\eta}\}$ are constructed on samples from different probability measures $\Pb$ and $\bar{\Pb}$, respectively. Then, under the same conditions of Theorem \ref{thm:error-bound-psi-mc}, we have
\begin{align*}
    & \Pn\left\{ \underline{\varphi}(Z;\bar{\gamma},\bar{\eta}) - \underline{\varphi}(Z;\gamma,\bar{\eta}) \right\} \\
    &= O_\Pb\left(\sum_a \Vert \bar{\pi}_a - {\pi}_a \Vert \Vert \bar{\nu}_a - {\nu}_a \Vert + \left ( \Vert \bar{p}_1 - p_1 \Vert_{\infty} + \Vert \bar{p}_0 - p_0 \Vert_{\infty} \right)^{\alpha+1}\right) + o_{\Pb}\left(\frac{1}{\sqrt{n}}\right),
\end{align*}
and
\begin{align*}
    & \Pn^*\left\{ \underline{\varphi}(Z;\bar{\gamma},\bar{\eta}) - \underline{\varphi}(Z;\gamma,\bar{\eta}) \right\} \\
    &= O_\Pb\left(\sum_a \Vert \bar{\pi}_a - {\pi}_a \Vert \Vert \bar{\nu}_a - {\nu}_a \Vert + \left ( \Vert \bar{p}_1 - p_1 \Vert_{\infty} + \Vert \bar{p}_0 - p_0 \Vert_{\infty} \right)^{\alpha+1}\right) + o_{\Pb}\left(\frac{1}{\sqrt{n}}\right).
\end{align*}
\end{corollary}
\begin{proof}
    The first result of Corollary \ref{cor:error-psi-mc-empirical-1} (Corollary \ref{cor:error-psi-mc-empirical-2}) follows by Lemma \ref{lem:mc-empirical-difference-measure-eta} (Lemma \ref{lem:mc-empirical-difference-measure-gamma}) and the same logic as used to analyze the term $(ii)$ (the term $(i)$) in the proof of Theorem \ref{thm:error-bound-psi-mc}. The second results follows by the first result, along with Lemma \ref{lem:sample-splitting-bootstrapped-sample} and the consistency condition in Assumption \ref{assumption:A6}.
\end{proof}

\section{Proofs for Section~\ref{sec:confidence_interval}}

\subsection{Proofs for Section \ref{subsec:confidence_band_for_density}}

Before proving Theorem \ref{thm:asymptotic_smooth_density}, we provide some technical results first.



\begin{claim}\label{claim:doubly_robust_vc}

(a) Recall the class $\mathcal{F}_{\bar{\eta},h}$ defined in \eqref{eq:notation_doubly_robust_class_general}.
Under Assumption \ref{assumption:kernel_vc}, $\mathcal{F}_{\bar{\eta},h}$ is of a uniformly bounded
VC-class, meaning that letting $\Updelta,\nu$ be the same constant used in Assumption~\ref{assumption:kernel_vc},
for every probability measure $Q$ on $\mathbb{R}^{d}$ and every $\epsilon\in\left(0,3\left\Vert K\right\Vert _{\infty}\frac{1}{\varepsilon h^d}\right)$,
the covering number $\mathcal{N}\left(\mathcal{F}_{\bar{\eta},h},L_{2}(Q),\epsilon\right)$
is upper bounded as
\[
\mathcal{N}\left(\mathcal{F}_{\bar{\eta},h},L_{2}(Q),\epsilon\right)\leq\left|\mathcal{A}\right|\left(\frac{3\Updelta\left\Vert K\right\Vert _{\infty}\left\Vert 1/\bar{\pi}_{a}\right\Vert _{\infty}}{\epsilon h^{d}}\right)^{\nu}.
\]
In particular, provided that $\left\Vert \frac{1}{\bar{\pi}_{a}}\right\Vert _{\infty}\leq\frac{1}{\varepsilon}$,
the covering number $\mathcal{N}\left(\mathcal{F}_{h},L_{2}(Q),\epsilon\right)$
is upper bounded as 
\[
\mathcal{N}\left(\mathcal{F}_{h},L_{2}(Q),\epsilon\right)\leq\left|\mathcal{A}\right|\left(\frac{3\Updelta \left\Vert K\right\Vert _{\infty}}{\varepsilon\epsilon h^{d}}\right)^{\nu}.
\]

(b) Recall the class $\bar{\mathcal{F}}_{h}$ defined in \eqref{eq:notation_doubly_robust_class_diff}.
Under Assumption~\ref{assumption:kernel_vc}, \ref{assumption:A2},
$\bar{\mathcal{F}}_{h}$ is of a uniformly bounded VC-class as 
\[
\mathcal{N}\left(\bar{\mathcal{F}}_{h},L_{2}(Q),\epsilon\right)\leq\left|\mathcal{A}\right|\left(\frac{6\Updelta\left\Vert K\right\Vert _{\infty}}{\varepsilon\epsilon h^{d}}\right)^{2\nu}.
\]

\end{claim}

\begin{proof}

(a) 
Fix a probability measure $Q$ on $\mathbb{R}^{d}$ and $\epsilon\in\left(0,3\left\Vert K\right\Vert _{\infty}\left\Vert 1/\pi_{a}\right\Vert _{\infty}h^{-d}\right)$.
For each $a\in\mathcal{A}$, consider a function subclass $\mathcal{F}_{\bar{\eta},h}^{a}\subset\mathcal{F}_{\bar{\eta},h}$
as 
\[
\mathcal{F}_{\bar{\eta},h}^{a}\coloneqq\left\{ f_{h,y}^{a}(\cdot;\bar{\eta}):\,y\in\mathcal{Y}_{h}\right\} ,
\]
then $\mathcal{F}_{\bar{\eta},h}=\bigcup_{a\in\mathcal{A}}\mathcal{F}_{\bar{\eta},h}^{a}$
and 
\begin{equation}
\mathcal{N}\left(\mathcal{F}_{\bar{\eta},h},L_{2}(Q),\epsilon\right)\leq\sum_{a\in\mathcal{A}}\mathcal{N}\left(\mathcal{F}_{\bar{\eta},h}^{a},L_{2}(Q),\epsilon\right).\label{eq:doubly_robust_vc_union_bound}
\end{equation}
Now we bound each $\mathcal{N}\left(\mathcal{F}_{\bar{\eta},h}^{a},L_{2}(Q),\epsilon\right)$,
$\forall a\in\mathcal{A}$. For any $y_{1},y_{2}\in\mathcal{Y}_{h}$,
\begin{align*}
\left|f_{h,y_{1}}^{a}(x',a',y';\bar{\eta})-f_{h,y_{2}}^{a}(x',a',y';\bar{\eta})\right| & =\left|\frac{1(a'=a)}{\bar{\pi}_{a}(x')}(K_{h,y_{1}}(y')-K_{h,y_{2}}(y'))\right|\\
 & \leq\left\Vert 1/\bar{\pi}_{a}\right\Vert _{\infty}\left|K_{h,y_{1}}(y')-K_{h,y_{2}}(y')\right|.
\end{align*}
Hence for any probability measure $Q$ on $\mathcal{X}\times\mathcal{A}\times\mathcal{Y}$
we have 
\[
\left\Vert f_{h,y_{1}}^{a}-f_{h,y_{2}}^{a}\right\Vert _{L_{2}(Q)}\leq\left\Vert 1/\bar{\pi}_{a}\right\Vert _{\infty}\left\Vert K_{h,y_{1}}-K_{h,y_{2}}\right\Vert _{L_{2}(\Pi_{\mathcal{Y}}Q)},
\]
where $\Pi_{\mathcal{Y}}(Q)$ is the marginal probability distribution
of $Q$ when projected on $\mathcal{Y}$. Hence this implies an upper
bound on $\mathcal{N}\left(\mathcal{F}_{\bar{\eta},h}^{a},L_{2}(Q),\epsilon\right)$
as 
\begin{equation}
\mathcal{N}\left(\mathcal{F}_{\bar{\eta},h}^{a},L_{2}(Q),\epsilon\right)\leq\mathcal{N}\left(\mathcal{G}_{h},L_{2}(\Pi_{\mathcal{Y}}Q),\frac{\epsilon}{\left\Vert 1/\bar{\pi}_{a}\right\Vert _{\infty}}\right).\label{eq:doubly_robust_vc_classes_bound}
\end{equation}
Then Assumption~\ref{assumption:kernel_vc} gives an upper bound
on $\mathcal{N}\left(\mathcal{G}_{h},L_{2}(\Pi_{\mathcal{Y}}Q),\epsilon/\left\Vert 1/\bar{\pi}_{a}\right\Vert _{\infty}\right)$
as 
\begin{equation}
\mathcal{N}\left(\mathcal{G}_{h},L_{2}(\Pi_{\mathcal{Y}}Q),\frac{\epsilon}{\left\Vert 1/\bar{\pi}_{a}\right\Vert _{\infty}}\right)\leq\left(\frac{3\Updelta\left\Vert K\right\Vert _{\infty}\left\Vert 1/\bar{\pi}_{a}\right\Vert _{\infty}}{\epsilon h^{d}}\right)^{\nu},\label{eq:doubly_robust_vc_uniform_bound_condition}
\end{equation}
where the additional constant $3$ in the above display comes from
the fact that $\epsilon/\left\Vert 1/\bar{\pi}_{a}\right\Vert _{\infty}$
ranges from $0$ to $3\left\Vert K\right\Vert _{\infty}h^{-d}$. Hence,
combining \eqref{eq:doubly_robust_vc_classes_bound} and \eqref{eq:doubly_robust_vc_uniform_bound_condition}
gives the uniformly bounded VC dimension of $\mathcal{N}\left(\mathcal{F}_{\bar{\eta},h}^{a},L_{2}(Q),\epsilon\right)$
as 
\[
\mathcal{N}\left(\mathcal{F}_{\bar{\eta},h}^{a},L_{2}(Q),\epsilon\right)\leq\left(\frac{3\Updelta\left\Vert K\right\Vert _{\infty}\left\Vert 1/\bar{\pi}_{a}\right\Vert _{\infty}}{\epsilon h^{d}}\right)^{\nu}.
\]
Finally, applying this to \eqref{eq:doubly_robust_vc_union_bound}
gives 
\[
\mathcal{N}\left(\mathcal{F}_{\bar{\eta},h},L_{2}(Q),\epsilon\right)\leq\left|\mathcal{A}\right|\left(\frac{3\Updelta\left\Vert K\right\Vert _{\infty}\left\Vert 1/\bar{\pi}_{a}\right\Vert _{\infty}}{\epsilon h^{d}}\right)^{\nu}.
\]
In particular, for $\mathcal{F}_{h}=\mathcal{F}_{\eta,h}$, along
with the condition $\left\Vert \frac{1}{\bar{\pi}_{a}}\right\Vert _{\infty}\leq\frac{1}{\varepsilon}$, we have
\[
\mathcal{N}\left(\mathcal{F}_{h},L_{2}(Q),\epsilon\right)\leq\left|\mathcal{A}\right|\left(\frac{3\Updelta\left\Vert K\right\Vert _{\infty}}{\varepsilon\epsilon h^{d}}\right)^{\nu}.
\]

(b) It is immediate to see that
\begin{align*}
\mathcal{N}\left(\mathcal{F}_{\eta,\hat{\eta},h},L_{2}(Q),\epsilon\right) & \leq\mathcal{N}\left(\mathcal{F}_{h},L_{2}(Q),\frac{\epsilon}{2}\right)\mathcal{N}\left(\hat{\mathcal{F}}_{h},L_{2}(Q),\frac{\epsilon}{2}\right)\\
 & \leq\left|\mathcal{A}\right|^{2}\left(\frac{6\Updelta\left\Vert K\right\Vert _{\infty}}{\varepsilon\epsilon h^{d}}\right)^{2\nu}.
\end{align*}

\end{proof}

\begin{lemma} 
\label{lem:asymptotic_empirical_observe}

Under Assumption \ref{assumption:kernel_vc}, we have the following weak convergence
\begin{align*}
 & \sqrt{n}(\mathbb{P}_{n}-\mathbb{P})\to\mathbb{G}\text{ weakly in }\ell^{\infty}(\mathcal{F}_{h}),\\
 & \sqrt{n}(\mathbb{P}_{n}^{*}-\mathbb{P}_{n})\to\mathbb{G}\text{ weakly in }\ell^{\infty}(\mathcal{F}_{h})\text{ a.s.},
\end{align*}
where $\mathbb{G}$ is a centered Gaussian process with $Cov\left[\mathbb{G}(f_{y_{1}}^{a_{1}}),\mathbb{G}(f_{y_{2}}^{a_{2}})\right]=\mathbb{E}_{\mathbb{P}}\left[f_{y_{1}}^{a_{1}}f_{y_{2}}^{a_{2}}\right]-\mathbb{E}_{\mathbb{P}}\left[f_{y_{1}}^{a_{1}}\right]\mathbb{E}_{\mathbb{P}}\left[f_{y_{2}}^{a_{2}}\right]$.
\end{lemma}

\begin{proof}
To show the desired weak convergence, we shall show that $\mathcal{F}_{h}$ is $P$-donsker for
any probability distribution $P$ using Proposition~\ref{prop:empirical_donsker_covering}.
Note that for any $f_{y}^{a}\in\mathcal{F}_{h}$, 
\[
\left\Vert f_{h}^{a}\right\Vert _{\infty}\leq3\left\Vert K\right\Vert _{\infty}\left\Vert 1/\pi_{a}\right\Vert _{\infty}h^{-d},
\]
so a constant function $3\left\Vert K\right\Vert _{\infty}\left\Vert 1/\pi_{a}\right\Vert _{\infty}h^{-d}$
acts as a uniform upper bound of $\mathcal{F}_{h}$. Then it is straightforward to see
that 
\begin{align*}
    \mathbb{E}_{P}\left[(3\left\Vert K\right\Vert _{\infty}\left\Vert 1/\pi_{a}\right\Vert _{\infty}h^{-d})^{2}\right]&=9\left\Vert K\right\Vert _{\infty}^{2}\left\Vert 1/\pi_{a}\right\Vert _{\infty}^{2}h^{-2d} \\
    &\leq 9\left\Vert K\right\Vert _{\infty}^{2}\frac{1}{\varepsilon^2 h^{2d}} < \infty.
\end{align*}
Under Assumption \ref{assumption:kernel_vc}, Claim~\ref{claim:doubly_robust_vc}
gives a uniformly bounded VC dimension of $\mathcal{F}_{h}$ as 
\[
\mathcal{N}\left(\mathcal{F}_{h},L_{2}(Q),\epsilon\right)\leq\left|\mathcal{A}\right|\left(\frac{3\Updelta\left\Vert K\right\Vert _{\infty}}{\varepsilon\epsilon h^{d}}\right)^{\nu}.
\]
Then it follows
\[
\int_{0}^{1}\sqrt{\log\sup_{Q}\mathcal{N}\left(\mathcal{F}_{h},L_{2}(Q),3\left\Vert K\right\Vert _{\infty}\varepsilon^{-1}{h}^{-d}\epsilon\right)}d\epsilon\leq\int_{0}^{1}\sqrt{\nu\log\left(\frac{\Updelta}{\epsilon}\right)+\log\left|\mathcal{A}\right|}d\epsilon<\infty.
\]
Thus, Proposition~\ref{prop:empirical_donsker_covering} implies that $\mathcal{F}_{h}$ is of $P$-Donsker class for any distribution $P$, and in particular,
\[
\sqrt{n}(\mathbb{P}_{n}-\mathbb{P})\to\mathbb{G}\text{ weakly in }\ell^{\infty}(\mathcal{F}_{h}),
\]
where $\mathbb{G}$ is a centered Gaussian process with  $Cov\left[\mathbb{G}(f_{y_{1}}^{a_{1}}),\mathbb{G}(f_{y_{2}}^{a_{2}})\right]=\mathbb{E}_{\mathbb{P}}\left[f_{y_{1}}^{a_{1}}f_{y_{2}}^{a_{2}}\right]-\mathbb{E}_{\mathbb{P}}\left[f_{y_{1}}^{a_{1}}\right]\mathbb{E}_{\mathbb{P}}\left[f_{y_{2}}^{a_{2}}\right]$.

Finally, by Proposition~\ref{prop:empirical_donsker_bootstrap} we get
\[
\sqrt{n}(\mathbb{P}_{n}^{*}-\mathbb{P}_{n})\to\mathbb{G}\text{ weakly in }\ell^{\infty}(\mathcal{F}_{h})\text{ a.s.}.
\]
\end{proof}

\begin{claim}

\label{claim:bounded_support_observe}

If $u\notin\mathcal{Y}_{h}$, 
\[
\widehat{p}_{a,h}(u)=p_{a,h}(u)=0.
\]

\end{claim}

\begin{proof}

For all $u\notin\mathcal{Y}_{h}$ and $y\in\mathcal{Y}$,
$K\left(\frac{u-y}{h}\right)=0$. Hence, $\widehat{p}_{a,h}(u)=p_{a,h}(u)=0$
if $u\notin\mathcal{Y}_{h}$.

\end{proof}

\begin{claim}

\label{claim:asymptotic_empirical_envelope}

For a fixed $\hat{\eta}$, recall the class $\bar{\mathcal{F}}_{h}$
defined in \eqref{eq:notation_doubly_robust_class_general}. Let $F$
be 
\[
F(Z)=\underset{a\in\mathcal{A},y\in\mathcal{Y}}{\sup}\left\vert \hat{f}_{h,y}^{a}(Z)-f_{h,y}^{a}(Z)\right\vert .
\]
Then $F$ becomes an envelope of $\bar{\mathcal{F}}_{h}$, and its
$L_{\infty}$ norm is bounded as 
\[
\left\Vert F\right\Vert _{\infty}\leq\frac{\left\Vert K\right\Vert _{\infty}}{\varepsilon^{2}h^{d}}\underset{a\in\mathcal{A}}{\sup}\left\Vert \hat{\pi}_{a}-\pi_{a}\right\Vert _{\infty}+\left(1+\frac{1}{\varepsilon}\right)\underset{a\in\mathcal{A}}{\sup}\left\Vert \widehat{\mu}_{a,y}-\mu_{a,y}\right\Vert _{\infty}.
\]
And when $\underset{a\in\mathcal{A}}{\sup}\left\Vert \hat{\pi}_{a}-\pi_{a}\right\Vert _{\infty}$
and $\underset{a\in\mathcal{A}}{\sup}\left\Vert \widehat{\mu}_{a,y}-\mu_{a,y}\right\Vert _{\infty}$
are small enough so that $\frac{\left\Vert K\right\Vert _{\infty}}{\varepsilon^{2}h^{d}}\underset{a\in\mathcal{A}}{\sup}\left\Vert \hat{\pi}_{a}-\pi_{a}\right\Vert _{\infty}<\frac{1}{2}$
and $\left(1+\frac{1}{\varepsilon}\right)\underset{a\in\mathcal{A}}{\sup}\left\Vert \widehat{\mu}_{a,y}-\mu_{a,y}\right\Vert _{\infty}<\frac{1}{2}$,
then
\begin{align*}
 & \left\Vert F\right\Vert _{\infty}\sqrt{\log(1/\left\Vert F\right\Vert _{\infty})}\\
 & \leq C_{\left\Vert K\right\Vert _{\infty},\varepsilon,h}\left(\underset{a\in\mathcal{A}}{\sup}\left\Vert \hat{\pi}_{a}-\pi_{a}\right\Vert _{\infty}\sqrt{\log\left(1/\underset{a\in\mathcal{A}}{\sup}\left\Vert \hat{\pi}_{a}-\pi_{a}\right\Vert _{\infty}\right)}+\underset{a\in\mathcal{A}}{\sup}\left\Vert \widehat{\mu}_{a,y}-\mu_{a,y}\right\Vert _{\infty}\sqrt{\log\left(1/\underset{a\in\mathcal{A}}{\sup}\left\Vert \widehat{\mu}_{a,y}-\mu_{a,y}\right\Vert _{\infty}\right)}\right),
\end{align*}where $C_{\left\Vert K\right\Vert _{\infty},\varepsilon,h}$ is a
constant depending only on $\left\Vert K\right\Vert _{\infty},\varepsilon,h$.

\end{claim}

\begin{proof}

From Claim~\ref{claim:pa-hat_expand}~(b), under Assumptions \ref{assumption:c3},
\ref{assumption:A1}, \ref{assumption:A2},
\begin{align*}
 & \left|\hat{f}_{h,y}^{a}(Z)-{f}_{h,y}^{a}(Z)\right|\\
 & \leq\left|\pi_{a}(X)-\hat{\pi}_{a}(X)\right|\left\Vert \frac{1}{\widehat{\pi}_{a}}\right\Vert _{\infty}\left|\frac{\mathbbm{1}(A=a)}{{\pi}_{a}(X)}\left\{ K_{h,y}(Y)-{\mu}_{A,y}(X)\right\} \right|+\left(1+\left\Vert \frac{1}{\widehat{\pi}_{a}}\right\Vert _{\infty}\right)\left|\widehat{\mu}_{a,y}(X)-{\mu}_{a,y}(X)\right|\\
 & \leq\frac{\left\Vert K\right\Vert _{\infty}}{\varepsilon^{2}h^{d}}\left|\pi_{a}(X)-\hat{\pi}_{a}(X)\right|+\left(1+\frac{1}{\epsilon}\right)\left|\widehat{\mu}_{a,y}(X)-{\mu}_{a,y}(X)\right|
\end{align*}
Therefore, $\left\Vert F\right\Vert _{\infty}$ is bounded as 
\begin{align*}
\left\Vert F\right\Vert _{\infty} & =\underset{a\in\mathcal{A},y\in\mathcal{Y}}{\sup}\left\Vert \hat{f}_{h,y}^{a}-f_{h,y}^{a}\right\Vert _{\infty}\\
 & \leq\frac{\left\Vert K\right\Vert _{\infty}}{\varepsilon^{2}h^{d}}\underset{a\in\mathcal{A}}{\sup}\left\Vert \hat{\pi}_{a}-\pi_{a}\right\Vert _{\infty}+\left(1+\frac{1}{\varepsilon}\right)\underset{a\in\mathcal{A}}{\sup}\left\Vert \widehat{\mu}_{a,y}-\mu_{a,y}\right\Vert _{\infty}.
\end{align*}

\end{proof}

\begin{claim}

\label{claim:asymptotic_smooth_density_diff}

Recall the class $\bar{\mathcal{F}}_{h}$ defined in \eqref{eq:notation_doubly_robust_class_general}.
Suppose $\underset{a\in\mathcal{A}}{\sup}\left\Vert \hat{\pi}_{a}-\pi_{a}\right\Vert _{\infty}\sqrt{\log\left(1/\underset{a\in\mathcal{A}}{\sup}\left\Vert \hat{\pi}_{a}-\pi_{a}\right\Vert _{\infty}\right)}\to0$
a.s. and $\underset{a\in\mathcal{A}}{\sup}\left\Vert \widehat{\mu}_{a,y}-\mu_{a,y}\right\Vert _{\infty}\sqrt{\log\left(1/\underset{a\in\mathcal{A}}{\sup}\left\Vert \widehat{\mu}_{a,y}-\mu_{a,y}\right\Vert _{\infty}\right)}\to0$
a.s. hold. Then,

\[
\underset{f\in\bar{\mathcal{F}}_{h}}{\sup}\left\vert \sqrt{n}(\mathbb{P}_{n}-\mathbb{P})f\right\vert =o_{\mathbb{P}}(1)\text{ a.s.},
\]
and 
\[
\underset{f\in\bar{\mathcal{F}}_{h}}{\sup}\left\vert \sqrt{n}(\mathbb{P}_{n}^{*}-\mathbb{P}_{n})f\right\vert =o_{\mathbb{P}}(1)\text{ a.s.}.
\]

\end{claim}

\begin{proof}

Under Assumption~\ref{assumption:kernel_vc}, \ref{assumption:A2},
Claim~\ref{claim:doubly_robust_vc}~(b) gives the upper bound for
the covering number as 
\[
\sup_{Q}\mathcal{N}\left(\bar{\mathcal{F}}_{h},L_{2}(Q),\epsilon\right)\leq\left|\mathcal{A}\right|\left(\frac{6\Updelta\left\Vert K\right\Vert _{\infty}}{\varepsilon\epsilon h^{d}}\right)^{2\nu}.
\]
And hence 
\begin{align*}
J(\bar{\mathcal{F}}_{h},\left\Vert F\right\Vert _{\infty},1) & =\int_{0}^{1}\sqrt{\log\sup_{Q}\mathcal{N}(\mathcal{F}_{n},L_{2}(Q),\epsilon\left\Vert F\right\Vert _{\infty})}d\epsilon\\
 & \leq\int_{0}^{1}\sqrt{2\nu\log\left(\frac{6\Updelta\left\Vert K\right\Vert _{\infty}}{\varepsilon\epsilon h^{d}\left\Vert F\right\Vert _{\infty}}\right)+\log\left|\mathcal{A}\right|}d\epsilon\\
 & \leq\sqrt{2\nu d\log h}+\sqrt{2\nu\log(1/\left\Vert F\right\Vert _{\infty})}+C_{\nu,\left\Vert K\right\Vert _{\infty},\varepsilon,\left|\mathcal{A}\right|},
\end{align*}
where $C_{\nu,\left\Vert K\right\Vert _{\infty},\varepsilon,\left|\mathcal{A}\right|}$
is a constant depending only on $\nu,\left\Vert K\right\Vert _{\infty},\varepsilon,\left|\mathcal{A}\right|$.
Hence by applying Theorem~\ref{thm:empirical_bound_envelope},
we obtain that 
\begin{align}
\mathbb{P}\left\{ \underset{f\in\mathcal{F}_{n}}{\sup}\left\vert \sqrt{n}(\mathbb{P}_{n}-\mathbb{P})f\right\vert \right\}  & \leq\left(8\sqrt{2}J(\mathcal{F},\left\Vert F\right\Vert _{\infty},1)+2\right)\left\Vert F\right\Vert _{\infty}\nonumber \\
 & \leq\left\Vert F\right\Vert _{\infty}\left(16\sqrt{\nu\log(1/\left\Vert F\right\Vert _{\infty})}+C_{\nu,\left\Vert K\right\Vert _{\infty},\varepsilon,\left|\mathcal{A}\right|,h,d}\right),\label{eq:asymptotic_smooth_density_diff_bound}
\end{align}
and by applying Theorem~\ref{thm:empirical_bound_envelope_bootstrap},
we obtain that 
\begin{align}
\mathbb{P}\left\{ \underset{f\in\mathcal{F}_{n}}{\sup}\left\vert \sqrt{n}(\mathbb{P}_{n}^{*}-\mathbb{P}_{n})f\right\vert \right\}  & \leq\left(8\sqrt{2}J(\mathcal{F},\left\Vert F\right\Vert _{\infty},1)+2\right)\left\Vert F\right\Vert _{\infty}\nonumber \\
 & \leq\left\Vert F\right\Vert _{\infty}\left(16\sqrt{\nu\log(1/\left\Vert F\right\Vert _{\infty})}+C_{\nu,\left\Vert K\right\Vert _{\infty},\varepsilon,\left|\mathcal{A}\right|,h,d}\right),\label{eq:asymptotic_smooth_density_diff_bound_bootstrap}
\end{align}
 where $C_{\nu,\left\Vert K\right\Vert _{\infty},\varepsilon,\left|\mathcal{A}\right|,h,d}$
is a constant depending only on $\nu,\left\Vert K\right\Vert _{\infty},\varepsilon,\left|\mathcal{A}\right|,h,d$. 

Now, consider how \eqref{eq:asymptotic_smooth_density_diff_bound}
and \eqref{eq:asymptotic_smooth_density_diff_bound_bootstrap} change
as $n\to\infty$. Note that when $\left\Vert F\right\Vert _{\infty}$
is small, both \eqref{eq:asymptotic_smooth_density_diff_bound} and
\eqref{eq:asymptotic_smooth_density_diff_bound_bootstrap} can be
further bounded as
\begin{align}
 & \max\left\{ \mathbb{P}\left\{ \underset{f\in\mathcal{F}_{n}}{\sup}\left\vert \sqrt{n}(\mathbb{P}_{n}-\mathbb{P})f\right\vert \right\} ,\mathbb{P}\left\{ \underset{f\in\mathcal{F}_{n}}{\sup}\left\vert \sqrt{n}(\mathbb{P}_{n}-\mathbb{P})f\right\vert \right\} \right\} \nonumber \\
 & \leq\left\Vert F\right\Vert _{\infty}\left(16\sqrt{\nu\log(1/\left\Vert F\right\Vert _{\infty})}+C_{\nu,\left\Vert K\right\Vert _{\infty},\varepsilon,\left|\mathcal{A}\right|,h,d}\right)\nonumber \\
 & \leq C'_{\nu,\left\Vert K\right\Vert _{\infty},\varepsilon,\left|\mathcal{A}\right|,h,d}\left\Vert F\right\Vert _{\infty}\sqrt{\log(1/\left\Vert F\right\Vert _{\infty})},\label{eq:asymptotic_smooth_density_diff_bound_small_F}
\end{align}
where $C'_{\nu,\left\Vert K\right\Vert _{\infty},\varepsilon,\left|\mathcal{A}\right|,h,d}$
is a constant depending only on $\nu,\left\Vert K\right\Vert _{\infty},\varepsilon,\left|\mathcal{A}\right|,h,d$.
Then from Claim~\ref{claim:asymptotic_empirical_envelope}, under
the conditions that $\underset{a\in\mathcal{A}}{\sup}\left\Vert \hat{\pi}_{a}-\pi_{a}\right\Vert _{\infty}\sqrt{\log\left(1/\underset{a\in\mathcal{A}}{\sup}\left\Vert \hat{\pi}_{a}-\pi_{a}\right\Vert _{\infty}\right)}\to0$
a.s. and $\underset{a\in\mathcal{A}}{\sup}\left\Vert \widehat{\mu}_{a,y}-\mu_{a,y}\right\Vert _{\infty}\sqrt{\log\left(1/\underset{a\in\mathcal{A}}{\sup}\left\Vert \widehat{\mu}_{a,y}-\mu_{a,y}\right\Vert _{\infty}\right)}\to0$
a.s., 

\[
\left\Vert F\right\Vert _{\infty}\sqrt{\log(1/\left\Vert F\right\Vert _{\infty})}\to0\text{ a.s..}
\]
By applying this to \eqref{eq:asymptotic_smooth_density_diff_bound_small_F},
we get 
\[
\underset{f\in\mathcal{F}_{n}}{\sup}\left\vert \sqrt{n}(\mathbb{P}_{n}-\mathbb{P})f\right\vert =o_{\mathbb{P}}(1)\text{ a.s.},
\]
and 
\[
\underset{f\in\mathcal{F}_{n}}{\sup}\left\vert \sqrt{n}(\mathbb{P}_{n}^{*}-\mathbb{P}_{n})f\right\vert =o_{\mathbb{P}}(1)\text{ a.s.}.
\]

\end{proof}

We are now prepared to present the proof of Theorem \ref{thm:asymptotic_smooth_density}.

\begin{proof}[Proof of Theorem \ref{thm:asymptotic_smooth_density}]
To show the desired weak convergence, for $\forall a\in\mathcal{A}$ and $y\in\mathcal{Y}$, we first expand $\sqrt{n}\left(\hat{p}_{h}^{a}(y)-p_{h}^{a}(y)\right)$
by Claim~\ref{claim:pa-hat_expand}~(a)
as 
\begin{align}
\sqrt{n}\left(\hat{p}_{h}^{a}(y)-p_{h}^{a}(y)\right) & =\sqrt{n}\left(\mathbb{P}_{n}\hat{f}_{h,y}^{a}-\mathbb{P}f_{h,y}^{a}\right)\nonumber \\
 & =\sqrt{n}(\mathbb{P}_{n}-\mathbb{P})f_{h,y}^{a}+\sqrt{n}\mathbb{P}\left(\hat{f}_{h,y}^{a}-f_{h,y}^{a}\right)+\sqrt{n}(\mathbb{P}_{n}-\mathbb{P})\left(\hat{f}_{h,y}^{a}-f_{h,y}^{a}\right).\label{eq:asymptotic_smooth_density_expand}
\end{align}
We compute the weak convergence limit for each term of \eqref{eq:asymptotic_smooth_density_expand}.

Applying Lemma~\ref{lem:asymptotic_empirical_observe} to the first term of \eqref{eq:asymptotic_smooth_density_expand} yields
$\sqrt{n}(\mathbb{P}_{n}-\mathbb{P})\to\tilde{\mathbb{G}}$
weakly in $\ell^{\infty}(\mathcal{F}_{h})$ where $\tilde{\mathbb{G}}$
is a centered Gaussian process with $Cov\left[\tilde{\mathbb{G}}(f_{y_{1}}^{a}),\tilde{\mathbb{G}}(f_{y_{2}}^{a})\right]=\mathbb{E}\left[f_{y_{1}}^{a}f_{y_{2}}^{a}\right]-\mathbb{E}\left[f_{y_{1}}^{a}\right]\mathbb{E}\left[f_{y_{2}}^{a}\right]$.
This is equivalent to that $\left\{ \sqrt{n}(\mathbb{P}_{n}-\mathbb{P})f_{h,y}^{a}\right\} _{a,y}\to\mathbb{G}$
weakly in $\ell^{\infty}(\mathcal{A}\times\mathcal{Y}_{h})$, where $\mathbb{G}$ is
a centered Gaussian process with $Cov\left[\mathbb{G}(a_{1},y_{1}),\mathbb{G}(a_{2},y_{2})\right]=\mathbb{E}\left[f_{y_{1}}^{a_{1}}f_{y_{2}}^{a_{2}}\right]-\mathbb{E}\left[f_{y_{1}}^{a_{1}}\right]\mathbb{E}\left[f_{y_{2}}^{a_{2}}\right]$.
Since Claim~\ref{claim:bounded_support_observe} implies that
$f_{h,y}^{a}=0$ for $y\in\mathbb{R}^{d}\backslash\mathcal{Y}_{h}$,
consequently we have 
\begin{equation}
\left\{ \sqrt{n}(\mathbb{P}_{n}-\mathbb{P})f_{h,y}^{a}\right\} _{y}\to\mathbb{G}\text{ weakly in }\ell^{\infty}(\mathcal{A}\times\mathbb{R}^{d}).\label{eq:asymptotic_smooth_density_convergence_first}
\end{equation}

For the second term, since $h$ is fixed, 
by 
Claim~\ref{claim:pa-hat_expand}~(c)
, the upper bound is given by
\[
\left|\sqrt{n}\mathbb{P}\left(\hat{f}_{h,y}^{a}-f_{h,y}^{a}\right)\right|\leq\sqrt{n}\left\Vert \frac{1}{\widehat{\pi}_{a}(X)}\right\Vert _{\infty}\left\Vert \widehat{\mu}_{a,y}-{\mu}_{a,y}\right\Vert \left\Vert \widehat{\pi}_{a}-{\pi}_{a}\right\Vert .
\]
Under Assumptions \ref{assumption:A2} and \ref{assumption:A4'}, we have $\left\Vert \frac{1}{\hat{\pi}_{a}(X)}\right\Vert_{\infty}<\infty$
and $\sup_{y\in\mathbb{R}^{d}}\left\Vert \widehat{\mu}_{a,y}-{\mu}_{a,y}\right\Vert \left\Vert \widehat{\pi}_{a}-{\pi}_{a}\right\Vert =o_{\Pb}(1/\sqrt{n})$, $\forall a\in\mathcal{A}$. Then, 
\begin{align*}
\sup_{a\in\mathcal{A},y\in\mathcal{Y}}\left|\sqrt{n}\mathbb{P}\left(\hat{f}_{h,y}^{a}-f_{h,y}^{a}\right)\right| & \leq\sqrt{n}\left\Vert \frac{1}{\widehat{\pi}_{a}(X)}\right\Vert _{\infty}\left\Vert \widehat{\mu}_{a,y}-{\mu}_{a,y}\right\Vert \left\Vert \widehat{\pi}_{a}-{\pi}_{a}\right\Vert \\
 & \leq\sqrt{n}\left\Vert \frac{1}{\widehat{\pi}_{a}(X)}\right\Vert _{\infty}\sup_{a\in\mathcal{A}}\sup_{y\in\mathbb{R}^{d}}\left\Vert \widehat{\mu}_{a,y}-{\mu}_{a,y}\right\Vert \left\Vert \widehat{\pi}_{a}-{\pi}_{a}\right\Vert \\
 & =o_{\mathbb{P}}(1),
\end{align*}
in particular since $\left|\mathcal{A}\right|<\infty$. Hence 
\[
\left\{ \sqrt{n}\mathbb{P}\left(\hat{f}_{h,y}^{a}-f_{h,y}^{a}\right)\right\} _{a\in\mathcal{A},y\in\mathbb{R}^{d}}=o_{\mathbb{P}}(1)\text{ in }\ell^{\infty}(\mathcal{A}\times\mathbb{R}^{d}),
\]
and since $o_{\mathbb{P}}(1)$ implies convergence in distribution,
\begin{equation}
\left\{ \sqrt{n}\mathbb{P}\left(\hat{f}_{h,y}^{a}-f_{h,y}^{a}\right)\right\} _{a\in\mathcal{A},y\in\mathbb{R}^{d}}\to0\text{ weakly in }\ell^{\infty}(\mathcal{A}\times\mathbb{R}^{d}).\label{eq:asymptotic_smooth_density_convergence_second}
\end{equation}

For the third term, we will show that 
\[
\left\{ \sqrt{n}(\mathbb{P}_{n}-\mathbb{P})\left(\hat{f}_{h,y}^{a}-f_{h,y}^{a}\right)\right\} _{a\in\mathcal{A},y\in\mathbb{R}^{d}}=o_{\mathbb{P}}(1)\text{ a.s. in }\ell^{\infty}(\mathcal{A}\times\mathbb{R}^{d}).
\]
Recall the class $\bar{\mathcal{F}}_{h}$ defined in \eqref{eq:notation_doubly_robust_class_general}.
Then the above is equivalent to that 
\[
\underset{f\in\bar{\mathcal{F}}_{h}}{\sup}\left\vert \sqrt{n}(\mathbb{P}_{n}-\mathbb{P})f\right\vert =o_{\mathbb{P}}(1)\text{ a.s.},
\]
which is shown by Claim~\ref{claim:asymptotic_smooth_density_diff} under Assumption \ref{assumption:A4'}.
Hence we get 
\begin{equation}
\left\{ \sqrt{n}(\mathbb{P}_{n}-\mathbb{P})\left(\hat{f}_{h,y}^{a}-f_{h,y}^{a}\right)\right\} _{a\in\mathcal{A},y\in\mathbb{R}^{d}}=o_{\mathbb{P}}(1)\text{ a.s. in }\ell^{\infty}(\mathcal{A}\times\mathbb{R}^{d}).\label{eq:asymptotic_smooth_density_convergence_third}
\end{equation}

Finally, combining \eqref{eq:asymptotic_smooth_density_convergence_first},
\eqref{eq:asymptotic_smooth_density_convergence_second}, and \eqref{eq:asymptotic_smooth_density_convergence_third}
with Slutsky Theorem \citep[][Theorem 7.15]{Kosorok2008} results in
the weak convergence of \eqref{eq:asymptotic_smooth_density_expand}
as 
\[
\sqrt{n}\left(\hat{p}_{h}-p_{h}\right)=\left\{ \sqrt{n}\left(\mathbb{P}_{n}\hat{f}_{h,y}^{a}-\mathbb{P}f_{h,y}^{a}\right)\right\} _{a,y}\to\mathbb{G}\text{ weakly in }\ell^{\infty}(\mathcal{A}\times\mathbb{R}^{d})\text{ a.s.},
\]
where $\mathbb{G}$ is a centered Gaussian process with $Cov\left[\mathbb{G}(a_{1},y_{1}),\mathbb{G}(a_{2},y_{2})\right]=\mathbb{E}\left[f_{y_{1}}^{a_{1}}f_{y_{2}}^{a_{2}}\right]-\mathbb{E}\left[f_{y_{1}}^{a_{1}}\right]\mathbb{E}\left[f_{y_{2}}^{a_{2}}\right]$.
\end{proof}

Before proving Corollary~\ref{cor:confidence_smooth_density}, we give the following propositions.

\begin{proposition} \label{cor:asymptotic_smooth_density_bootstrap}
Suppose that Assumptions \ref{assumption:A2}, \ref{assumption:A3}, \ref{assumption:A4'},
and \ref{assumption:kernel_vc} hold. When $\sqrt{n}\left(\hat{p}_{h}^{*}-\hat{p}_{h}\right)$
is understood as a stochastic process indexed by $(a,y)\in\mathcal{A}\times\mathbb{R}^{d}$
such that $\left\{ \sqrt{n}\left(\hat{p}_{h}^{*}-\hat{p}_{h}\right)\right\} _{(a,y)}=\sqrt{n}\left(\hat{p}_{a,h}^{*}(y)-\hat{p}_{a,h}(y)\right)$,
we have the following weak convergence: 
\[
\sqrt{n}\left(\hat{p}_{h}^{*}-\hat{p}_{h}\right)\to\mathbb{G}\text{ weakly in }\ell^{\infty}(\mathcal{A}\times\mathbb{R}^{d})\text{ a.s.},
\]
where $\mathbb{G}$ is a centered Gaussian process with $Cov\left[\mathbb{G}(a_{1},y_{1}),\mathbb{G}(a,y_{2})\right]=\mathbb{E}\left[f_{y_{1}}^{a_{1}}f_{y_{2}}^{a_{2}}\right]-\mathbb{E}\left[f_{y_{1}}^{a_{1}}\right]\mathbb{E}\left[f_{y_{2}}^{a_{2}}\right]$,
$f_{y}^{a}(Z)\coloneqq\frac{\mathbf{1}(A=a)}{\pi_{a}(X)}\big(K_{h,y}(Y)-\mu_{A,y}(X)\big)+\mu_{a,y}(X)$.
\end{proposition}

\begin{proof}[Proof of Proposition~\ref{cor:asymptotic_smooth_density_bootstrap}]
For $\sqrt{n}\left(\hat{p}_{h}^{*}-\hat{p}_{h}\right)=\left\{ \sqrt{n}\left(\hat{p}_{a,h}^{*}(y)-\hat{p}_{a,h}(y)\right)\right\} _{a,y}$, $\forall a\in\mathcal{A}$ and $y\in\mathcal{Y}$, we have the following decomposition:
\begin{align}
\sqrt{n}\left(\hat{p}_{a,h}^{*}(y)-\hat{p}_{a,h}(y)\right) & =\sqrt{n}\left(\mathbb{P}_{n}^{*}\hat{f}_{h,y}^{a}-\mathbb{P}_{n}\hat{f}_{h,y}^{a}\right)\nonumber \\
 & =\sqrt{n}(\mathbb{P}_{n}^{*}-\mathbb{P}_{n})f_{h,y}^{a}+\sqrt{n}(\mathbb{P}_{n}^{*}-\mathbb{P}_{n})\left(\hat{f}_{h,y}^{a}-f_{h,y}^{a}\right).\label{eq:asymptotic_smooth_density_bootstrap_expand}
\end{align}
We compute the weak convergence limit for each term of \eqref{eq:asymptotic_smooth_density_bootstrap_expand}.

For the first term,
note that under Assumption \ref{assumption:kernel_vc}, Lemma~\ref{lem:asymptotic_empirical_observe}
implies that 
\[
\sqrt{n}(\mathbb{P}_{n}^{*}-\mathbb{P}_{n})\to\tilde{\mathbb{G}}\text{ weakly in }\ell^{\infty}(\mathcal{F}_{h})\text{ a.s.},
\]
 where $\tilde{\mathbb{G}}$ is a centered Gaussian process with $Cov\left[\tilde{\mathbb{G}}(f_{y_{1}}^{a_{1}}),\tilde{\mathbb{G}}(f_{y_{2}}^{a_{2}})\right]=\mathbb{E}\left[f_{y_{1}}^{a_{1}}f_{y_{2}}^{a_{2}}\right]-\mathbb{E}\left[f_{y_{1}}^{a_{1}}\right]\mathbb{E}\left[f_{y_{2}}^{a_{2}}\right]$.
Consequently, we have 
\[
\left\{ \sqrt{n}(\mathbb{P}_{n}^{*}-\mathbb{P}_{n})f_{h,y}^{a}\right\} _{a,y}\to\mathbb{G}\text{ weakly in }\ell^{\infty}(\mathcal{A}\times\mathcal{Y}_{h})\text{ a.s.},
\]
where $\mathbb{G}$ is a centered Gaussian process with $Cov\left[\mathbb{G}(a_{1},y_{1}),\mathbb{G}(a_{2},y_{2})\right]=\mathbb{E}\left[f_{y_{1}}^{a_{1}}f_{y_{2}}^{a_{2}}\right]-\mathbb{E}\left[f_{y_{1}}^{a_{1}}\right]\mathbb{E}\left[f_{y_{2}}^{a_{2}}\right]$.
And since Claim~\ref{claim:bounded_support_observe} implies that
$f_{h,y}^{a}=0$ for $y\in\mathbb{R}^{d}\backslash\mathcal{Y}_{h}$,
we have 
\begin{equation}
\left\{ \sqrt{n}(\mathbb{P}_{n}^{*}-\mathbb{P}_{n})f_{h,y}^{a}\right\} _{a,y}\to\mathbb{G}\text{ weakly in }\ell^{\infty}(\mathcal{A}\times\mathbb{R}^{d})\text{ a.s.}.\label{eq:asymptotic_smooth_density_bootstrap_convergence_first}
\end{equation}

For the second term, we will show that 
\[
\left\{ \sqrt{n}(\mathbb{P}_{n}^{*}-\mathbb{P}_{n})\left(\hat{f}_{h,y}^{a}-f_{h,y}^{a}\right)\right\} _{a\in\mathcal{A},y\in\mathbb{R}^{d}}=o_{\mathbb{P}}(1)\text{ a.s. in }\ell^{\infty}(\mathcal{A}\times\mathbb{R}^{d}).
\]
Recall the class $\bar{\mathcal{F}}_{h}$ defined in \eqref{eq:notation_doubly_robust_class_general}.
Then the above is equivalent to that 
\[
\underset{f\in\bar{\mathcal{F}}_{h}}{\sup}\left\vert \sqrt{n}(\mathbb{P}_{n}^{*}-\mathbb{P}_{n})f\right\vert =o_{\mathbb{P}}(1)\text{ a.s.},
\]
which is shown by Claim~\ref{claim:asymptotic_smooth_density_diff} under Assumption \ref{assumption:A4'}.
Hence we have
\begin{equation}
\left\{ \sqrt{n}(\mathbb{P}_{n}^{*}-\mathbb{P}_{n})\left(\hat{f}_{h,y}^{a}-f_{h,y}^{a}\right)\right\} _{a\in\mathcal{A},y\in\mathbb{R}^{d}}=o_{\mathbb{P}}(1)\text{ a.s. in }\ell^{\infty}(\mathcal{A}\times\mathbb{R}^{d}).\label{eq:asymptotic_smooth_density_bootstrap_convergence_third}
\end{equation}

Finally, as before, combining \eqref{eq:asymptotic_smooth_density_bootstrap_convergence_first}
and \eqref{eq:asymptotic_smooth_density_bootstrap_convergence_third}
with Slutsky Theorem \citep[][Theorem 7.15]{Kosorok2008} characterizes
the weak convergence of \eqref{eq:asymptotic_smooth_density_bootstrap_expand}
as 
\[
\sqrt{n}\left(\hat{p}_{h}^{*}-\hat{p}_{h}\right)=\left\{ \sqrt{n}\left(\mathbb{P}_{n}^{*}\hat{f}_{h,y}^{a}-\mathbb{P}_{n}\hat{f}_{h,y}^{a}\right)\right\} _{a,y}\to\mathbb{G}\text{ weakly in }\ell^{\infty}(\mathcal{A}\times\mathbb{R}^{d})\text{ a.s.},
\]
where $\mathbb{G}$ is a centered Gaussian process with $Cov\left[\mathbb{G}(a_{1},y_{1}),\mathbb{G}(a_{2},y_{2})\right]=\mathbb{E}\left[f_{y_{1}}^{a_{1}}f_{y_{2}}^{a_{2}}\right]-\mathbb{E}\left[f_{y_{1}}^{a_{1}}\right]\mathbb{E}\left[f_{y_{2}}^{a_{2}}\right]$.

\end{proof}

\begin{proposition} \label{cor:asymptotic_linfty} Assume that Assumptions \ref{assumption:A2}, \ref{assumption:A3}, \ref{assumption:A4'}, and \ref{assumption:kernel_vc}
hold. Then for a given $h$ and each $a\in\mathcal{A}$, we have the
following weak convergence as 
\[
\sqrt{n}\left\Vert \hat{p}_{a,h}-p_{a,h}\right\Vert _{\infty}\to\sup_{y\in\mathcal{Y}_{h}}\left|\mathbb{G}(a,y)\right|\text{ weakly in }\mathbb{R}\text{ a.s.},
\]
where $\mathbb{G}$ is a centered Gaussian process with $Cov\left[\mathbb{G}(a_{1},y_{1}),\mathbb{G}(a_{2},y_{2})\right]=\mathbb{E}\left[f_{y_{1}}^{a_{1}}f_{y_{2}}^{a_{2}}\right]-\mathbb{E}\left[f_{y_{1}}^{a_{1}}\right]\mathbb{E}\left[f_{y_{2}}^{a_{2}}\right]$,
$f_{y}^{a}(Z)\coloneqq\frac{\mathbf{1}(A=a)}{\pi_{a}(X)}\big(K_{h,y}(Y)-\mu_{A,y}(X)\big)+\mu_{a,y}(X)$.
\end{proposition}

\begin{proof}[Proof of Proposition~\ref{cor:asymptotic_linfty}]

To show the weak convergence of the integral, first from Theorem~\ref{thm:asymptotic_smooth_density}
under the same condition implies 
\[
\left\{ \sqrt{n}\left(\hat{p}_{a,h}-p_{a,h}\right)\right\} _{a,y}\to\mathbb{G}\text{ weakly in }\ell^{\infty}(\mathcal{A}\times\mathbb{R}^{d})\text{ a.s.}.
\]
Then by restricting the domain to $\mathcal{Y}_{h}$, for each $a\in\mathcal{A}$,
under the same condition, we have 
\begin{equation}
\left\{ \sqrt{n}\left(\hat{p}_{a,h}-p_{a,h}\right)\right\} _{y}\to\tilde{\mathbb{G}}\text{ weakly in }\ell^{\infty}(\mathcal{Y}_{h})\text{ a.s.},\label{eq:asymptotic_integrated_pointwise_a-1}
\end{equation}
where $\tilde{\mathbb{G}}(y)=\mathbb{G}(a,y)$. Now, consider the
supreme functional $\Phi:\ell^{\infty}(\mathcal{Y}_{h})\to\mathbb{R}$
defined as $\Phi(f)=\sup_{y\in\mathcal{Y}_{h}}\left|f(y)\right|$,
then $\Phi$ is continuous on $\ell^{\infty}(\mathcal{Y}_{h})$. Hence,
from \eqref{eq:asymptotic_integrated_pointwise_a-1} and $\Phi$ being
continuous, the continuous mapping theorem \citep[e.g.,][Theorem 7.7]{Kosorok2008}
implies 
\[
\sqrt{n}\left\Vert \hat{p}_{a,h}-p_{a,h}\right\Vert _{\infty}=\Phi\left(\left\{ \sqrt{n}\left(\hat{p}_{a,h}(y)-p_{a,h}(y)\right)\right\} _{y}\right)\to\Phi(\tilde{\mathbb{G}})=\sup_{y\in\mathcal{Y}_{h}}\left|\mathbb{G}(a,y)\right|\text{ weakly in }\mathbb{R}\text{ a.s.}.
\]
\end{proof}

\begin{proposition} \label{cor:asymptotic_linfty_bootstrap} Assume
that Assumptions \ref{assumption:A2} - \ref{assumption:A4} and
\ref{assumption:kernel_vc} hold. Then for a given $h$ and each $a\in\mathcal{A}$,
we have the following weak convergence as 
\[
\sqrt{n}\left\Vert \widehat{p}_{a,h}^{*}-\hat{p}_{a,h}\right\Vert _{\infty}\to\sup_{y\in\mathcal{Y}_{h}}\left|\mathbb{G}(a,y)\right|\text{ weakly in }\mathbb{R}\text{ a.s.},
\]
where $\mathbb{G}$ is a centered Gaussian process with $Cov\left[\mathbb{G}(a_{1},y_{1}),\mathbb{G}(a_{2},y_{2})\right]=\mathbb{E}\left[f_{y_{1}}^{a_{1}}f_{y_{2}}^{a_{2}}\right]-\mathbb{E}\left[f_{y_{1}}^{a_{1}}\right]\mathbb{E}\left[f_{y_{2}}^{a_{2}}\right]$,
$f_{y}^{a}(Z)\coloneqq\frac{\mathbf{1}(A=a)}{\pi_{a}(X)}\big(K_{h,y}(Y)-\mu_{A,y}(X)\big)+\mu_{a,y}(X)$.
\end{proposition}

\begin{proof}[Proof of Proposition~\ref{cor:asymptotic_linfty_bootstrap}]

To show the weak convergence of the integral, first from Proposition~\ref{cor:asymptotic_smooth_density_bootstrap}
under the same condition implies 
\[
\left\{ \sqrt{n}\left(\hat{p}_{a,h}^{*}-\hat{p}_{a,h}\right)\right\} _{a,y}\to\mathbb{G}\text{ weakly in }\ell^{\infty}(\mathcal{A}\times\mathbb{R}^{d})\text{ a.s.}.
\]
Then by restricting the domain to $\mathcal{Y}_{h}$, for each $a\in\mathcal{A}$,
under the same condition, we have 
\begin{equation}
\left\{ \sqrt{n}\left(\hat{p}_{a,h}^{*}-\hat{p}_{a,h}\right)\right\} _{y}\to\tilde{\mathbb{G}}\text{ weakly in }\ell^{\infty}(\mathcal{Y}_{h})\text{ a.s.},\label{eq:asymptotic_integrated_pointwise_a_bootstrap-1}
\end{equation}
where $\tilde{\mathbb{G}}(y)=\mathbb{G}(a,y)$. Now, consider the
supreme functional $\Phi:\ell^{\infty}(\mathcal{Y}_{h})\to\mathbb{R}$
defined as $\Phi(f)=\sup_{y\in\mathcal{Y}_{h}}\left|f(y)\right|$,
then $\Phi$ is continuous on $\ell^{\infty}(\mathcal{Y}_{h})$. Hence,
from \eqref{eq:asymptotic_integrated_pointwise_a_bootstrap-1} and
$\Phi$ being continuous, the continuous mapping theorem \citep[e.g.,][Theorem 7.7]{Kosorok2008}
implies 
\[
\sqrt{n}\left\Vert \widehat{p}_{a,h}^{*}-\hat{p}_{a,h}\right\Vert _{\infty}=\Phi\left(\left\{ \sqrt{n}\left(\hat{p}_{a,h}^{*}(y)-\hat{p}_{a,h}(y)\right)\right\} _{y}\right)\to\Phi(\tilde{\mathbb{G}})=\sup_{y\in\mathcal{Y}_{h}}\left|\mathbb{G}(a,y)\right|\text{ weakly in }\mathbb{R}\text{ a.s.}.
\]
\end{proof}

Finally, Corollary~\ref{cor:confidence_smooth_density} immediately follows by Propositions~\ref{cor:asymptotic_linfty} and \ref{cor:asymptotic_linfty_bootstrap}.

\begin{proof}[Proof of Corollary~\ref{cor:confidence_smooth_density}]

To show the validity of the bootstrap confidence interval, we first
note that $\hat{z}_{\alpha}$ in Algorithm \ref{algorithm:density-band}
satisfies that for each $a\in\mathcal{A}$, 
\[
\liminf_{n\to\infty}P\left(\sqrt{n}D(\hat{p}_{a,h}^{*},\hat{p}_{a,h})\leq\hat{z}_{\alpha}^{a}\right)\geq1-\alpha.
\]
Then by Propositions~\ref{cor:asymptotic_linfty} and \ref{cor:asymptotic_linfty_bootstrap},
given that $\mathbb{E}\left[(f_{y}^{a})^{2}\right]\in(0,\infty)$,
$\sqrt{n}D(\hat{p}_{a,h},p_{a,h})$ and $\sqrt{n}D(\hat{p}_{a,h}^{*},\hat{p}_{a,h})$
have the same asymptotic distribution limit that is nonsingular, and
hence for each $a\in\mathcal{A}$, 
\[
\liminf_{n\to\infty}P\left(\sqrt{n}D(\hat{p}_{a,h},p_{a,h})\leq\hat{z}_{\alpha}^{a}\right)\geq1-\alpha.
\]
And in particular, 
\[
\liminf_{n\to\infty}P\left(\left\Vert \hat{p}_{a,h}-p_{a,h}\right\Vert _{\infty}\leq\frac{\hat{z}_{\alpha}^{a}}{\sqrt{n}}\right)\geq1-\alpha.
\]

\end{proof}

\subsection{Proofs for Section \ref{subsec:inference-density-effect}}

\textbf{Results in Section \ref{subsubsec:inference-density-effect-cd}}

\begin{proof}[Proof of Corollary~\ref{cor:asymptotic_l1_one}]

First note that Theorem~\ref{thm:asymptotic_smooth_density} implies 
\[
\left\{ \sqrt{n}\left(\hat{p}_{a,h}-p_{a,h}\right)\right\} _{a,y}\to\mathbb{G}\text{ weakly in }\ell^{\infty}(\mathcal{A}\times\mathbb{R}^{d})\text{ a.s.}.
\]
Restricting the domain to $\mathcal{Y}_{h}$ yields
\begin{equation}
\left\{ \sqrt{n}\left(\hat{p}_{a,h}-p_{a,h}\right)\right\} _{y}\to\tilde{\mathbb{G}}\text{ weakly in }\ell^{\infty}(\mathcal{Y}_{h})\text{ a.s.},\label{eq:asymptotic_integrated_pointwise_a}
\end{equation}
where $\tilde{\mathbb{G}}(y)=\mathbb{G}(a,y)$. Now, consider the integral functional $\Phi:\ell^{\infty}(\mathcal{Y}_{h})\to\mathbb{R}$
defined as $\Phi(f)=\int_{\mathcal{Y}_{h}}\left|f(y)\right|dy$. Then
from the boundedness condition on $\mathcal{Y}$, $\mathcal{Y}_{h}$ is bounded as well, which implies that $\Phi$ is continuous
on $\ell^{\infty}(\mathcal{Y}_{h})$. Hence, from \eqref{eq:asymptotic_integrated_pointwise_a}
and $\Phi$ being continuous, the continuous mapping theorem \citep[e.g.,][Theorem 7.7]{Kosorok2008}
implies 
\[
\sqrt{n}D\left(\hat{p}_{a,h},p_{a,h}\right)=\Phi\left(\left\{ \sqrt{n}\left(\hat{p}_{a,h}(y)-p_{a,h}(y)\right)\right\} _{y}\right)\to\Phi(\tilde{\mathbb{G}})=\int_{\mathcal{Y}_{h}}\left|\mathbb{G}(a,y)\right|dy\text{ weakly in }\mathbb{R}\text{ a.s.}.
\]
\end{proof}

Before proving Corollaries \ref{cor:confidence_l1_one} and \ref{cor:asymptotic_l1_twodiff}, we give the following result.

\begin{corollary} \label{cor:asymptotic_l1_one_bootstrap} Assume
that Assumptions \ref{assumption:A2}, \ref{assumption:A3}, \ref{assumption:A4'}, and
\ref{assumption:kernel_vc} hold. Further assume that $\mathcal{Y}$ is bounded. Then for a given $h$ and each $a\in\mathcal{A}$,
we have the following weak convergence as 
\[
\sqrt{n}D\left(\widehat{p}_{a,h}^{*},\hat{p}_{a,h}\right)\to\int\left|\mathbb{G}(a,y)\right|dy\text{ weakly in }\mathbb{R}\text{ a.s.},
\]
where $\mathbb{G}$ is a centered Gaussian process with $Cov\left[\mathbb{G}(a_{1},y_{1}),\mathbb{G}(a_{2},y_{2})\right]=\mathbb{E}\left[f_{y_{1}}^{a_{1}}f_{y_{2}}^{a_{2}}\right]-\mathbb{E}\left[f_{y_{1}}^{a_{1}}\right]\mathbb{E}\left[f_{y_{2}}^{a_{2}}\right]$,
$f_{y}^{a}(Z)\coloneqq\frac{\mathbf{1}(A=a)}{\pi_{a}(X)}\big(K_{h,y}(Y)-\mu_{A,y}(X)\big)+\mu_{a,y}(X)$.
\end{corollary}

\begin{proof}[Proof of Corollary~\ref{cor:asymptotic_l1_one_bootstrap}]

To show the weak convergence of the integral, first from Corollary~\ref{cor:asymptotic_smooth_density_bootstrap}
under the same condition implies 
\[
\left\{ \sqrt{n}\left(\hat{p}_{a,h}^{*}-\hat{p}_{a,h}\right)\right\} _{a,y}\to\mathbb{G}\text{ weakly in }\ell^{\infty}(\mathcal{A}\times\mathbb{R}^{d})\text{ a.s.}.
\]
Then by restricting the domain to $\mathcal{Y}_{h}$, for each $a\in\mathcal{A}$,
under the same condition, we have 
\begin{equation}
\left\{ \sqrt{n}\left(\hat{p}_{a,h}^{*}-\hat{p}_{a,h}\right)\right\} _{y}\to\tilde{\mathbb{G}}\text{ weakly in }\ell^{\infty}(\mathcal{Y}_{h})\text{ a.s.},\label{eq:asymptotic_integrated_pointwise_a_bootstrap}
\end{equation}
where $\tilde{\mathbb{G}}(y)=\mathbb{G}(a,y)$. Now, consider the
integral functional $\Phi:\ell^{\infty}(\mathcal{Y}_{h})\to\mathbb{R}$
defined as $\Phi(f)=\int_{\mathcal{Y}_{h}}\left|f(y)\right|dy$, then
from $\mathcal{Y}$ being bounded, $\mathcal{Y}_{h}$ is bounded as well, and this implies
that $\Phi$ is continuous on $\ell^{\infty}(\mathcal{Y}_{h})$. Hence,
from \eqref{eq:asymptotic_integrated_pointwise_a_bootstrap} and $\Phi$
being continuous, the continuous mapping theorem \citep[e.g.,][Theorem 7.7]{Kosorok2008}
implies 
\[
\sqrt{n}D\left(\hat{p}_{a,h}^{*},\hat{p}_{a,h}\right)=\Phi\left(\left\{ \sqrt{n}\left(\hat{p}_{a,h}^{*}(y)-\hat{p}_{a,h}(y)\right)\right\} _{y}\right)\to\Phi(\tilde{\mathbb{G}})=\int_{\mathcal{Y}_{h}}\left|\mathbb{G}(a,y)\right|dy\text{ weakly in }\mathbb{R}\text{ a.s.}.
\]
\end{proof}

Now we provide the proofs of Corollaries \ref{cor:confidence_l1_one} and \ref{cor:asymptotic_l1_twodiff}.

\begin{proof}[Proof of Corollary~\ref{cor:confidence_l1_one}]

To show the validity of the bootstrap confidence interval, we first
note that $\hat{z}_{\alpha/2}^{a}$ in Algorithm \ref{algorithm:density-effect-cd1} satisfies that
for each $a\in\mathcal{A}$, 
\[
\liminf_{n\to\infty}P\left(\sqrt{n}D(\hat{p}_{a,h}^{*},\hat{p}_{a,h})\leq\hat{z}_{\alpha/2}^{a}\right)\geq1-\frac{\alpha}{2}.
\]
Then by Corollary~\ref{cor:asymptotic_l1_one} and \ref{cor:asymptotic_l1_one_bootstrap},
given that $\mathbb{E}\left[(f_{y}^{a})^{2}\right]\in(0,\infty)$,
$\sqrt{n}D(\hat{p}_{a,h},p_{a,h})$ and $\sqrt{n}D(\hat{p}_{a,h}^{*},\hat{p}_{a,h})$
have the same asymptotic distribution limit that is nonsingular, and
hence for each $a\in\mathcal{A}$,
\[
\liminf_{n\to\infty}P\left(\sqrt{n}D(\hat{p}_{a,h},p_{a,h})\leq\hat{z}_{\alpha/2}^{a}\right)\geq1-\frac{\alpha}{2},
\]
which implies that 
\begin{equation}
\liminf_{n\to\infty}P\left(D(\hat{p}_{0,h},p_{0,h})+D(\hat{p}_{1,h},p_{1,h})\leq\frac{\hat{z}_{\alpha/2}^{0}}{\sqrt{n}}+\frac{\hat{z}_{\alpha/2}^{1}}{\sqrt{n}}\right)\geq1-\alpha.\label{eq:confidence_l1_one_base}
\end{equation}
Now, for all $y\in\mathbb{R}^{d}$, we have
\[
\left|\left|\hat{p}_{1,h}(y)-\hat{p}_{0,h}(y)\right|-\left|p_{1,h}(y)-p_{0,h}(y)\right|\right|\leq\left|\hat{p}_{0,h}(y)-p_{0,h}(y)\right|+\left|\hat{p}_{1,h}(y)-p_{1,h}(y)\right|,
\]
and thus
\[
\left|D(\hat{p}_{1,h},\hat{p}_{0,h})-D(p_{1,h},p_{0,h})\right|\leq D(\hat{p}_{0,h},p_{0,h})+D(\hat{p}_{1,h},p_{1,h}).
\]
Combining this with \eqref{eq:confidence_l1_one_base}
gives 
\[
\liminf_{n\to\infty}P\left(\left|D(\hat{p}_{1,h},\hat{p}_{0,h})-D(p_{1,h},p_{0,h})\right|\leq\frac{\hat{z}_{\alpha/2}^{0}}{\sqrt{n}}+\frac{\hat{z}_{\alpha/2}^{1}}{\sqrt{n}}\right)\geq1-\alpha.
\]

\end{proof}

\begin{proof}[Proof of Corollary~\ref{cor:asymptotic_l1_twodiff}]

To show the weak convergence of the integral, first from 
Theorem~\ref{thm:asymptotic_smooth_density} under the same condition
implies that 
\[
\left\{ \sqrt{n}\left(\hat{p}_{a,h}-p_{a,h}\right)\right\} _{a,y}\to\mathbb{G}\text{ weakly in }\ell^{\infty}(\mathcal{A}\times\mathbb{R}^{d})\text{ a.s.}.
\]
Then by restricting the domain to $\mathcal{A}\times\mathcal{Y}_{h}$,
under the same condition, we have 
\begin{equation}
\left\{ \sqrt{n}\left(\hat{p}_{a,h}(y)-p_{a,h}(y)\right)\right\}_{a,y}\to\mathbb{G}\text{ weakly in }\ell^{\infty}(\mathcal{A}\times\mathcal{Y}_{h})\text{ a.s.}.\label{eq:asymptotic_integrated_pointwise}
\end{equation}
Now, consider the integral functional $\Phi:\ell^{\infty}(\mathcal{A}\times\mathcal{Y}_{h})\to\mathbb{R}$
defined as $\Phi(f)=\int_{\mathcal{Y}_{h}}\left|f(1,y)-f(0,y)\right|dy$,
then from $\mathcal{Y}$ being bounded, $\mathcal{Y}_{h}$ is bounded as well, and this implies that $\Phi$ is continuous
on $\ell^{\infty}(\mathcal{A}\times\mathcal{Y}_{h})$. Hence, from
\eqref{eq:asymptotic_integrated_pointwise} and $\Phi$ being continuous,
the continuous mapping theorem \citep[e.g.,][Theorem 7.7]{Kosorok2008}
implies 
\begin{align*}
 & \sqrt{n}D\left(\hat{p}_{h}^{1}-\hat{p}_{h}^{0},p_{h}^{1}-p_{h}^{0}\right)=\Phi\left(\left\{ \sqrt{n}\left(\hat{p}_{a,h}(y)-p_{a,h}(y)\right)\right\} _{a,y}\right)\\
 & \to\Phi(\mathbb{G})=\int_{\mathcal{Y}_{h}}\left|\mathbb{G}(1,y)-\mathbb{G}(0,y)\right|dy\text{ weakly in }\mathbb{R}\text{ a.s.}.
\end{align*}

\end{proof}

We give another weak convergence result
before the proof of Corollary~\ref{cor:confidence_l1_twodiff} as below.

\begin{corollary} \label{cor:asymptotic_l1_twodiff_bootstrap} Assume
that the same conditions of Corollary~\ref{cor:asymptotic_l1_one} hold. Then for a given $h$, we have the
following weak convergence as 
\[
\sqrt{n}D\left(\widehat{p}_{1,h}^{*}-\widehat{p}_{0,h}^{*},\hat{p}_{1,h}-\hat{p}_{0,h}\right)\to\int\left|\mathbb{G}(1,y)-\mathbb{G}(0,y)\right|dy\text{ weakly in }\mathbb{R}\text{ a.s.},
\]
where $\mathbb{G}$ is a centered Gaussian process with $Cov\left[\mathbb{G}(a_{1},y_{1}),\mathbb{G}(a_{2},y_{2})\right]=\mathbb{E}\left[f_{y_{1}}^{a_{1}}f_{y_{2}}^{a_{2}}\right]-\mathbb{E}\left[f_{y_{1}}^{a_{1}}\right]\mathbb{E}\left[f_{y_{2}}^{a_{2}}\right]$,
$f_{y}^{a}(Z)\coloneqq\frac{\mathbf{1}(A=a)}{{\pi}_{a}(X)}\big(K_{h,y}(Y)-\mu_{A,y}(X)\big)+\mu_{a,y}(X)$.
\end{corollary}

\begin{proof}[Proof of Corollary~\ref{cor:asymptotic_l1_twodiff_bootstrap}]

Under the given conditions,
Corollary~\ref{cor:asymptotic_smooth_density_bootstrap} implies 
\[
\left\{ \sqrt{n}\left(\hat{p}_{a,h}^{*}-\hat{p}_{a,h}\right)\right\} _{a,y}\to\mathbb{G}\text{ weakly in }\ell^{\infty}(\mathcal{A}\times\mathbb{R}^{d})\text{ a.s.}.
\]
Then by restricting the domain to $\mathcal{A}\times\mathcal{Y}_{h}$,
we have 
\begin{equation}
\left\{ \sqrt{n}\left(\hat{p}_{a,h}^{*}(y)-\hat{p}_{a,h}(y)\right)\right\} _{a,y}\to\mathbb{G}\text{ weakly in }\ell^{\infty}(\mathcal{A}\times\mathcal{Y}_{h})\text{ a.s.}.\label{eq:asymptotic_integrated_pointwise_bootstrap}
\end{equation}
Now, consider the integral functional $\Phi:\ell^{\infty}(\mathcal{A}\times\mathcal{Y}_{h})\to\mathbb{R}$
defined as $\Phi(f)=\int_{\mathcal{Y}_{h}}\left|f(1,y)-f(0,y)\right|dy$.
Since $\mathcal{Y}$ is bounded, $\mathcal{Y}_{h}$ is bounded as well. Hence, $\Phi$ is continuous on $\ell^{\infty}(\mathcal{A}\times\mathcal{Y}_{h})$.

Now that we have \eqref{eq:asymptotic_integrated_pointwise_bootstrap}
and $\Phi$ being continuous, by the continuous mapping theorem \citep[e.g.,][Theorem 7.7]{Kosorok2008}
it follows that 
\begin{align*}
 & \sqrt{n}D\left(\hat{p}_{1,h}^{*}-\hat{p}_{0,h}^{*},\hat{p}_{1,h}-\hat{p}_{0,h}\right)=\Phi\left(\left\{ \sqrt{n}\left(\hat{p}_{a,h}^{*}(y)-\hat{p}_{a,h}(y)\right)\right\} _{a,y}\right)\\
 & \to\Phi(\mathbb{G})=\int_{\mathcal{Y}_{h}}\left|\mathbb{G}(1,y)-\mathbb{G}(0,y)\right|dy\text{ weakly in }\mathbb{R}\text{ a.s.}.
\end{align*}

\end{proof}

\begin{proof}[Proof of Corollary~\ref{cor:confidence_l1_twodiff}]

To show the validity of the bootstrap confidence interval, we first
note that $\hat{z}_{\alpha}$ in Algorithm \ref{algorithm:density-effect-cd2} satisfies that 
\[
\liminf_{n\to\infty}P\left(\sqrt{n}D(\hat{p}_{1,h}^{*}-\hat{p}_{0,h}^{*},\hat{p}_{1,h}-\hat{p}_{0,h})\leq\hat{z}_{\alpha}\right)\geq1-\alpha.
\]
Then by Corollary~\ref{cor:asymptotic_l1_twodiff} and \ref{cor:asymptotic_l1_twodiff_bootstrap}, given that $\mathbb{E}\left[(f_{y}^{1}-f_{y}^{0})^{2}\right]\in(0,\infty)$, it follows that
$\sqrt{n}D(\hat{p}_{1,h}-\hat{p}_{0,h},p_{1,h}-p_{0,h})$ and $\sqrt{n}D(\hat{p}_{1,h}^{*}-\hat{p}_{0,h}^{*},\hat{p}_{1,h}-\hat{p}_{0,h})$
have the same asymptotic distribution limit that is nonsingular, and
hence
\[
\liminf_{n\to\infty}P\left(\sqrt{n}D(\hat{p}_{1,h}-\hat{p}_{0,h},p_{1,h}-p_{0,h})\leq\hat{z}_{\alpha}\right)\geq1-\alpha.
\]
Then this implies that 
\begin{equation}
\liminf_{n\to\infty}P\left(D(\hat{p}_{1,h}-\hat{p}_{0,h},p_{1,h}-p_{0,h})\leq\frac{\hat{z}_{\alpha}}{\sqrt{n}}\right)\geq1-\alpha.\label{eq:confidence_l1_twodiff_base}
\end{equation}
Now, for all $y\in\mathbb{R}^{d}$, the following inequality holds:
\[
\left|\left|\hat{p}_{1,h}(y)-\hat{p}_{0,h}(y)\right|-\left|p_{1,h}(y)-p_{0,h}(y)\right|\right|\leq\left|\left(\hat{p}_{1,h}(y)-\hat{p}_{0,h}(y)\right)-\left(p_{1,h}(y)-p_{0,h}(y)\right)\right|.
\]
Therefore, 
\[
\left|D(\hat{p}_{1,h},\hat{p}_{0,h})-D(p_{1,h},p_{0,h})\right|\leq D(\hat{p}_{1,h}-\hat{p}_{0,h},p_{1,h}-p_{0,h}).
\]
 holds. And combining this with \eqref{eq:confidence_l1_twodiff_base}
gives 
\[
\liminf_{n\to\infty}P\left(\left|D(\hat{p}_{1,h},\hat{p}_{0,h})-D(p_{1,h},p_{0,h})\right|\leq\frac{\hat{z}_{\alpha}}{\sqrt{n}}\right)\geq1-\alpha.
\]
\end{proof}




\textbf{Results in Section \ref{subsubsec:inference-density-effect-sm}}

Before the proof of Corollary~\ref{cor:asymptotic_smooth_l1}, we make the following brief claim.

\begin{claim} \label{claim:asymptotic_smooth_l1_weakly} Suppose
$\sigma_{sm}^{2}\coloneqq\mathbb{E}\left[\phi_{s}^{sm}(Z;\eta^{sm})^{2}\right]<\infty$,
then 
\begin{align*}
\sqrt{n}(\mathbb{P}_{n}-\mathbb{P})\phi_{s}^{sm}(\cdot;\eta^{sm}) & \to\mathcal{N}(0,\sigma_{sm}^{2})\text{ weakly},\\
\sqrt{n}(\mathbb{P}_{n}^{*}-\mathbb{P}_{n})\phi_{s}^{sm}(\cdot;\eta^{sm}) & \to\mathcal{N}(0,\sigma_{sm}^{2})\text{ weakly a.s.}.
\end{align*}

\end{claim}

\begin{proof}

From $\sigma_{sm}^{2}=\mathbb{E}\left[\phi_{s}^{sm}(Z;\eta^{sm})^{2}\right]<\infty$,
by the central limit theorem it follows
\[
\sqrt{n}(\mathbb{P}_{n}-\mathbb{P})\phi_{s}^{sm}(\cdot;\eta^{sm})\to\mathcal{N}(0,\sigma_{sm}^{2})\text{ weakly}.
\]
Then Proposition~\ref{prop:empirical_donsker_bootstrap} implies that
\[
\sqrt{n}(\mathbb{P}_{n}^{*}-\mathbb{P}_{n})\phi_{s}^{sm}(\cdot;\eta^{sm})\to\mathcal{N}(0,\sigma_{sm}^{2})\text{ weakly a.s.}.
\]

\end{proof}

The proof of Corollary~\ref{cor:asymptotic_smooth_l1} follows immediately by Theorem~\ref{thm:error-bound-psi-sm} and Claim \ref{claim:asymptotic_smooth_l1_weakly} as given below.

\begin{proof}[Proof of Corollary~\ref{cor:asymptotic_smooth_l1}]

Under Assumptions \ref{assumption:A2}, \ref{assumption:A3},
\ref{assumption:A5}, from Theorem~\ref{thm:error-bound-psi-sm},
\begin{align*}
 & \left|\sqrt{n}\left(\widehat{\psi}_{s}^{sm}-\psi_{s}^{sm}\right)-\sqrt{n}(\mathbb{P}_{n}-\mathbb{P})\phi_{s}^{sm}(\cdot;\eta^{sm})\right|\\
 & =O_{\mathbb{P}}\left(\sqrt{n}\sum_{a}\left\{ \Vert h'_{s}\Vert_{\infty}\Vert\widehat{\pi}_{a}-\pi_{a}\Vert\Vert\widehat{\nu}_{a}-\nu_{a}\Vert+\Vert h''_{s}\Vert_{\infty}\Vert\widehat{p}_{a}-p_{a}\Vert^{2}\right\} \right).
\end{align*}
Hence under the conditions $\Vert h'_{s}\Vert_{\infty},\Vert h''_{s}\Vert_{\infty}<\infty$,
$\Vert\widehat{\pi}_{a}-\pi_{a}\Vert\Vert\widehat{\nu}_{a}-\nu_{a}\Vert=o_{\mathbb{P}}\left(n^{-\frac{1}{2}}\right)$,
and $\Vert\widehat{p}_{a}-p_{a}\Vert=o_{\mathbb{P}}\left(n^{-\frac{1}{4}}\right)$,
\begin{equation}
\left|\sqrt{n}\left(\widehat{\psi}_{s}^{sm}-\psi_{s}^{sm}\right)-\sqrt{n}(\mathbb{P}_{n}-\mathbb{P})\varphi_{s}^{sm}(\cdot;\eta^{sm})\right|=o_{\mathbb{P}}(1).\label{eq:asymptotic_smooth_l1_op1}
\end{equation}
Now under the condition $\sigma_{sm}^{2}=\mathbb{E}\left[\phi_{s}^{sm}(Z;\eta^{sm})^{2}\right]<\infty$,
Claim~\ref{claim:asymptotic_smooth_l1_weakly} implies 
\begin{equation}
\sqrt{n}(\mathbb{P}_{n}-\mathbb{P})\phi_{s}^{sm}(\cdot;\eta^{sm})\to\mathcal{N}(0,\sigma_{sm}^{2})\text{ weakly},\label{eq:asymptotic_smooth_l1_normal}
\end{equation}
where $\sigma_{sm}^{2}=\mathbb{E}\left[\phi_{s}^{sm}(Z;\eta^{sm})^{2}\right]$.
Hence combining \eqref{eq:asymptotic_smooth_l1_op1} and \eqref{eq:asymptotic_smooth_l1_normal}
with Slutsky Theorem \citep[][Theorem 7.15]{Kosorok2008} provides
the weak convergence of \eqref{eq:asymptotic_smooth_l1} as 
\[
\sqrt{n}\left(\hat{\psi}_{s}^{sm}-\psi_{s}^{sm}\right)\to\mathcal{N}(0,\sigma_{sm}^{2})\text{ weakly}.
\]

\end{proof}


The following weak convergence is required to prove Corollary \ref{cor:confidence_smooth_l1}.

\begin{corollary} \label{cor:asymptotic_smooth_l1_bootstrap} Suppose
that the assumptions \ref{assumption:A2}, \ref{assumption:A3}, \ref{assumption:A4''},
\ref{assumption:A5} hold. Then we have
the following weak convergence as 
\begin{equation}
\sqrt{n}\left(\left(\hat{\psi}_{s}^{sm}\right)^{*}-\hat{\psi}_{s}^{sm}\right)\to\mathcal{N}(0,\sigma_{sm}^{2})\text{ weakly a.s.},\label{eq:asymptotic_smooth_l1_bootstrap}
\end{equation}
where $\sigma_{sm}^{2}=\mathbb{E}\left[\phi_{s}^{sm}(Z;\eta^{sm})^{2}\right]<\infty$.

\end{corollary}

\begin{proof} 
Since $\hat{\psi}_{s}^{sm}(Z;\widehat{\eta}^{sm},\{\hat{p}_{a}\})=\mathbb{P}_{n}\phi_{s}^{sm}(\cdot;\widehat{\eta}^{sm},\{\hat{p}_{a}\})$
and $(\hat{\psi}_{s}^{sm})^{*}(Z;\widehat{\eta}^{sm},\{\hat{p}_{a}\})=\mathbb{P}_{n}^{*}\phi_{s}^{sm}(\cdot;\widehat{\eta}^{sm},\{\hat{p}_{a}\})$, Claim~\ref{claim:error-psi-sm-expand} gives 
\begin{equation}
\sqrt{n}\left(\left(\hat{\psi}_{s}^{sm}\right)^{*}-\hat{\psi}_{s}^{sm}\right)=\sqrt{n}(\mathbb{P}_{n}^{*}-\mathbb{P}_{n})\phi_{s}^{sm}(\cdot;{\eta}^{sm})+\sqrt{n}\mathbb{P}_{n}^{*}\left\{ \phi_{s}^{sm}(\cdot;\widehat{\eta}^{sm})-\phi_{s}^{sm}(\cdot;{\eta}^{sm})\right\}. \label{eq:asymptotic_smooth_l1_bootstrap_expand}
\end{equation}
Now under the condition $\sigma_{sm}^{2}=\mathbb{E}\left[\phi_{s}^{sm}(Z;\eta^{sm})^{2}\right]<\infty$,
Claim~\ref{claim:asymptotic_smooth_l1_weakly} implies 
\begin{equation}
\sqrt{n}(\mathbb{P}_{n}^{*}-\mathbb{P}_{n})\phi_{s}^{sm}(\cdot;{\eta}^{sm})\to\mathcal{N}(0,\sigma_{sm}^{2})\text{ weakly},\label{eq:asymptotic_smooth_l1_bootstrap_normal}
\end{equation}
where $\sigma_{sm}^{2}=\mathbb{E}\left[\phi_{s}^{sm}(Z;{\eta}^{sm})^{2}\right]$. And under the assumptions \ref{assumption:A2}, \ref{assumption:A3},
\ref{assumption:A4''}, \ref{assumption:A5}, Corollary~\ref{cor:sm-empirical-difference-measure}
implies
\begin{equation}
\sqrt{n}\mathbb{P}_{n}^{*}\left\{ \phi_{s}^{sm}(\cdot;\widehat{\eta}^{sm})-\phi_{s}^{sm}(\cdot;{\eta}^{sm})\right\} \to0\ \text{in probability a.s.}.\label{eq:asymptotic_smooth_l1_bootstrap_op1}
\end{equation}
Hence applying \eqref{eq:asymptotic_smooth_l1_bootstrap_normal} and
\eqref{eq:asymptotic_smooth_l1_bootstrap_op1} with Slutsky Theorem
\citep[][Theorem 7.15]{Kosorok2008} to \eqref{eq:asymptotic_smooth_l1_bootstrap_expand}
provides the weak convergence of \eqref{eq:asymptotic_smooth_l1}
as 
\[
\sqrt{n}\left(\left(\hat{\psi}_{s}^{sm}\right)^{*}-\hat{\psi}_{s}^{sm}\right)\to\mathcal{N}(0,\sigma_{sm}^{2})\text{ weakly a.s.}.
\]

\end{proof}

Corollary \ref{cor:confidence_smooth_l1} follows by Corollary~\ref{cor:asymptotic_smooth_l1}
and \ref{cor:asymptotic_smooth_l1_bootstrap}.

\begin{proof}[Proof of Corollary~\ref{cor:confidence_smooth_l1}]

To show the validity of the bootstrap confidence interval, we first
note that $\hat{z}_{\alpha}$ in Algorithm \ref{algorithm:density-effect-sm}
satisfies that, 
\[
\liminf_{n\to\infty}P\left(\sqrt{n}\left|\left(\hat{\psi}_{s}^{sm}\right)^{*}-\hat{\psi}_{s}^{sm}\right|\leq\hat{z}_{\alpha}\right)\geq1-\alpha.
\]
Then by Corollary~\ref{cor:asymptotic_smooth_l1} and \ref{cor:asymptotic_smooth_l1_bootstrap},
given that $\sigma_{sm}^{2}\coloneqq\mathbb{E}\left[\phi_{s}^{sm}(Z;\eta^{sm})^{2}\right]\in(0,\infty)$,
$\sqrt{n}\left(\hat{\psi}_{s}^{sm}-\psi_{s}^{sm}\right)$ and $\sqrt{n}\left(\left(\hat{\psi}_{s}^{sm}\right)^{*}-\hat{\psi}_{s}^{sm}\right)$
have the same asymptotic distribution limit that is nonsingular, and
hence, 
\[
\liminf_{n\to\infty}P\left(\sqrt{n}\left|\hat{\psi}_{s}^{sm}-\psi_{s}^{sm}\right|\leq\hat{z}_{\alpha}\right)\geq1-\alpha.
\]
And in particular, 
\[
\liminf_{n\to\infty}P\left(\left|\hat{\psi}_{s}^{sm}-\psi_{s}^{sm}\right|\leq\frac{\hat{z}_{\alpha}}{\sqrt{n}}\right)\geq1-\alpha.
\]

\end{proof}

\textbf{Results in Section \ref{subsubsec:inference-density-effect-mc}}

The following claim is analogous to Claim \ref{claim:asymptotic_smooth_l1_weakly}.

\begin{claim} \label{claim:asymptotic_margin_weakly} Suppose $\sigma_{mc}^{2}\coloneqq\mathbb{E}\left[\varphi^{mc}(Z;\gamma^{mc},\eta^{mc})^{2}\right]<\infty$,
then
\begin{align*}
\sqrt{n}(\mathbb{P}_{n}-\mathbb{P})\varphi^{mc}(\cdot;\gamma^{mc},\eta^{mc}) & \to\mathcal{N}(0,\sigma_{mc}^{2})\text{ weakly},\\
\sqrt{n}(\mathbb{P}_{n}^{*}-\mathbb{P}_{n})\varphi^{mc}(\cdot;\gamma^{mc},\eta^{mc}) & \to\mathcal{N}(0,\sigma_{mc}^{2})\text{ weakly a.s.}.
\end{align*}

\end{claim}

\begin{proof}

From $\sigma_{mc}^{2}=\mathbb{E}\left[\varphi^{mc}(Z;\gamma^{mc},\eta^{mc})^{2}\right]<\infty$, by the central limit theorem we have
\[
\sqrt{n}(\mathbb{P}_{n}-\mathbb{P})\varphi^{mc}(\cdot;\gamma^{mc},\eta^{mc})\to\mathcal{N}(0,\sigma_{mc}^{2})\text{ weakly}.
\]
Then Proposition~\ref{prop:empirical_donsker_bootstrap} implies that 
\[
\sqrt{n}(\mathbb{P}_{n}^{*}-\mathbb{P}_{n})\varphi^{mc}(\cdot;\gamma^{mc},\eta^{mc})\to\mathcal{N}(0,\sigma_{mc}^{2})\text{ weakly a.s.}.
\]

\end{proof}

Corollary~\ref{cor:asymptotic_margin} follows by Theorem~\ref{thm:error-bound-psi-mc} and Claim \ref{claim:asymptotic_smooth_l1_weakly}.

\begin{proof}[Proof of Corollary~\ref{cor:asymptotic_margin}]

Under the assumptions \ref{assumption:A2}, \ref{assumption:A3},
\ref{assumption:A6} from Theorem~\ref{thm:error-bound-psi-mc},
\begin{align*}
 & \left|\sqrt{n}\left(\widehat{\psi}^{mc}-\psi\right)-\sqrt{n}(\mathbb{P}_{n}-\mathbb{P})\varphi^{mc}(\cdot;\gamma^{mc},\eta^{mc})\right|\\
 & =O_{\mathbb{P}}\left(\sqrt{n}\sum_{a}
 \left\{ \Vert\widehat{\pi}_{a}-\pi_{a}\Vert\Vert\widehat{\nu}_{a}-\nu_{a}\Vert + \left\Vert\widehat{p}_{a}-p_{a}\right\Vert_{\infty}^{\alpha+1}\right\}
 \right)+o_{\mathbb{P}}(1).
\end{align*}
Hence under the conditions $\Vert\widehat{\pi}_{a}-\pi_{a}\Vert\Vert\widehat{\nu}_{a}-\nu_{a}\Vert=o_{\mathbb{P}}\left(n^{-\frac{1}{2}}\right)$
and $\Vert\widehat{p}_{a}-p_{a}\Vert=o_{\mathbb{P}}\left(n^{-\frac{1}{2\alpha}}\right)$
for each $a\in \mathcal{A}$, 
\begin{equation}
\left|\sqrt{n}\left(\widehat{\psi}^{mc}-\psi\right)-\sqrt{n}(\mathbb{P}_{n}-\mathbb{P})\varphi^{mc}(\cdot;\gamma^{mc},\eta^{mc})\right|=o_{\mathbb{P}}(1).\label{eq:asymptotic_margin_op1}
\end{equation}
Now under the condition $\sigma_{mc}^{2}=\mathbb{E}\left[\varphi^{mc}(Z;\gamma^{mc},\eta^{mc})^{2}\right]<\infty$,
Claim~\ref{claim:asymptotic_margin_weakly} implies 
\begin{equation}
\sqrt{n}(\mathbb{P}_{n}-\mathbb{P})\varphi^{mc}(\cdot;\gamma^{mc},\eta^{mc})\to\mathcal{N}(0,\sigma_{mc}^{2})\text{ weakly},\label{eq:asymptotic_margin_normal}
\end{equation}
where $\sigma_{mc}^{2}=\mathbb{E}\left[\varphi^{mc}(Z;\gamma^{mc},\eta^{mc})^{2}\right]$.
Hence combining \eqref{eq:asymptotic_margin_op1} and \eqref{eq:asymptotic_margin_normal}
with Slutsky Theorem \citep[][Theorem 7.15]{Kosorok2008} provides
the weak convergence of \eqref{eq:asymptotic_margin} as 
\[
\sqrt{n}\left(\widehat{\psi}^{mc}-\psi\right)\to\mathcal{N}(0,\sigma_{mc}^{2})\text{ weakly}.
\]
\end{proof}


The following weak convergence is required for the proof of Corollary \ref{cor:confidence_smooth_margin}.

\begin{corollary} \label{cor:asymptotic_margin_bootstrap} Suppose
that the assumptions \ref{assumption:A2}, \ref{assumption:A3}, \ref{assumption:A4'''}, and 
\ref{assumption:A6} hold. Then we have
the following weak convergence as 
\begin{equation}
\sqrt{n}\left(\left(\widehat{\psi}^{mc}\right)^{*}-\widehat{\psi}^{mc}\right)\to\mathcal{N}(0,\sigma_{mc}^{2})\text{ weakly a.s.},\label{eq:asymptotic_margin_bootstrap}
\end{equation}
where $\sigma_{mc}^{2}=\mathbb{E}\left[\varphi^{mc}(Z;\gamma^{mc},\eta^{mc})^{2}\right]<\infty$.
\end{corollary}

\begin{proof} 
Since $\widehat{\psi}^{mc}(Z;\widehat{\gamma},\widehat{\eta}^{mc})=\mathbb{P}_{n}\varphi^{mc}(\cdot;\widehat{\gamma},\widehat{\eta}^{mc})$
and $(\widehat{\psi}^{mc})^{*}(Z;\widehat{\gamma},\widehat{\eta}^{mc})=\mathbb{P}_{n}^{*}\varphi^{mc}(\cdot;\widehat{\gamma},\widehat{\eta}^{mc})$,
applying Claim~\ref{claim:error-psi-mc-expand} gives the expansion
as 
\begin{align}
 & \sqrt{n}\left(\left(\widehat{\psi}^{mc}\right)^{*}-\widehat{\psi}^{mc}\right) = \sqrt{n}(\mathbb{P}_{n}^{*}-\mathbb{P}_{n})\varphi^{mc}(\cdot;\gamma,{\eta}^{mc})\nonumber \\
 & \qquad + \sqrt{n}\mathbb{P}_{n}^{*}\left\{ \varphi^{mc}(\cdot;\widehat{\gamma},\widehat{\eta}^{mc})-\varphi^{mc}(\cdot;\gamma,\widehat{\eta}^{mc})\right\} + \sqrt{n}\mathbb{P}_{n}^{*}\left\{ \varphi^{mc}(\cdot;\gamma,\widehat{\eta}^{mc})-\varphi^{mc}(\cdot;\gamma,{\eta}^{mc})\right\} .\label{eq:asymptotic_margin_bootstrap_expand}
\end{align}
Then under the condition $\sigma_{mc}^{2}=\mathbb{E}\left[\varphi^{mc}(Z;\gamma^{mc},\eta^{mc})^{2}\right]<\infty$,
Claim~\ref{claim:asymptotic_margin_weakly} implies 
\begin{equation}
\sqrt{n}(\mathbb{P}_{n}^{*}-\mathbb{P}_{n})\varphi^{mc}(\cdot;\gamma,{\eta}^{mc})\to\mathcal{N}(0,\sigma_{mc}^{2})\text{ weakly a.s.},\label{eq:asymptotic_margin_bootstrap_normal}
\end{equation}
where $\sigma_{mc}^{2}=\mathbb{E}\left[\varphi^{mc}(Z;\gamma,{\eta}^{mc})^{2}\right]$.

Now, Corollary~\ref{cor:error-psi-mc-empirical-2} and Corollary~\ref{cor:error-psi-mc-empirical-1} imply
\begin{equation}
\sqrt{n}\mathbb{P}_{n}^{*}\left\{ \varphi^{mc}(\cdot;\widehat{\gamma},\widehat{\eta}^{mc})-\varphi^{mc}(\cdot;\gamma,\widehat{\eta}^{mc})\right\} \to 0\ \text{in probability a.s.}, \label{eq:asymptotic_margin_bootstrap_op1-1}
\end{equation}
and
\begin{equation}
\sqrt{n}\mathbb{P}_{n}^{*}\left\{ \varphi^{mc}(\cdot;\gamma,\widehat{\eta}^{mc})-\varphi^{mc}(\cdot;\gamma,{\eta}^{mc})\right\} \to 0\ \text{in probability a.s.}, \label{eq:asymptotic_margin_bootstrap_op1-2}
\end{equation}
respectively. 

Hence applying \eqref{eq:asymptotic_margin_bootstrap_normal}, \eqref{eq:asymptotic_margin_bootstrap_op1-1},
and \eqref{eq:asymptotic_margin_bootstrap_op1-2} with Slutsky's theorem
\citep[][Theorem 7.15]{Kosorok2008} to \eqref{eq:asymptotic_margin_bootstrap_expand}
provides the weak convergence of \eqref{eq:asymptotic_margin} as
\[
\sqrt{n}\left(\left(\widehat{\psi}^{mc}\right)^{*}-\widehat{\psi}\right)\to\mathcal{N}(0,\sigma_{sm}^{2})\text{ weakly a.s.}.
\]
\end{proof}

Corollary \ref{cor:confidence_smooth_margin} follows by Corollary~\ref{cor:asymptotic_margin}
and \ref{cor:asymptotic_margin_bootstrap}.

\begin{proof}[Proof of Corollary~\ref{cor:confidence_smooth_margin}]

To show the validity of the bootstrap confidence interval, we first
note that $\hat{z}_{\alpha}$ in Algorithm \ref{algorithm:density-effect-mc}
satisfies that, 
\[
\liminf_{n\to\infty}P\left(\sqrt{n}\left|\left(\hat{\psi}^{mc}\right)^{*}-\hat{\psi}^{mc}\right|\leq\hat{z}_{\alpha}\right)\geq1-\alpha.
\]
Then by Corollary~\ref{cor:asymptotic_margin} and \ref{cor:asymptotic_margin_bootstrap},
given that $\sigma_{mc}^{2}\coloneqq\mathbb{E}\left[\varphi^{mc}(Z;\gamma^{mc},\eta^{mc})^{2}\right]\in(0,\infty)$,
$\sqrt{n}\left(\widehat{\psi}^{mc}-\psi\right)$ and $\sqrt{n}\left(\left(\hat{\psi}^{mc}\right)^{*}-\hat{\psi}^{mc}\right)$
have the same asymptotic distribution limit that is nonsingular, and
hence, 
\[
\liminf_{n\to\infty}P\left(\sqrt{n}\left|\widehat{\psi}^{mc}-\psi\right|\leq\hat{z}_{\alpha}\right)\geq1-\alpha.
\]
And in particular, 
\[
\liminf_{n\to\infty}P\left(\left|\widehat{\psi}^{mc}-\psi\right|\leq\frac{\hat{z}_{\alpha}}{\sqrt{n}}\right)\geq1-\alpha.
\]

\end{proof}

\textbf{Results in Section \ref{subsubsec:testing-no-effect}}




\begin{proof}[Proof of Proposition \ref{prop:testing-the-null}]

Suppose the support of $K$ is contained in $\mathcal{B}(0,R)$, i.e.,
$K(x)=0$ if $\left\Vert x\right\Vert \geq R$. And since $K$ is
continuous at $0$ with $K(0)>0$, there exists $R_{0}\in(0,R)$ such
that for all $x\in\mathcal{B}(0,R_{0})$, $K(x)\geq\frac{1}{2}K(0)>0$.

Now without loss of generality, suppose $p(x_{0})<q(x_{0})$ for some
$x_{0}\in\mathbb{R}^{d}$. Since $q-p$ is a continuous at $x_{0}$
with $(q-p)(x_{0})>0$, similarly there exists $r_{0}>0$ such that
for all $x\in\mathcal{B}(x_{0,}r_{0})$, $(q-p)(x)\geq\frac{1}{2}(q-p)(x_{0})$.
Now let $h_{0}:=\frac{r_{0}}{R}$, and fix any $h\leq h_{0}$. Then
first, $\left\Vert \frac{x-x_{0}}{h}\right\Vert \leq R$ is equivalent
to $x\in\mathcal{B}(x_{0},hR)$, so 

\begin{align*}
q_{h}(x_{0})-p_{h}(x_{0}) & =\int_{\mathbb{R}^{d}}\frac{1}{h^{d}}K\left(\frac{x-x_{0}}{h}\right)(q-p)(x)dx\\
 & =\frac{1}{h^{d}}\int_{\mathcal{B}(x_{0},hR)}K\left(\frac{x-x_{0}}{h}\right)(q-p)(x)dx.
\end{align*}
Then from $h\leq h_{0}=\frac{r_{0}}{R}$, $\mathcal{B}(x_{0},hR)\subset\mathcal{B}(x_{0},r_{0})$
holds, and $K\left(\frac{x-x_{0}}{h}\right)(q-p)(x)$ is nonnegative
on $\mathcal{B}(x_{0},hR)$. Hence RHS is further lower bounded as
\begin{align*}
 & \frac{1}{h^{d}}\int_{\mathcal{B}(x_{0},hR)}K\left(\frac{x-x_{0}}{h}\right)(q-p)(x)dx\\
 & \geq\frac{1}{h^{d}}\int_{\mathcal{B}(x_{0},hR_{0})}K\left(\frac{x-x_{0}}{h}\right)(q-p)(x)dx\\
 & \geq\frac{1}{4h^{d}}\int_{\mathcal{B}(x_{0},hR_{0})}K(0)(q-p)(x_{0})dx\\
 & =\frac{1}{4}R_{0}^{d}\omega_{d}K(0)(q-p)(x_{0})>0,
\end{align*}
where $\omega_{d}$ is the volume of the $d$-dimensional ball of
radius $1$. Therefore, $p_{h}(x_{0})<q_{h}(x_{0})$ for all $h\leq h_{0}$,
and $p_{h}\neq q_{h}$ as well.

\end{proof}